\documentclass[10pt,twocolumn,letterpaper]{article}
\usepackage[pagenumbers]{cvpr}

\usepackage{algorithm,algpseudocode}
\DeclareMathOperator{\SiLU}{SiLU}
\DeclareMathOperator{\LN}{LN}
\DeclareMathOperator{\entmax}{entmax}
\DeclareMathOperator{\sg}{sg}
\DeclareMathOperator{\normtwo}{norm_2}
\newcommand{\R}{\mathbb{R}}

\usepackage{multirow}

\usepackage{stfloats}
\definecolor{cvprblue}{rgb}{0.21,0.49,0.74}
\usepackage{xr}
\usepackage[breaklinks,colorlinks,allcolors=cvprblue]{hyperref}

\title{Interpretable Prototype-Conditioned Adaptation for Weather Forecasting}
\author{%
Zhixuan Li$^{1}$ \quad Hao Jiang$^{1}$ \quad Chen Hui$^{2}$ \quad Kai Geng$^{2}$ \quad Weisi Lin$^{1}$\thanks{Corresponding author.}\\[0.4em]
{\normalsize $^{1}$Nanyang Technological University, Singapore}\\
{\normalsize $^{2}$Nanjing University of Information Science and Technology, China}
}
\hypersetup{
  pdftitle={Interpretable Prototype-Conditioned Adaptation for Weather Forecasting},
  pdfauthor={Zhixuan Li, Hao Jiang, Chen Hui, Kai Geng, Weisi Lin}
}

\begin{document}
\maketitle
\begin{abstract}
Data-driven weather forecasting requires distinguishing local states and their temporal evolution. However, learned patterns remain implicit in neural forecasters, making their influence difficult to interpret. Failing to distinguish local evolution can lead to similar forecasts for locations that develop differently, increasing prediction error. This motivates explicit references for distinguishing and interpreting local patterns. Prototypes provide explicit pattern references, but sample- or task-level conditioning leaves open how to adapt to distinct locations and variable groups. To deal with this problem, we design ProtoCast, which uses group-specific prototype banks to adapt local features while accounting for cross-variable dependencies. Specifically, (1) we combine each group's historical features and temporal change with forecast context from cross-variable attention for evolution-aware prototype matching. (2) Sparse mixtures of prototype-specific low-rank operators tailor corrections to each group and location. A gated attention block then exchanges the corrected features across groups before prediction. Each group's correction decomposes into explicit prototype contributions before this fusion. (3) We jointly train the forecasting network and operators from scratch and refit each group's prototypes before every training epoch. Slot tracking preserves prototype-operator associations, making each prototype's contribution to local corrections more interpretable. Experiments on GlobalForecast and TyphoonTracking show improved forecasting accuracy, while prototype deletion links local feature contributions to changes in prediction error. This work advances the development of more accurate and interpretable data-driven weather forecasting models. Project page: \url{https://zhixuanli.github.io/projects/li2026protocast}.
\par
\end{abstract}

\section{Introduction}
\label{sec:intro}

\begin{figure*}[t]
    \centering
    \includegraphics[width=\textwidth]{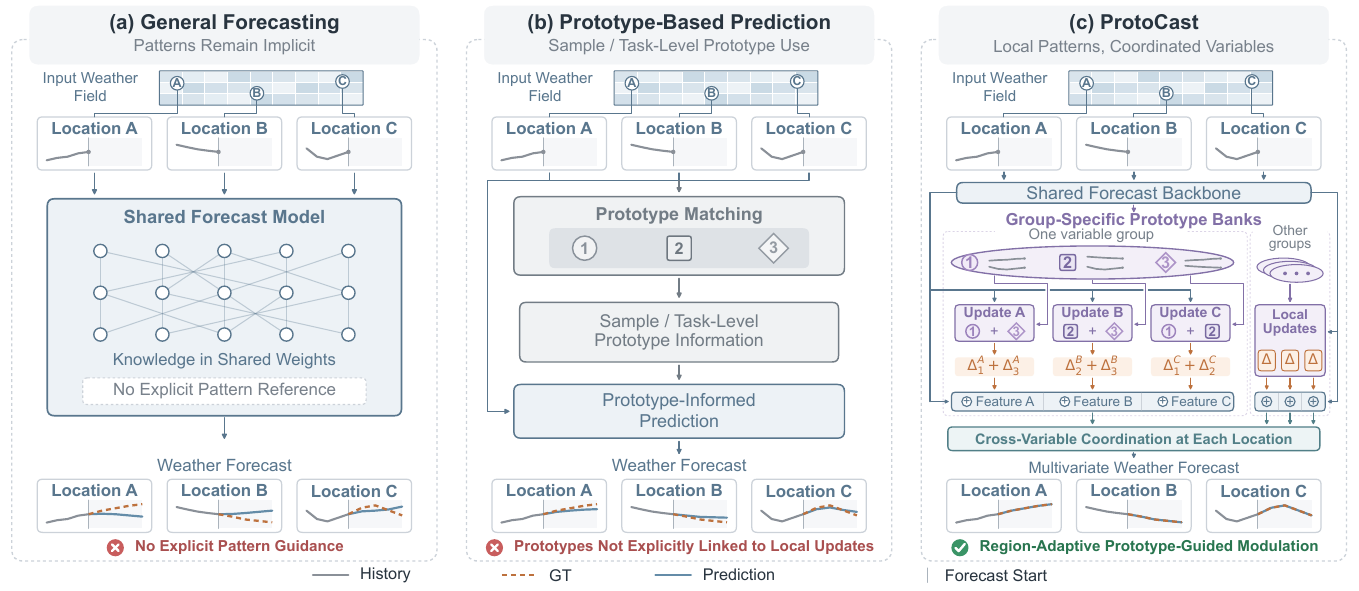}
    \caption{Motivation for local prototype-conditioned forecasting. (a) A general forecaster represents pattern knowledge implicitly in shared parameters. (b) Prototype matching provides information for sample- or task-level prediction. (c) ProtoCast uses group-specific prototype banks, with one group expanded and the others summarized. Locations A, B, and C form distinct operator mixtures from the same group-specific bank. The operators process local features to produce the \textcolor[HTML]{BD713D}{orange} weighted contributions. At each location, the contributions within each group are summed and added to its features before cross-variable coordination. Curves beside the prototypes depict representative histories within the expanded group. Solid \textcolor[HTML]{8A9098}{gray} curves show historical inputs. Dashed \textcolor[HTML]{BD713D}{orange} and solid \textcolor[HTML]{648CA8}{blue} curves denote \textcolor[HTML]{BD713D}{ground truth} and \textcolor[HTML]{648CA8}{predictions}, respectively.}
    \label{fig:motivation}
\end{figure*}

Data-driven weather forecasting uses historical multivariate observations to predict future meteorological fields. Forecasting models aim to capture both spatial relationships and temporal evolution. Pangu-Weather~\cite{bi2023pangu} and GraphCast~\cite{lam2023graphcast} demonstrate progress in data-driven weather prediction. Within a weather field, different locations exhibit distinct states and evolution patterns. Forecasting models therefore need to distinguish these local evolution patterns. Making their influence on predictions explicit also helps interpret model behavior.

Representative Transformer forecasters such as Earthformer~\cite{gao2022earthformer} and PredFormer~\cite{tang2026predformer} model these spatial relationships and local evolution patterns through spatiotemporal attention. However, the learned patterns remain implicit in features and parameters, as illustrated in Figure~\ref{fig:motivation}(a). This makes it difficult to trace which representative patterns influence predictions at a particular location. Additionally, if these local evolution patterns are not adequately distinguished, the model can produce similar forecasts for states that develop differently, increasing prediction error. 
 
To support this distinction, explicit references in the form of prototypes can help relate local states to representative patterns. A common use of these references is to guide sample- or task-level prediction, as illustrated in Figure~\ref{fig:motivation}(b). For example, ProtoPNet~\cite{chen2019protopnet} aggregates similarities between image parts and prototypes into image-level classification evidence, while ProtoTS~\cite{peng2026protots} combines hierarchical temporal prototypes to construct future sequences. HyperAdapter~\cite{zhao2023hyperadapter} uses task prototypes to generate adapter parameters. However, using prototype information at the sample or task level leaves open \textit{how different positions within a weather field should be updated according to their own evolution}. Different locations can follow different trajectories even when their present states are similar. At the same location, meteorological variables also exhibit distinct but interdependent dynamics. An intuitive solution is to retain separate references for variable groups and use them to adapt each location's features, while exchanging information across variables. This adapts feature updates to local dynamics for more accurate and interpretable forecasts.

To realize this adaptation, we propose ProtoCast, a spatiotemporal forecasting framework that retains distinct representations and prototype banks for meteorological variable groups. As illustrated in Figure~\ref{fig:motivation}(c), each group shares its prototype bank across locations, while local evolution determines the prototype mixture used at each location. Our design combines group-specific low-rank residual corrections with cross-variable interactions before and after correction. Tracked prototype refitting preserves the association between pattern references and learned correction operators during joint training.

Specifically, first, group-specific encoders retain separate variable representations, while shared temporal and spatial blocks model their evolution. Pre-correction attention supplies cross-variable context at each patch and lead. Second, evolution-aware descriptors combine history, temporal change, and forecast context to select sparse prototype mixtures. The associated low-rank operators produce group-specific residual corrections. Gated attention exchanges corrected features across groups before final spatial processing and group-specific decoding. Finally, joint training updates the network and correction operators, while epoch-wise refitting and slot tracking preserve prototype-operator associations. Each group's residual decomposes into weighted prototype contributions before fusion, making local corrections more interpretable.

Our contributions can be summarized as follows.
\begin{itemize}
    \item We introduce ProtoCast with group-specific representations and prototype banks alongside shared spatiotemporal modeling. Evolution-aware matching adapts features to distinct dynamics across locations and variable groups.
    \item We combine sparse mixtures of prototype-specific low-rank corrections with cross-variable interactions before and after correction. Tracked refitting preserves prototype-operator associations, making local corrections more interpretable.
    \item Among the compared methods, ProtoCast ranks first on 13 of the 16 metrics on GlobalForecast and all 16 metrics on TyphoonTracking. Prototype deletion reveals how individual local corrections affect forecast accuracy.
\end{itemize}

\section{Related Work}
\label{sec:related_work}

\subsection{Spatiotemporal Weather Forecasting}

Data-driven weather forecasting explores spatial representations and multi-step prediction. FourCastNet~\cite{pathak2022fourcastnet} and SFNO~\cite{bonev2023sfno} use Fourier operators, while Pangu-Weather~\cite{bi2023pangu} and GraphCast~\cite{lam2023graphcast} use 3D Transformers and multiscale graph networks. FuXi~\cite{chen2023fuxi} and FengWu~\cite{chen2025fengwu} reduce longer-range errors through range-specific cascaded models and replayed predictions, respectively. Stormer~\cite{nguyen2024stormer} combines interval randomization and pressure-weighted losses, while NeuralGCM~\cite{kochkov2024neuralgcm} couples differentiable atmospheric dynamics with learned parameterizations. GenCast~\cite{price2025gencast} generates diffusion-based ensembles, and ATLAS~\cite{kossaifi2026atlas} combines downsampled latents with history-conditioned local projection for probabilistic forecasting. ClimaX~\cite{nguyen2023climax} aggregates variables for weather and climate transfer, and Aurora~\cite{bodnar2025aurora} learns unified 3D representations from diverse Earth-system data. WeatherNext~3~\cite{rasp2026weathernext3} incorporates satellite observations for hourly forecasts and learns precipitation, cyclone, and station targets.

Related spatiotemporal models use convolutional recurrence~\cite{shi2015convlstm}, convolutional temporal modeling~\cite{gao2022simvp}, and attention in Earthformer~\cite{gao2022earthformer} and PredFormer~\cite{tang2026predformer}. These architectures capture local variation, but their learned evolution patterns remain implicit in distributed representations. ProtoCast provides explicit group-specific prototype references for local feature correction. Each variable group follows its own dynamics while exchanging information across groups for multivariate forecasting.

\subsection{Prototype Learning in Meteorology}

Prototype learning grounds prediction in shared class references~\cite{snell2017prototypical} or visual parts~\cite{chen2019protopnet,nauta2021prototree,rymarczyk2022protopool,donnelly2022deformable,nauta2023pipnet}. These methods connect prototype matches to classification evidence through similarities, tree decisions, or sparse weights. ProtoTS~\cite{peng2026protots} combines hierarchical temporal prototypes for sequence forecasting.

In meteorology, ProtoLNet~\cite{barnes2022protolnet} uses location-scaled matching to identify Madden-Julian oscillation phases. Its subseasonal-to-seasonal application~\cite{gordillo2025protolnet} predicts temperature-anomaly classes using precipitation-anomaly prototypes, transferring the model from CESM2 to ERA5. Mechanism Learning~\cite{jiang2026mechanism} combines historical queries and prototypes into forecasting representations. These methods form class evidence or forecast representations.

Prototype-based HyperAdapter~\cite{zhao2023hyperadapter} generates adapter parameters from task prototypes and layer information through a hypernetwork, but selecting a single prototype when adapting to a new task can leave useful information from other prototypes unused. ProtoCast matches evolving local states to separate variable-group prototype banks and combines multiple low-rank correction operators within the forecasting network. Cross-variable context informs matching, and gated fusion exchanges corrected features before prediction. Each correction decomposes into prototype contributions, making local updates more interpretable.

\begin{figure*}[t]
    \centering
    \includegraphics[width=\textwidth]{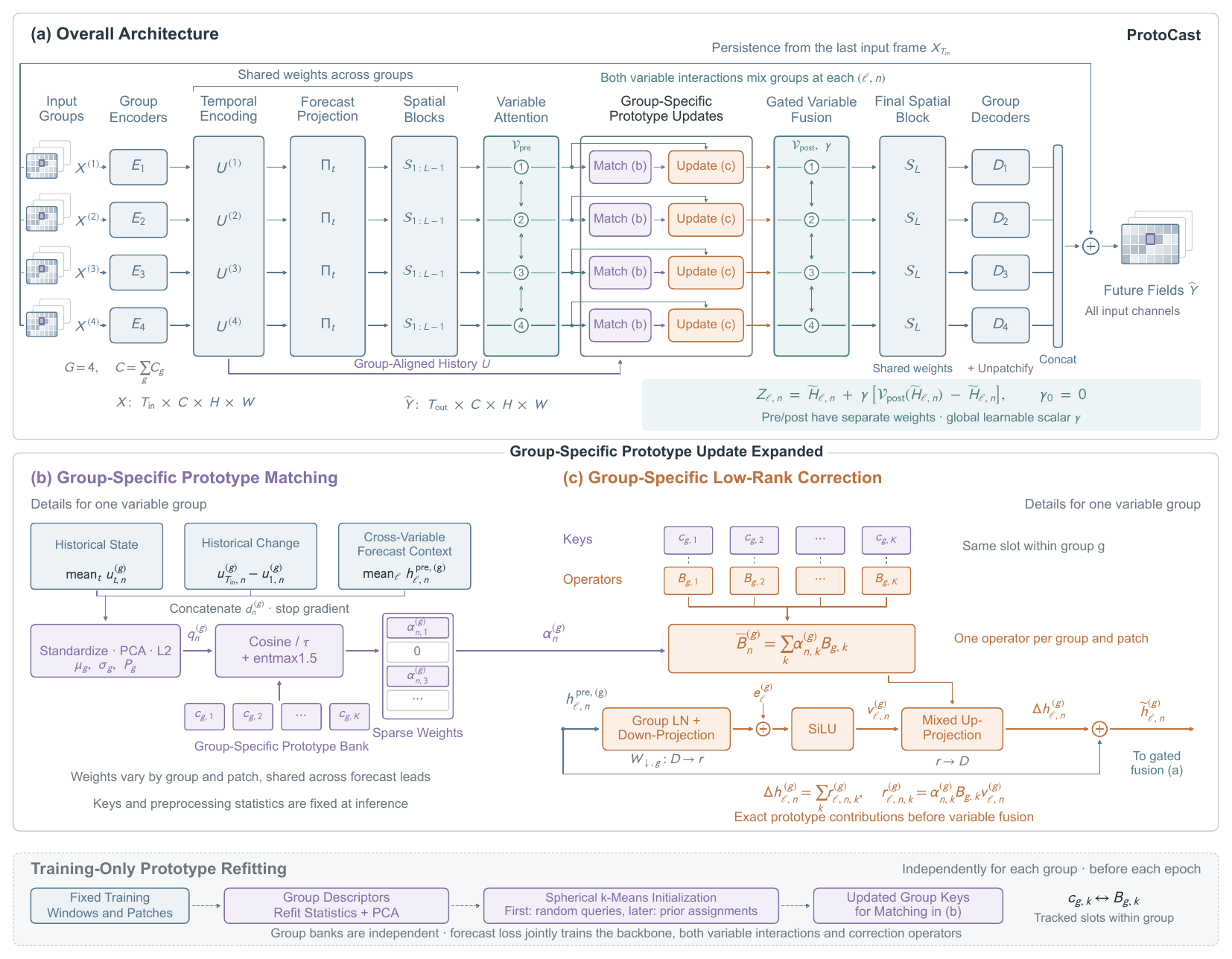}
    \caption{ProtoCast with four variable groups. (a) Groups use independent encoders, decoders, and prototype banks with shared temporal and spatial modules. Cross-variable attention provides matching context, while gated fusion exchanges corrected features. The last input supplies persistence. Panels (b) and (c) detail one group. (b) History, temporal change, and cross-variable forecast context yield sparse, patch-specific weights shared across forecast leads. (c) Mixed operators map lead-conditioned low-rank features to residuals with explicit prototype contributions before fusion. The bottom strip shows group-wise refitting of matching statistics and keys before each epoch. Previous assignments keep keys paired with their operators across refits.}
    \label{fig:method}
\end{figure*}

\section{Method}
\label{sec:method}

\subsection{Problem Formulation}
\label{sec:problem_formulation}

Given $T_{\mathrm{in}}$ historical weather fields $X\in\R^{T_{\mathrm{in}}\times C\times H\times W}$, we predict $\widehat{Y}=f_{\theta,\phi}(X)\in\R^{T_{\mathrm{out}}\times C\times H\times W}$. Here $C$ counts meteorological channels, $H\times W$ is the spatial grid, and $Y$ contains the next $T_{\mathrm{out}}$ observations. Parameters $\theta$ govern forecasting and cross-variable interactions, while $\phi$ governs group-specific corrections.

\subsection{Overall Architecture}
\label{sec:method_overall_architecture}

Figure~\ref{fig:method}(a) summarizes ProtoCast, which combines shared spatiotemporal modeling with group-specific prototype corrections. First, group encoders convert the historical fields into patch features. Shared temporal modeling, forecast projection, and spatial modeling then produce forecast representations. A first cross-variable attention block supplies information from related variables before prototype matching. Next, for each group and patch, the matching branch in Figure~\ref{fig:method}(b) combines historical features, temporal change, and forecast context to obtain sparse prototype weights. These weights combine the low-rank operators in Figure~\ref{fig:method}(c) to correct the local forecast features. Finally, gated cross-variable fusion exchanges the corrected features, followed by the final spatial block and group-specific decoders. The decoded changes are added to the last observed fields to obtain the forecasts. The first cross-variable interaction thus informs pattern selection, while the second integrates the feature corrections before prediction.

During training, the network and correction operators are optimized jointly, and tracked refitting updates the prototype banks shown at the bottom of Figure~\ref{fig:method}. The banks remain fixed at inference.

\subsection{Group-Structured Spatiotemporal Modeling}
\label{sec:method_overview}
\label{sec:method_group_modeling}

Since different meteorological variables follow distinct dynamics, we preserve group-specific feature streams throughout forecasting. Shared temporal and spatial modules model common structure, and cross-variable attention provides information from related fields.

We split $X$ into $G$ variable groups $X^{(g)}$, each with $C_g$ channels. Each group uses $N=(H/p)(W/p)$ non-overlapping $p\times p$ patches with $D$-dimensional features.

Inspired by factorized spatiotemporal modeling in PredFormer~\cite{tang2026predformer}, we design separate group encoders and decoders with shared temporal and spatial modules (Figure~\ref{fig:method}(a)). The encoder $E_g$ maps each flattened patch to $D$ dimensions. Shared temporal blocks $\mathcal{T}$ encode history, a learned projection $\Pi_t$ maps features to future leads, and the first $L-1$ of $L$ shared spatial blocks process the result:
\begin{equation}
\begin{aligned}
U^{(g)}&=\mathcal{T}\!\left(E_g(\operatorname{Patch}(X^{(g)}))+P_t+P_s\right),\\
V^{(g)}&=\mathcal{S}_{1:L-1}\!\left(\Pi_t(U^{(g)})\right).
\end{aligned}
\label{eq:method_group_features}
\end{equation}
Here $U^{(g)}\in\R^{T_{\mathrm{in}}\times N\times D}$ contains group $g$'s temporally encoded historical patch features. The tensor $V^{(g)}\in\R^{T_{\mathrm{out}}\times N\times D}$ contains its forecast patch features after temporal projection and spatial modeling. $P_t$ and $P_s$ are temporal and spatial positional encodings. Cross-variable attention then exchanges information at each patch $n$ and lead $\ell$:
\begin{equation}
H^{\mathrm{pre}}_{\ell,n}=\mathcal{V}_{\mathrm{pre}}(V_{\ell,n}),\qquad
V_{\ell,n}\in\R^{G\times D}.
\label{eq:method_pre_fusion}
\end{equation}
Here $V_{\ell,n}$ stacks group features. The vector $u_{t,n}^{(g)}$ is the historical feature of patch $n$ at time $t$, and $h_{\ell,n}^{(g)}$ is group $g$'s forecast feature in $H^{\mathrm{pre}}_{\ell,n}$. These forecast features provide cross-variable matching context and receive group-specific prototype corrections. Section~\ref{sec:supp_hyperparameters} details tensor shapes, normalization, and feature computation.

\subsection{Evolution-Aware Prototype Matching}
\label{sec:method_matching}

As local states that look similar can follow different trajectories, prototype selection must account for their temporal evolution. We therefore match each group's history and forecast context to its own prototype bank, a set of shared reference vectors for that group's local evolution patterns. For patch $n$ in group $g$, the descriptor $d_n^{(g)}\in\R^{3D}$ in Figure~\ref{fig:method}(b) combines the mean historical feature, its endpoint change, and the mean forecast feature after the first cross-variable interaction.
The historical mean feature summarizes the local state, while the endpoint difference retains its net temporal change. The forecast-context term incorporates information from related variables, allowing matching to account for dependencies between fields. All terms are computed from the historical input $X$.

We standardize each group's descriptors using training statistics, apply principal component analysis (PCA), and normalize the projected vectors to unit length, obtaining queries $q_n^{(g)}\in\R^{d_q}$. Matching receives no prediction-loss gradients. Section~\ref{sec:supp_hyperparameters} details the descriptor and query construction.
Group $g$ has $K$ normalized prototype keys $\mathcal{C}_g=\{c_{g,k}\}_{k=1}^{K}$, with $c_{g,k}\in\R^{d_q}$. The keys are shared across samples and locations within that group. Cosine-based scores determine sparse weights through entmax~\cite{peters2019sparse}:
\begin{equation}
\begin{aligned}
s_{n,k}^{(g)}&=(q_n^{(g)})^\top c_{g,k}/\tau,\qquad
\alpha_n^{(g)}=\entmax(s_n^{(g)}),\\
\alpha_{n,k}^{(g)}&\geq 0,\qquad \sum_{k=1}^{K}\alpha_{n,k}^{(g)}=1,
\end{aligned}
\label{eq:method_routing}
\end{equation}
where $\tau>0$ controls the score scale. Entmax gives low-scoring prototypes exact zero weights while allowing multiple relevant patterns to participate. The mixture varies across groups and patches, but $\alpha_n^{(g)}$ is shared across all forecast leads at that patch. Each group's local trajectory therefore uses a consistent set of pattern references. Representative training samples characterize the meteorological patterns associated with these keys, making the selected pattern references more interpretable through observed historical trajectories.

\subsection{Prototype-Conditioned Low-Rank Correction}
\label{sec:method_correction}

To turn the selected pattern mixture into a feature update, we associate each prototype with a low-rank residual operator. Combining these operators with the matching weights makes the correction specific to each group and location.

In the correction branch of Figure~\ref{fig:method}(c), each key $c_{g,k}$ indexes a learned up-projection $B_{g,k}\in\R^{D\times r}$, where $r\ll D$. The prototypes within group $g$ share a down-projection $W_{\downarrow,g}\in\R^{r\times D}$, layer normalization $\LN_g$, and lead embeddings $e_\ell^{(g)}\in\R^r$. These parameters are independent across groups. We apply the shared down-projection to normalized forecast features and add the lead embedding before SiLU, obtaining the low-rank feature $v_{\ell,n}^{(g)}\in\R^r$. The prototype weights combine the up-projections, and the resulting operator maps this feature to a residual $\Delta h_{\ell,n}^{(g)}$. Adding the residual to $h_{\ell,n}^{(g)}$ gives the corrected feature $\widetilde{h}_{\ell,n}^{(g)}$.

The shared low-rank representation keeps each prototype-specific up-projection compact. The forecast features and lead embeddings allow a shared pattern mixture to produce lead-specific corrections. Each operator mixture is formed once per patch and reused across forecast leads. Each correction branch initially preserves its input. Section~\ref{sec:supp_hyperparameters} gives formulas and parameter counts.

\subsection{Post-Correction Variable Fusion}
\label{sec:method_fusion}

The prototype branch updates each variable group separately. To account for dependencies among the corrected variables, a second cross-variable block exchanges their updated features before decoding. The earlier interaction informs prototype selection, while this block integrates the resulting corrections. Let $\widetilde{H}_{\ell,n}\in\R^{G\times D}$ stack all groups at patch $n$ and lead $\ell$. We compute:
\begin{equation}
\begin{aligned}
Z_{\ell,n}&=\widetilde{H}_{\ell,n}+\gamma R_{\ell,n},\\
R_{\ell,n}&=\mathcal{V}_{\mathrm{post}}(\widetilde{H}_{\ell,n})-\widetilde{H}_{\ell,n}.
\end{aligned}
\label{eq:method_post_fusion}
\end{equation}
Here $\gamma$ is a shared trainable scalar, and $\mathcal{V}_{\mathrm{post}}$ follows the structure of $\mathcal{V}_{\mathrm{pre}}$ with independent parameters. Subtracting $\widetilde{H}_{\ell,n}$ isolates the block's update. The learned gate controls the strength of this exchange while retaining a direct path for each group's corrected features.

Let $Z_\ell^{(g)}$ collect group $g$'s fused patch features. The final shared spatial block $\mathcal{S}_L$ and group-specific decoder $D_g$ reconstruct the forecast:
\begin{equation}
\widehat{Y}_\ell^{(g)}=\operatorname{Unpatch}\!\left(D_g(\mathcal{S}_L(Z_\ell^{(g)}))\right)+X_{T_{\mathrm{in}}}^{(g)}.
\label{eq:method_prediction}
\end{equation}
Here $\operatorname{Unpatch}$ restores the spatial grid. Group outputs are assembled in their original channel order. Adding the last observed fields at every lead makes the decoders predict changes from the last input.

\begin{table*}[!t]
\centering
    \caption{Global multivariate forecasting on GlobalForecast. Models use 12 hourly observations to predict leads of 1-12 hours. Scores are computed after inverse normalization. Lower errors and higher $R^2$ indicate better predictions. The \textcolor{cvprblue}{\textbf{best}} and \underline{second-best} values are highlighted. The 1st Count column records the number of best scores for each method.}
    \label{tab:weatherbench_main}
    \small
    \setlength{\tabcolsep}{2pt}
    \renewcommand{\arraystretch}{1.12}
    \resizebox{\textwidth}{!}{%
    \begin{tabular}{ll*{16}{r}c}
    \toprule
    \multirow[c]{2}{*}{Method} & \multirow[c]{2}{*}{Venue} & \multicolumn{4}{c}{T2M} & \multicolumn{4}{c}{R} & \multicolumn{4}{c}{UV10} & \multicolumn{4}{c}{TCC} & \multirow[c]{2}{*}{\shortstack{1st\\Count}} \\
    \cmidrule(lr){3-6}\cmidrule(lr){7-10}\cmidrule(lr){11-14}\cmidrule(lr){15-18}
     &  & MSE$\downarrow$ & MAE$\downarrow$ & RMSE$\downarrow$ & $R^2\!\uparrow$ & MSE$\downarrow$ & MAE$\downarrow$ & RMSE$\downarrow$ & $R^2\!\uparrow$ & MSE$\downarrow$ & MAE$\downarrow$ & RMSE$\downarrow$ & $R^2\!\uparrow$ & MSE$\downarrow$ & MAE$\downarrow$ & RMSE$\downarrow$ & $R^2\!\uparrow$ &  \\
    \midrule
    ConvLSTM~\cite{shi2015convlstm} & NeurIPS'15 & 2.8330 & 1.2108 & 1.6831 & 0.9936 & 22.2595 & 3.4032 & 4.7180 & 0.9193 & 1.9612 & 0.9698 & 1.4004 & 0.9155 & 0.0211 & 0.1050 & 0.1454 & 0.7695 & 0 \\
    PredRNN~\cite{wang2017predrnn} & NeurIPS'17 & 2.4768 & 1.1439 & 1.5738 & 0.9944 & 19.0061 & 3.0984 & 4.3596 & 0.9311 & 1.6057 & 0.8767 & 1.2671 & 0.9308 & 0.0200 & 0.1025 & 0.1413 & 0.7824 & 0 \\
    MAU~\cite{chang2021mau} & NeurIPS'21 & 4.5610 & 1.4979 & 2.1357 & 0.9897 & 30.4721 & 4.0525 & 5.5202 & 0.8896 & 2.5267 & 1.1423 & 1.5895 & 0.8911 & 0.0234 & 0.1084 & 0.1531 & 0.7444 & 0 \\
    SimVP~\cite{gao2022simvp} & CVPR'22 & 3.4210 & 1.3681 & 1.8496 & 0.9923 & 26.7747 & 3.8266 & 5.1744 & 0.9030 & 2.4634 & 1.1450 & 1.5695 & 0.8938 & 0.0291 & 0.1283 & 0.1707 & 0.6825 & 0 \\
    ConvNeXt~\cite{liu2022convnext} & CVPR'22 & 2.7980 & 1.2215 & 1.6727 & 0.9937 & 23.1956 & 3.5768 & 4.8162 & 0.9159 & 1.7352 & 0.9702 & 1.3173 & 0.9252 & 0.0242 & 0.1151 & 0.1554 & 0.7366 & 0 \\
    HorNet~\cite{rao2022hornet} & NeurIPS'22 & 2.3653 & 1.1085 & 1.5380 & 0.9947 & 20.5402 & 3.3645 & 4.5321 & 0.9256 & 1.3913 & 0.8670 & 1.1795 & 0.9400 & 0.0221 & 0.1075 & 0.1487 & 0.7591 & 0 \\
    MogaNet~\cite{li2024moganet} & ICLR'24 & 2.3759 & 1.1072 & 1.5414 & 0.9946 & 20.0429 & 3.3261 & 4.4769 & 0.9274 & 1.3874 & 0.8635 & 1.1779 & 0.9402 & 0.0229 & 0.1093 & 0.1512 & 0.7509 & 0 \\
    TAU~\cite{tan2023tau} & CVPR'23 & 2.3887 & 1.1139 & 1.5455 & 0.9946 & 19.6103 & 3.2911 & 4.4283 & 0.9289 & 1.3588 & 0.8559 & 1.1657 & 0.9414 & 0.0218 & 0.1078 & 0.1477 & 0.7621 & 0 \\
    Met$^2$Net~\cite{li2025met2net} & ICCV'25 & \underline{0.9608} & \underline{0.6753} & \underline{0.9802} & \underline{0.9978} & 12.3275 & 2.4849 & 3.5110 & 0.9553 & 1.0297 & 0.7307 & 1.0147 & 0.9556 & 0.0175 & 0.0928 & 0.1324 & 0.8090 & 0 \\
    PredFormer~\cite{tang2026predformer} & TMLR'26 & 1.1814 & 0.7591 & 1.0869 & 0.9973 & \underline{11.0174} & \underline{2.3495} & \underline{3.3193} & \underline{0.9601} & \underline{0.7817} & \underline{0.6163} & \underline{0.8842} & \underline{0.9663} & \textcolor{cvprblue}{\textbf{0.0162}} & \underline{0.0897} & \textcolor{cvprblue}{\textbf{0.1273}} & \textcolor{cvprblue}{\textbf{0.8234}} & 3 \\
    \midrule
    Ours & -- & \textcolor{cvprblue}{\textbf{0.7850}} & \textcolor{cvprblue}{\textbf{0.5611}} & \textcolor{cvprblue}{\textbf{0.8860}} & \textcolor{cvprblue}{\textbf{0.9982}} & \textcolor{cvprblue}{\textbf{10.5224}} & \textcolor{cvprblue}{\textbf{2.2646}} & \textcolor{cvprblue}{\textbf{3.2438}} & \textcolor{cvprblue}{\textbf{0.9619}} & \textcolor{cvprblue}{\textbf{0.7626}} & \textcolor{cvprblue}{\textbf{0.6013}} & \textcolor{cvprblue}{\textbf{0.8733}} & \textcolor{cvprblue}{\textbf{0.9671}} & \underline{0.0165} & \textcolor{cvprblue}{\textbf{0.0892}} & \underline{0.1285} & \underline{0.8199} & \textcolor{cvprblue}{\textbf{13}} \\
    \bottomrule
    \end{tabular}%
    }
\par\vspace{0pt}
\begingroup
\normalsize
\captionsetup{type=figure,position=bottom,skip=1pt}
\centering
\includegraphics[width=\textwidth]{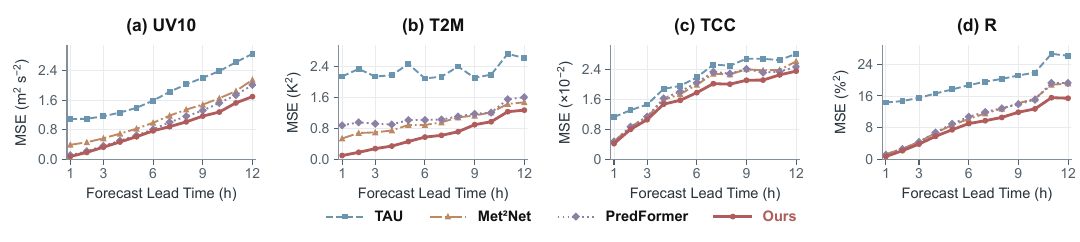}
\caption{MSE across 1-12-hour forecast leads for a group of GlobalForecast test cases. Panels (a)-(d) show UV10, T2M, TCC, and R, respectively, with one case per variable. Errors are computed over the full spatial field after inverse normalization. UV10 averages the squared errors of U10 and V10. Ours achieves the lowest MSE at every lead in the four displayed cases.}
\label{fig:leadwise_mse}
\endgroup
\vspace{-0.7\dbltextfloatsep}
\end{table*}

\subsection{Joint Training and Tracked Prototype Refitting}
\label{sec:method_training}

As the forecasting network learns, the features used for prototype matching change. We therefore refit the prototype banks and track their associations with the learned correction operators as the feature space evolves.

We jointly train the forecasting modules, cross-variable interactions, and correction operators from scratch using normalized-space MSE, with all output-channel elements weighted equally. Matching statistics and keys stay fixed within each epoch, and matching receives no prediction-loss gradients. 
As representations evolve, we refit each group's matching space before every epoch using fixed training samples without gradients. We initialize each center from updated queries assigned to its slot in the previous epoch, preserving the slot's association with its operator $B_{g,k}$ as PCA coordinates change. Maintaining this association lets updated prototype references reuse their learned correction operators across epochs. Validation selects the complete model state, including all group-specific banks, which remain fixed at inference. The supplementary material provides the training procedure in Algorithm~\ref{alg:method_training} and details the training loss and tracked-refitting equations in Section~\ref{sec:supp_training}.

\subsection{Interpretable Prototype Contributions}
\label{sec:method_interpretability}

To interpret a local correction, we need to identify which matched patterns contributed to it and how they affected the features. We therefore decompose the residual before post-correction fusion into weighted prototype contributions $r_{\ell,n,k}^{(g)}$. Each contribution combines its matching weight with the corresponding low-rank operator response.
The weights identify relevant patterns, while residual vectors give their contributions at each forecast lead. Training patches with high query-to-prototype similarity characterize the associated historical evolution. We interpret local corrections by tracing the matched prototypes, their weighted residual contributions, and the forecast changes caused by removing individual contributions.

To examine its effect on prediction, we remove a selected contribution $r_{\ell,n,k}^{(g)}$ from $\widetilde{H}$ and recompute the subsequent layers. Section~\ref{sec:supp_prototype_analysis} gives the decomposition in Equation~\eqref{eq:method_decomposition} and details the removal procedure. Section~\ref{sec:supp_prototype_results} reports its effects on forecast accuracy.

\raggedbottom
\section{Experiments}
\label{sec:experiments}

\subsection{Experimental Setup}
\label{sec:experimental_setup}

\noindent\textbf{Benchmarks.} We evaluate global forecasting on GlobalForecast from WeatherBench~\cite{rasp2020weatherbench} at $32\times64$ resolution and regional forecasting on TyphoonTracking from ERA5~\cite{hersbach2020era5} at $32\times32$ resolution. GlobalForecast includes 2-m temperature (T2M), relative humidity at 1000 hPa (R), 10-m zonal and meridional winds (U10 and V10, jointly reported as UV10), and total cloud cover (TCC). TyphoonTracking includes mean sea level pressure (MSL), U10, V10, and T2M. The two benchmarks therefore cover six distinct meteorological variables. Both benchmarks use 12 hourly observations to predict all channels at leads of 1-12 hours. Section~\ref{sec:supp_data_evaluation} details preprocessing and data splits.

All baselines were retrained using one GPU, BF16, and 50 epochs. Section~\ref{sec:supp_baselines} details their settings.

\noindent\textbf{Evaluation.} Tables~\ref{tab:weatherbench_main} and~\ref{tab:era5_main} report mean squared error (MSE), mean absolute error (MAE), root mean squared error (RMSE), and $R^2$ after inverse normalization, aggregating all test windows, leads, and grid elements. For all methods, the MSE and MAE of UV10 are computed as the arithmetic means of the corresponding U10 and V10 scores. Its RMSE is the square root of the mean MSE, and its $R^2$ averages the component scores with weights proportional to each component's target variance.

\subsection{Global Forecasting}
\label{sec:forecasting_performance}

\begin{figure*}[t]
\centering
    \includegraphics[width=\textwidth]{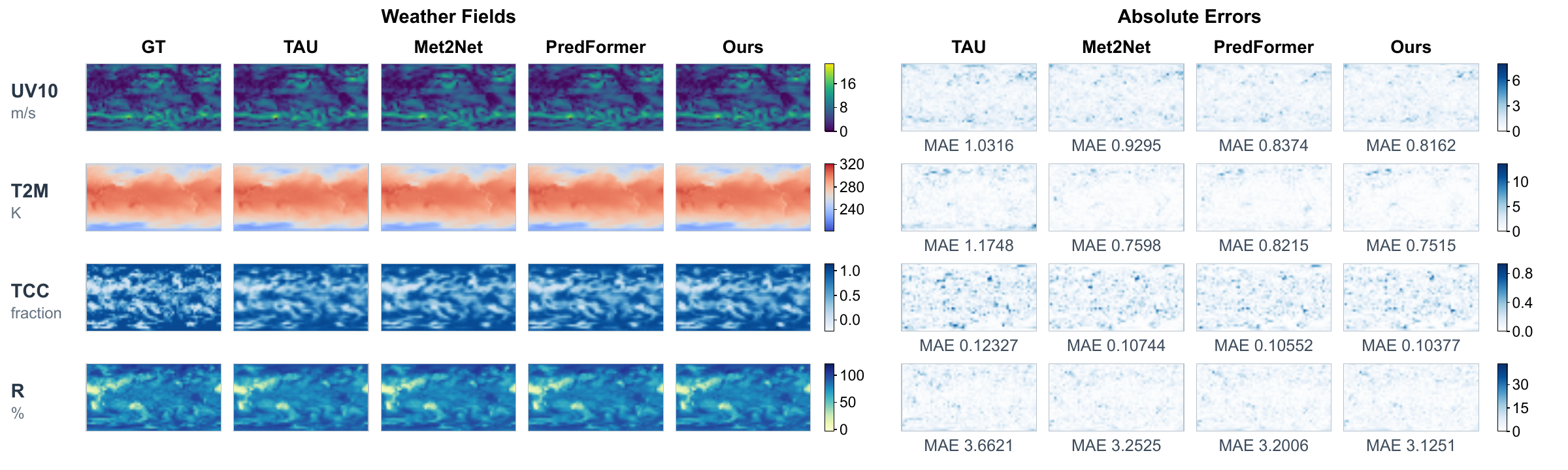}
    \captionsetup{skip=1.4pt}
    \caption{Qualitative comparison of 12-hour forecasts on the GlobalForecast benchmark. The left panels show ground truth (GT) and predictions from TAU, Met$^2$Net, PredFormer, and Ours. The right panels show absolute errors.}
    \label{fig:qualitative_forecast}
\end{figure*}

\setcounter{topnumber}{1}
\begin{figure}[t]
\captionsetup{type=figure}
\addtocounter{figure}{1}
    \centering
    \vspace{3pt}
    \includegraphics[width=\columnwidth]{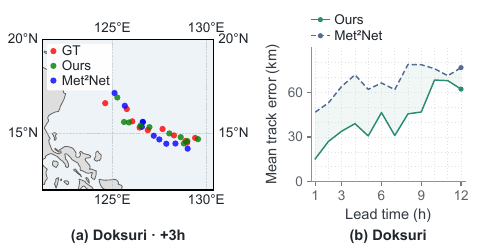}
    \captionsetup{skip=9pt}
    \caption{Typhoon track prediction for Doksuri (2023), comparing Met$^2$Net and Ours. (a) Three-hour predicted centers and ground truth from ERA5 MSL fields. (b) Mean Haversine track error at leads of 1-12 hours (lower is better).}
    \label{fig:typhoon_tracks}
\addtocounter{figure}{-2}
\end{figure}

\begin{figure}[t]
\addtocounter{table}{1}
\captionsetup{type=table}
\centering
\footnotesize
\setlength{\tabcolsep}{2pt}
\renewcommand{\arraystretch}{1.08}
\caption{Core design ablations on TyphoonTracking. MSE is reported for each variable, with lower values indicating better performance. \textcolor{cvprblue}{\textbf{Best}} and \underline{second-best} values are marked.}
\label{tab:ablation_core}
\begin{tabular*}{\columnwidth}{@{\extracolsep{\fill}}lrrrr@{}}
\toprule
Variant & MSL & U10 & V10 & T2M \\
\midrule
w/o correction & 6074.21 & 0.9993 & \underline{1.1361} & 0.5560 \\
Plain adapter & 6117.54 & \underline{0.9957} & 1.1414 & 0.5548 \\
Uniform weights & 6124.51 & 1.0138 & 1.1481 & 0.5605 \\
w/o pre-attention & 6326.96 & 1.0145 & 1.1585 & 0.5758 \\
w/o post-fusion & 6258.18 & 1.0101 & 1.1713 & 0.5626 \\
Fixed gate ($\gamma=1$) & \underline{5993.01} & 1.0006 & 1.1405 & \textcolor{cvprblue}{\textbf{0.5491}} \\
\textbf{Ours} & \textcolor{cvprblue}{\textbf{5888.81}} & \textcolor{cvprblue}{\textbf{0.9886}} & \textcolor{cvprblue}{\textbf{1.1292}} & \underline{0.5529} \\
\bottomrule
\end{tabular*}
\addtocounter{table}{-2}
\vspace{2.73pt}
\end{figure}

\begin{table*}[t]
    \centering
    \caption{Regional multivariate forecasting on TyphoonTracking. Models use 12 hourly observations to predict leads of 1-12 hours. Metrics for MSL, U10, V10, and T2M are computed after inverse normalization. Lower errors and higher $R^2$ indicate better predictions. The \textcolor{cvprblue}{\textbf{best}} and \underline{second-best} values are highlighted. The 1st Count column records the number of best scores for each method.}
    \label{tab:era5_main}
    \small
    \setlength{\tabcolsep}{2pt}
    \renewcommand{\arraystretch}{1.0}
    \resizebox{\textwidth}{!}{%
    \begin{tabular}{ll*{16}{r}c}
    \toprule
    \multirow[c]{2}{*}{Method} & \multirow[c]{2}{*}{Venue} & \multicolumn{4}{c}{MSL} & \multicolumn{4}{c}{U10} & \multicolumn{4}{c}{V10} & \multicolumn{4}{c}{T2M} & \multirow[c]{2}{*}{\shortstack{1st\\Count}} \\
    \cmidrule(lr){3-6}\cmidrule(lr){7-10}\cmidrule(lr){11-14}\cmidrule(lr){15-18}
     &  & MSE$\downarrow$ & MAE$\downarrow$ & RMSE$\downarrow$ & $R^2\!\uparrow$ & MSE$\downarrow$ & MAE$\downarrow$ & RMSE$\downarrow$ & $R^2\!\uparrow$ & MSE$\downarrow$ & MAE$\downarrow$ & RMSE$\downarrow$ & $R^2\!\uparrow$ & MSE$\downarrow$ & MAE$\downarrow$ & RMSE$\downarrow$ & $R^2\!\uparrow$ &  \\
    \midrule
    ConvLSTM~\cite{shi2015convlstm} & NeurIPS'15 & 20742.42 & 103.0576 & 144.0223 & 0.9642 & 2.1290 & 1.0321 & 1.4591 & 0.8879 & 2.3270 & 1.0849 & 1.5254 & 0.8736 & 2.3518 & 1.0257 & 1.5335 & 0.9758 & 0 \\
    PredRNN~\cite{wang2017predrnn} & NeurIPS'17 & 9613.80 & 68.6191 & 98.0500 & 0.9834 & 1.3928 & 0.8253 & 1.1802 & 0.9267 & 1.5688 & 0.8810 & 1.2525 & 0.9148 & 1.0492 & 0.7064 & 1.0243 & 0.9892 & 0 \\
    MAU~\cite{chang2021mau} & NeurIPS'21 & 10361.82 & 69.3611 & 101.7930 & 0.9821 & 1.5269 & 0.8564 & 1.2357 & 0.9196 & 1.7368 & 0.9131 & 1.3179 & 0.9057 & 1.0331 & 0.7067 & 1.0164 & 0.9894 & 0 \\
    SimVP~\cite{gao2022simvp} & CVPR'22 & 16821.50 & 95.4607 & 129.6977 & 0.9710 & 2.1898 & 1.0740 & 1.4798 & 0.8847 & 2.4468 & 1.1429 & 1.5642 & 0.8671 & 1.6990 & 0.9274 & 1.3035 & 0.9825 & 0 \\
    ConvNeXt~\cite{liu2022convnext} & CVPR'22 & 12387.68 & 82.9830 & 111.3000 & 0.9786 & 1.3465 & 0.8196 & 1.1604 & 0.9291 & 1.5473 & 0.8750 & 1.2439 & 0.9160 & 0.9961 & 0.7120 & 0.9981 & 0.9898 & 0 \\
    HorNet~\cite{rao2022hornet} & NeurIPS'22 & 11604.45 & 79.4172 & 107.7239 & 0.9800 & 1.2826 & 0.7957 & 1.1325 & 0.9325 & 1.4777 & 0.8513 & 1.2156 & 0.9197 & 0.9895 & 0.7058 & 0.9947 & 0.9898 & 0 \\
    MogaNet~\cite{li2024moganet} & ICLR'24 & 8894.43 & 67.7602 & 94.3103 & 0.9847 & 1.2165 & 0.7790 & 1.1030 & 0.9360 & 1.3821 & 0.8284 & 1.1756 & 0.9249 & 0.8215 & 0.6348 & 0.9064 & 0.9915 & 0 \\
    TAU~\cite{tan2023tau} & CVPR'23 & 8187.11 & 65.0016 & 90.4827 & 0.9859 & 1.1516 & 0.7579 & 1.0731 & 0.9394 & 1.3247 & 0.8086 & 1.1509 & 0.9281 & \underline{0.7467} & \underline{0.6038} & \underline{0.8641} & \underline{0.9923} & 0 \\
    Met$^2$Net~\cite{li2025met2net} & ICCV'25 & \underline{6902.47} & \underline{60.0469} & \underline{83.0811} & \underline{0.9881} & \underline{1.0825} & \underline{0.7440} & \underline{1.0404} & \underline{0.9430} & \underline{1.2081} & \underline{0.7827} & \underline{1.0991} & \underline{0.9344} & 0.8403 & 0.6404 & 0.9167 & 0.9914 & 0 \\
    PredFormer~\cite{tang2026predformer} & TMLR'26 & 9197.57 & 68.9641 & 95.9040 & 0.9841 & 1.2689 & 0.8009 & 1.1265 & 0.9332 & 1.4384 & 0.8550 & 1.1994 & 0.9219 & 1.2235 & 0.7188 & 1.1061 & 0.9874 & 0 \\
    \midrule
    Ours & -- & \textcolor{cvprblue}{\textbf{5888.81}} & \textcolor{cvprblue}{\textbf{53.1169}} & \textcolor{cvprblue}{\textbf{76.7386}} & \textcolor{cvprblue}{\textbf{0.9898}} & \textcolor{cvprblue}{\textbf{0.9886}} & \textcolor{cvprblue}{\textbf{0.6793}} & \textcolor{cvprblue}{\textbf{0.9943}} & \textcolor{cvprblue}{\textbf{0.9479}} & \textcolor{cvprblue}{\textbf{1.1292}} & \textcolor{cvprblue}{\textbf{0.7248}} & \textcolor{cvprblue}{\textbf{1.0626}} & \textcolor{cvprblue}{\textbf{0.9387}} & \textcolor{cvprblue}{\textbf{0.5529}} & \textcolor{cvprblue}{\textbf{0.4804}} & \textcolor{cvprblue}{\textbf{0.7436}} & \textcolor{cvprblue}{\textbf{0.9943}} & \textcolor{cvprblue}{\textbf{16}} \\
    \bottomrule
    \end{tabular}%
    }

\addtocounter{table}{1}
\end{table*}

\noindent\textbf{Quantitative comparison.} Table~\ref{tab:weatherbench_main} shows that ProtoCast achieves the highest 1st Count among all compared methods, ranking first on 13 of the 16 metrics. For T2M, our method reduces MSE from 0.9608 to 0.7850 and MAE from 0.6753 to 0.5611 compared with Met$^2$Net, corresponding to reductions of 18.30\% and 16.91\%, respectively. For R and UV10, MSE decreases by 4.49\% and 2.44\% compared with PredFormer. For TCC, our method achieves the lowest MAE of 0.0892 and ranks second on MSE, RMSE, and $R^2$. These gains are consistent with evolution-aware matching against group-specific prototype banks in our design, which lets each location combine corrections suited to its variable group and temporal dynamics. Cross-variable attention supplies related-variable information to account for dependencies when adapting local features.

\begingroup
\setlength{\baselineskip}{11.6pt}
\noindent\textbf{Errors across forecast leads.} Figure~\ref{fig:leadwise_mse} compares 1-12-hour full-field MSE for four GlobalForecast test cases, one per variable. In these cases, ProtoCast achieves the lowest MSE at every forecast lead. Three additional groups of cases are provided in Section~\ref{sec:supp_leadwise_mse}.
\par
\endgroup

\noindent\textbf{Qualitative comparison.} Figure~\ref{fig:qualitative_forecast} compares 12-hour forecasts on GlobalForecast. We visualize UV10 as wind speed $\sqrt{u^2+v^2}$ and compute its error map as $(|\hat{u}-u|+|\hat{v}-v|)/2$, where hats denote predictions. For scalar variables, the error maps show absolute differences from ground truth in physical units. Each reported MAE averages the corresponding errors over the full $32\times64$ field. Our method achieves the lowest full-field MAE for UV10, T2M, TCC, and R. Six additional test examples and their error maps are provided in Section~\ref{sec:supp_visualizations}.

\subsection{Regional Forecasting}
\label{sec:regional_forecasting}

\noindent\textbf{Quantitative comparison.} Table~\ref{tab:era5_main} shows that ProtoCast ranks first on all 16 metrics for MSL, U10, V10, and T2M. Compared with Met$^2$Net, our method reduces the MSE of MSL from 6902.47 to 5888.81, a reduction of 14.69\%, and lowers the MSEs of U10 and V10 by 8.67\% and 6.53\%, respectively. For T2M, MSE decreases from 0.7467 to 0.5529 compared with TAU, representing the largest relative MSE reduction of 25.95\%. The MAE of T2M also decreases from 0.6038 to 0.4804, a reduction of 20.44\%. ProtoCast improves forecasts for all four regional variables, including pressure, both wind components, and temperature.

\noindent\textbf{Typhoon track forecasting.}\label{sec:typhoon_track_prediction} We examine typhoon track forecasts through case studies of Doksuri and Mawar (2023). ProtoCast is compared with Met$^2$Net on Doksuri and with TAU on Mawar, using the same minimum-pressure procedure to extract centers from physical-scale MSL forecasts and ERA5 targets. Track error is their Haversine distance in kilometers.

Figure~\ref{fig:typhoon_tracks} shows the Doksuri comparison. We initialize forecasts every six hours, using the preceding 12 hourly observations to predict the next 12 hours at each start time. The map shows the three-hour predicted centers, and the curves show mean track errors at each lead. Averaged over the 10 evaluated Doksuri start times, ProtoCast achieves lower mean track error than Met$^2$Net at each forecast lead from 1 to 12 hours. The Mawar comparison is provided in Figure~\ref{fig:supp_typhoon_mawar}. Case windows and center extraction are detailed in Section~\ref{sec:supp_typhoon_evaluation}.

\setcounter{topnumber}{2}
\begin{figure*}[!t]
    \centering
    \includegraphics[width=\textwidth]{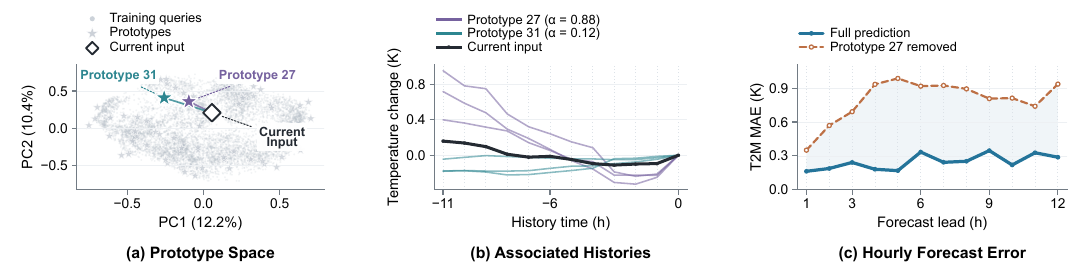}
    \captionsetup{skip=2pt}
    \caption{Interpreting a local T2M prototype contribution. (a) PCA projection of training queries, prototype keys, and the current query. (b) Current and retrieved histories, with original-space prototype matching weights. (c) MAE of T2M w/ and w/o the local weighted Prototype 27 contribution. MAE averages gridwise errors over the $4\times4$ patch at each lead. Lower MAE is better.}
    \label{fig:prototype_t2m_case}
\addtocounter{figure}{1}
\end{figure*}

\subsection{Ablation Studies}

\label{sec:ablation_studies}

We ablate ProtoCast on TyphoonTracking using the training and evaluation protocol of Table~\ref{tab:era5_main}. As a result, Table~\ref{tab:ablation_core} compares prototype correction, cross-variable attention, and gated fusion through MSE for all four variables. Section~\ref{sec:supp_ablation_protocol} provides the settings and analysis for each row. Sections~\ref{sec:supp_ablation_routing}-\ref{sec:supp_ablation_complete} report additional ablations of matching, refitting, routing, and model capacity, with all four metrics.

\subsection{Prototype Contribution Analysis}
\label{sec:prototype_analysis}

To analyze prototype contributions, we present a case study in Figure~\ref{fig:prototype_t2m_case} and report further quantitative analyses in the supplementary material. Section~\ref{sec:supp_prototype_analysis} details the evaluation protocol, and Section~\ref{sec:supp_prototype_results} reports results on both benchmarks. In the case study, removing Prototype 27's local weighted correction increases the MAE of T2M at every forecast lead time. Across both benchmarks, removing the largest-magnitude temperature contribution produces a larger mean forecast response and a greater increase in local MAE than removing a randomly selected other active contribution. These findings link prototype contributions to forecast accuracy and make corrections more interpretable.

\par\vspace{-4.5pt}
\section{Conclusion and Future Work}
\label{sec:conclusion}

We presented ProtoCast, which adapts local features to distinct weather dynamics through evolution-aware prototype matching and cross-variable interactions. Each variable group maintains its own prototype bank, and matching combines historical features, temporal changes, and forecast context to distinguish local trajectories. Sparse mixtures of prototype-specific low-rank operators produce corrections that can be decomposed into weighted prototype contributions. Linking these contributions to historical patterns makes local corrections more interpretable, while tracked refitting preserves pattern-operator associations during training. Experiments on GlobalForecast and TyphoonTracking demonstrate improved global and regional forecasting accuracy. Ablations show the benefits of prototype conditioning and cross-variable interactions, and fixed-model tests show that retaining the selected prototype contributions lowers mean local forecast error.

\vspace{-1pt}
Future work will investigate prototype transfer across regions and seasons. Reusing shared weather patterns while adapting prototype banks to local conditions could reduce retraining for new forecasting settings.

\par\medskip
\flushbottom
{
    \fontsize{9}{10.5}\selectfont
    \setlength{\bibsep}{0pt}
    \bibliographystyle{ieeenat_fullname}
    \bibliography{main}
}

\clearpage
\setcounter{figure}{0}
\renewcommand{\thefigure}{S\arabic{figure}}
\renewcommand{\theHfigure}{supp.\arabic{figure}}
\setcounter{table}{0}
\renewcommand{\thetable}{S\arabic{table}}
\renewcommand{\theHtable}{supp.\arabic{table}}
\setcounter{section}{0}
\renewcommand{\thesection}{S\arabic{section}}
\renewcommand{\theHsection}{supp.\arabic{section}}
\setcounter{equation}{0}
\renewcommand{\theequation}{S\arabic{equation}}
\renewcommand{\theHequation}{supp.\arabic{equation}}
\maketitlesupplementary

\raggedbottom
\setcounter{dbltopnumber}{3}
\setcounter{topnumber}{3}

\section*{Contents}
\begingroup
\small
\noindent\hyperref[sec:supp_implementation]{\ref*{sec:supp_implementation}. Implementation Details}\par
\noindent\hspace*{1em}\hyperref[sec:supp_hyperparameters]{\ref*{sec:supp_hyperparameters}. Model and Refitting Settings}\par
\noindent\hspace*{1em}\hyperref[sec:supp_training]{\ref*{sec:supp_training}. Joint Training and Tracked Prototype Refitting}\par
\noindent\hspace*{1em}\hyperref[sec:supp_prototype_analysis]{\ref*{sec:supp_prototype_analysis}. Prototype Contribution Analysis}\par
\noindent\hspace*{1em}\hyperref[sec:supp_data_evaluation]{\ref*{sec:supp_data_evaluation}. Data Preparation and Evaluation}\par
\noindent\hspace*{1em}\hyperref[sec:supp_typhoon_evaluation]{\ref*{sec:supp_typhoon_evaluation}. Typhoon Track Evaluation}\par
\noindent\hspace*{1em}\hyperref[sec:supp_training_cost]{\ref*{sec:supp_training_cost}. Model Size and Training Cost}\par
\noindent\hspace*{1em}\hyperref[sec:supp_baselines]{\ref*{sec:supp_baselines}. Baseline Configurations and Result Sources}\par
\noindent\hyperref[sec:supp_experiments]{\ref*{sec:supp_experiments}. Additional Experiments}\par
\noindent\hspace*{1em}\hyperref[sec:supp_ablation_protocol]{\ref*{sec:supp_ablation_protocol}. Core Ablation Settings and Analysis}\par
\noindent\hspace*{1em}\hyperref[sec:supp_ablation_routing]{\ref*{sec:supp_ablation_routing}. Prototype Matching and Refitting}\par
\noindent\hspace*{1em}\hyperref[sec:supp_descriptor_components]{\ref*{sec:supp_descriptor_components}. Matching Descriptor Components}\par
\noindent\hspace*{1em}\hyperref[sec:supp_router_tracking]{\ref*{sec:supp_router_tracking}. Learned Routing and Slot Tracking}\par
\noindent\hspace*{1em}\hyperref[sec:supp_ablation_capacity]{\ref*{sec:supp_ablation_capacity}. Prototype and Correction Capacity}\par
\noindent\hspace*{1em}\hyperref[sec:supp_ablation_retrieval]{\ref*{sec:supp_ablation_retrieval}. Matching Temperature and Dimension}\par
\noindent\hspace*{1em}\hyperref[sec:supp_ablation_complete]{\ref*{sec:supp_ablation_complete}. Complete Ablation Results}\par
\noindent\hspace*{1em}\hyperref[sec:supp_prototype_results]{\ref*{sec:supp_prototype_results}. Extended Prototype Contribution Results}\par
\noindent\hspace*{1em}\hyperref[sec:supp_history_consistency]{\ref*{sec:supp_history_consistency}. Prototype History Consistency}\par
\noindent\hspace*{1em}\hyperref[sec:supp_pairing_intervention]{\ref*{sec:supp_pairing_intervention}. Prototype-Operator Pairing}\par
\noindent\hspace*{1em}\hyperref[sec:supp_persistence]{\ref*{sec:supp_persistence}. Comparison with Persistence}\par
\noindent\hspace*{1em}\hyperref[sec:supp_area_weighted]{\ref*{sec:supp_area_weighted}. Area-Weighted Global Forecast Evaluation}\par
\noindent\hspace*{1em}\hyperref[sec:supp_correction_effects]{\ref*{sec:supp_correction_effects}. Predictive Role of Prototype Corrections}\par
\noindent\hspace*{1em}\hyperref[sec:supp_training_runs]{\ref*{sec:supp_training_runs}. Comparison Across Training Runs}\par
\noindent\hyperref[sec:supp_visualizations]{\ref*{sec:supp_visualizations}. Additional Visualizations}\par
\noindent\hspace*{1em}\hyperref[sec:supp_forecast_fields]{\ref*{sec:supp_forecast_fields}. Forecast Fields and Spatial Errors}\par
\noindent\hspace*{1em}\hyperref[sec:supp_leadwise_mse]{\ref*{sec:supp_leadwise_mse}. Errors Across Forecast Leads}\par
\noindent\hspace*{1em}\hyperref[sec:supp_typhoon_visualization]{\ref*{sec:supp_typhoon_visualization}. Additional Typhoon Track Prediction}\par
\noindent\hspace*{1em}\hyperref[sec:supp_prototype_cases]{\ref*{sec:supp_prototype_cases}. Prototype Matching and Forecast Contributions}\par
\endgroup

\section{Implementation Details}
\label{sec:supp_implementation}
\subsection{Model and Refitting Settings}
\label{sec:supp_hyperparameters}

\paragraph{Task notation.} The input and output contain the same channels, with $C_{\mathrm{in}}=C_{\mathrm{out}}=\sum_{g=1}^{G}C_g$. Their tensor shapes are:
\begin{equation}
\begin{aligned}
X &\in \R^{T_{\mathrm{in}}\times C_{\mathrm{in}}\times H\times W},\\
X^{(g)} &\in \R^{T_{\mathrm{in}}\times C_g\times H\times W},\\
\widehat{Y}=f_{\theta,\phi}(X) &\in \R^{T_{\mathrm{out}}\times C_{\mathrm{out}}\times H\times W},
\end{aligned}
\label{eq:method_task}
\end{equation}
Here $X^{(g)}$ contains the channels assigned to group $g$, and $Y$ denotes the ground-truth future fields. The parameters $\theta$ include the forecasting modules and cross-variable interactions, while $\phi$ contains the group-specific correction operators. We omit batch indices. Each channel uses its training-set normalization for both observations and targets, as specified in Section~\ref{sec:supp_data_evaluation}.

\paragraph{Variable grouping and sparse matching.} Both benchmarks use $G=4$ variable groups. GlobalForecast groups U10 and V10 together and treats T2M, TCC, and R separately. TyphoonTracking uses one group for each of MSL, U10, V10, and T2M. Sparse matching in Equation~\eqref{eq:method_routing} uses $1.5$-entmax for every group.

\paragraph{Architecture and parameter sharing.} Each variable group has its own patch encoder, decoder, prototype bank, and correction branch. The temporal blocks, time-axis projection, and spatial blocks share parameters across groups. Within a correction branch, layer normalization, the down-projection, and lead embeddings are shared across prototypes, while each prototype has its own up-projection. These correction parameters are independent across groups.

The two cross-variable blocks use separate parameters and attend along the group axis at a fixed patch and forecast lead. The first block follows all but the final spatial block and supplies the forecast context used for matching. The second exchanges the corrected group features through $\gamma$, a single unconstrained trainable scalar shared across samples, groups, patches, and leads. Both blocks contain multi-head self-attention and a gated feed-forward network with residual connections. Equation~\eqref{eq:method_post_fusion} therefore subtracts the block input to isolate the post-correction update before gating.

\paragraph{Group feature extraction.} Equation~\eqref{eq:method_group_features} gives the group-wise feature computation. Temporal blocks operate along the historical time axis at each patch, while spatial blocks operate across patches at each forecast lead. The parameters of $\mathcal{T}$, $\Pi_t$, and $\mathcal{S}$ are shared across groups. The group features have shapes $U^{(g)}\in\R^{T_{\mathrm{in}}\times N\times D}$ and $V^{(g)}\in\R^{T_{\mathrm{out}}\times N\times D}$. Stacking all groups gives $U\in\R^{T_{\mathrm{in}}\times G\times N\times D}$ and $H^{\mathrm{pre}}\in\R^{T_{\mathrm{out}}\times G\times N\times D}$ after pre-correction attention. In Figure~\ref{fig:method}(c), the group-$g$ vector $h_{\ell,n}^{(g)}$ is labeled $h_{\ell,n}^{\mathrm{pre},(g)}$.

\paragraph{Matching descriptor.} We concatenate historical means, endpoint changes, and forecast-context means:
\begin{equation}
\begin{aligned}
\bar{u}_n^{(g)}&=\frac{1}{T_{\mathrm{in}}}\sum_{t=1}^{T_{\mathrm{in}}}u_{t,n}^{(g)},\\
\delta u_n^{(g)}&=u_{T_{\mathrm{in}},n}^{(g)}-u_{1,n}^{(g)},\qquad
\bar{h}_n^{(g)}=\frac{1}{T_{\mathrm{out}}}\sum_{\ell=1}^{T_{\mathrm{out}}}h_{\ell,n}^{(g)},\\
d_n^{(g)}&=\operatorname{concat}(\bar{u}_n^{(g)},\delta u_n^{(g)},\bar{h}_n^{(g)})\in\R^{3D}.
\end{aligned}
\label{eq:method_descriptor} 
\end{equation}

\begin{samepage}
\paragraph{Query construction.} We fit descriptor standardization and PCA independently for each group using training samples, then apply $\ell_2$ normalization:
\begin{equation}
\begin{aligned}
q_n^{(g)}&=\normtwo\!\left(P_g^\top\big[(\sg(d_n^{(g)})-\mu_g)\oslash\sigma_g\big]\right),\\
\normtwo(z)&=\frac{z}{\max(\lVert z\rVert_2,\epsilon)}.
\end{aligned}
\label{eq:method_query}
\end{equation}
\end{samepage}
Here $\mu_g,\sigma_g\in\R^{3D}$ are the group's descriptor mean and standard deviation, $P_g\in\R^{3D\times d_q}$ contains its leading PCA directions, and $\oslash$ denotes elementwise division. A positive floor on $\sigma_g$ and a small $\epsilon$ stabilize normalization. The stop-gradient operator $\sg$ separates prototype retrieval from the differentiable forecast path. Section~\ref{sec:supp_training} specifies the fitting schedule and numerical settings.

\paragraph{Low-rank correction.} Layer normalization precedes a bias-free down-projection. Adding the lead embedding and applying SiLU gives a lead-specific low-rank feature. The prototype weights combine the up-projections before the resulting residual is added to the forecast feature:
\begin{equation}
\begin{aligned}
v_{\ell,n}^{(g)}&=\SiLU\!\left(W_{\downarrow,g}\LN_g(h_{\ell,n}^{(g)})+e_\ell^{(g)}\right),\\
\overline{B}_n^{(g)}&=\sum_{k=1}^{K}\alpha_{n,k}^{(g)}B_{g,k},\\
\Delta h_{\ell,n}^{(g)}&=\overline{B}_n^{(g)}v_{\ell,n}^{(g)},\qquad
\widetilde{h}_{\ell,n}^{(g)}=h_{\ell,n}^{(g)}+\Delta h_{\ell,n}^{(g)}.
\end{aligned}
\label{eq:method_operators}
\end{equation}

\paragraph{Output reconstruction.} In Equation~\eqref{eq:method_prediction}, each group decoder $D_g$ applies layer normalization and a linear projection before spatial reconstruction.

\paragraph{Correction computation and parameter count.} Each operator mixture is formed once per patch and reused across forecast leads, avoiding a separate $D$-dimensional response for every prototype, patch, and lead. The $G$ correction branches contain $G[(K+1)Dr+T_{\mathrm{out}}r+2D]$ trainable parameters, including affine layer normalization. The cross-variable blocks and forecasting modules contribute their own parameters.

\paragraph{Structural settings.} The reference configuration uses $p=4$, $D=192$, four temporal blocks, four spatial blocks, and six heads in each attention block. Each group uses $K=64$ prototypes on GlobalForecast and $K=32$ on TyphoonTracking, with matching dimension $d_q=64$. The correction rank is $r=16$, and the matching temperature is $\tau=0.2$. The sparse weights for a group and patch are reused across all forecast leads, while the lead-specific features and embeddings produce different corrections.

\paragraph{Initialization and refitting frequency.} We initialize all prototype-specific up-projections $B_{g,k}$ and lead embeddings $e_\ell^{(g)}$ to zero. The post-correction fusion gate $\gamma$ is also initialized to zero. These initializations preserve the input of each correction branch and the corrected features entering post-correction fusion. Every group's matching statistics and prototype keys are refitted once before each training epoch and held fixed during that epoch's minibatch updates.

\subsection{Joint Training and Tracked Prototype Refitting}
\label{sec:supp_training}

Algorithm~\ref{alg:method_training} details the training schedule in Section~\ref{sec:method_training}. We specify the optimization settings, prototype refitting, objective, and model selection below.

\paragraph{Training configuration.} We use OpenSTL~\cite{tan2023openstl}. All compared methods use one GPU, BF16 mixed precision, and 50 training epochs. We use Adam~\cite{kingma2015adam} with cosine decay, zero weight decay, and no warmup, gradient clipping, or augmentation. For GlobalForecast, ProtoCast uses learning rate $3\times10^{-3}$ and Adam $\beta_2=0.995$. Batch sizes are 128 for training and 64 for evaluation. Training shuffles windows and drops incomplete batches. Evaluation retains all windows.

\paragraph{Prototype refitting.} We fix an ordered set $\mathcal{S}$ of training windows and patch indices for all epochs. The reference configuration samples 2,048 training windows without replacement and four patches per window, giving 8,192 descriptors per group. All groups use the same sampled windows, patch positions, and ordering. Before each epoch, the current model runs in evaluation mode to recompute each group's descriptors without gradient tracking. For group $g$, we independently refit the descriptor statistics $\mu_g$ and $\sigma_g$ and the PCA projection $P_g$, then obtain normalized queries using Equation~\eqref{eq:method_query}. Sample identities and order stay fixed as the descriptors evolve. Descriptor standardization uses the population standard deviation with a floor of $10^{-4}$. We compute the PCA directions by singular value decomposition and normalize the projected queries to unit length. Each refit runs 25 iterations of spherical $k$-means.

The first fit initializes cluster centers from sampled queries. At refit $e$, let $Q_g^{(e)}=\{q_{g,m}^{(e)}\}_{m=1}^{M}$ denote the updated normalized queries from the same $M$ sampled patches. Later fits initialize centers using the previous nearest-prototype assignments $a_{g,m}^{(e-1)}$:
\begin{equation}
\begin{aligned}
c_{g,k}^{(e,0)}&=\normtwo\!\left(\sum_{m\,\colon\,a_{g,m}^{(e-1)}=k}q_{g,m}^{(e)}\right),\\
(\mathcal{C}_g^{(e)},a_g^{(e)})&=\operatorname{SphericalKMeans}\!\left(Q_g^{(e)},\mathcal{C}_g^{(e,0)}\right).
\end{aligned}
\label{eq:method_refitting}
\end{equation} Spherical $k$-means alternates maximum-cosine assignment and normalized cluster-mean updates independently within each bank. Empty clusters are reseeded from poorly represented queries. Reinitializing clusters independently could associate an existing correction operator with different training patterns. Membership-based initialization instead transfers slot identities into the updated matching space, even when the group's PCA coordinates change. Each up-projection $B_{g,k}$ stays attached to slot $k$ in group $g$ throughout training. Assignments are then updated.

\paragraph{Joint optimization and model selection.} We train ProtoCast from scratch. After refitting, we return the model to training mode and hold all $\mu_g$, $\sigma_g$, $P_g$, and $\mathcal{C}_g$ fixed for the epoch. Each minibatch updates the forecasting modules, cross-variable interactions, and correction operators using normalized-space MSE:
\begin{equation}
\begin{aligned}
\mathcal{L}_{\mathrm{pred}}&=\frac{1}{B T_{\mathrm{out}} C_{\mathrm{out}} H W}\\
&\quad\times\sum_{b,\ell,c,i,j}\big(\widehat{Y}_{b,\ell,c,i,j}-Y_{b,\ell,c,i,j}\big)^2.
\end{aligned}
\label{eq:method_loss}
\end{equation}
Here $B$ is the batch size, $c$ indexes output channels, and $i,j$ index grid cells. The loss averages over all forecast leads, output channels, and grid cells, weighting channel elements equally regardless of group size. This includes both channels in the GlobalForecast wind group. Predictions and targets follow Section~\ref{sec:supp_data_evaluation} for normalization. The stop-gradient in Equation~\eqref{eq:method_query} blocks prediction-loss gradients through matching, while the forecasting and correction modules learn through the residual and prediction paths. Validation uses the same banks as that epoch's training. We retain the best validation checkpoint with all group-specific matching statistics and keys for test inference. Evaluation uses only historical observations.

\setcounter{algorithm}{0}
\renewcommand{\thealgorithm}{S\arabic{algorithm}}
\begin{algorithm}[!t]
\caption{Joint training with tracked group-wise prototype refitting}
\label{alg:method_training}
\small
\begin{algorithmic}[1]
\Require Training set $\mathcal{D}_{\mathrm{tr}}$, validation set $\mathcal{D}_{\mathrm{val}}$, $E$ epochs
\State Initialize forecasting parameters $\theta$, including both variable interactions and $\gamma$, and group-specific correction parameters $\phi$
\State Set $B_{g,k}=0$, $e_\ell^{(g)}=0$, and $\gamma=0$
\State Fix ordered training-window and patch samples $\mathcal{S}$
\For{$e=1,\ldots,E$}
  \State Extract all groups' descriptors on $\mathcal{S}$ in evaluation mode without gradients
  \For{each variable group $g$}
    \State Refit $\mu_g$, $\sigma_g$, and $P_g$, then compute $Q_g^{(e)}$
    \If{$e=1$}
      \State Initialize centers from sampled queries in group $g$
    \Else
      \State Initialize centers using $a_g^{(e-1)}$ and Equation~\eqref{eq:method_refitting}
    \EndIf
    \State Run spherical $k$-means and store $\mathcal{C}_g^{(e)}$ and $a_g^{(e)}$
  \EndFor
  \State Return to training mode and fix all groups' matching statistics and keys
  \For{each training batch $(X,Y)$}
    \State Encode groups, apply shared temporal blocks and forecast projection, then the first $L-1$ spatial blocks
    \State Apply pre-correction attention across groups
    \State Match stop-gradient descriptors to each group's bank and apply its prototype-weighted correction
    \State Apply gated fusion, the final spatial block, and group decoders, then add the last input to obtain $\widehat{Y}$
    \State Update $\theta,\phi$ using $\nabla_{\theta,\phi}\mathcal{L}_{\mathrm{pred}}$
  \EndFor
  \State Evaluate in evaluation mode on $\mathcal{D}_{\mathrm{val}}$ and retain the best complete model state
\EndFor
\State \Return Selected $\theta$, $\phi$, and $\{\mu_g,\sigma_g,P_g,\mathcal{C}_g\}_{g=1}^{G}$
\end{algorithmic}
\end{algorithm}

\subsection{Prototype Contribution Analysis}
\label{sec:supp_prototype_analysis}

\paragraph{Frozen models and sampling.} We analyze the validation-selected checkpoints used in Tables~\ref{tab:weatherbench_main} and~\ref{tab:era5_main}, at epochs 49 and 48, respectively. Inference uses evaluation mode and FP32 throughout, with autocast and TF32 disabled and no prototype refitting. For GlobalForecast, we sample 96 test windows from a 24-hour candidate grid, with 12 windows from each quarter of 2017-2018. One spatial patch is sampled uniformly per window and reused for removal tests in all four groups. TyphoonTracking uses 48 windows, sampled equally from four contiguous index ranges of its test split. Of these, 45 have multiple active temperature prototypes and enter the paired comparison. The remaining three single-active cases are retained separately because no other active prototype is available for the random control. Case selection does not use forecast error.

\paragraph{Training pattern references.} For GlobalForecast temperature, we sample 2,048 training windows across the 24 quarters of 2010-2015 and retain all 128 patch queries per window. From these 262,144 queries, each of the 64 prototypes retrieves three high-similarity histories from different dates, giving 192 examples. Retrieval uses the frozen matching space and training histories. Figure~\ref{fig:prototype_contributions}(a) plots each history relative to its first observation, exposing its temporal variation.

\paragraph{Contribution decomposition.} The weighted operator mixture in Equation~\eqref{eq:method_operators} gives an additive decomposition of the residual before post-correction fusion:
\begin{equation}
\begin{aligned}
\Delta h_{\ell,n}^{(g)}&=\sum_{k=1}^{K}r_{\ell,n,k}^{(g)},\\
r_{\ell,n,k}^{(g)}&=\alpha_{n,k}^{(g)}B_{g,k}v_{\ell,n}^{(g)}.
\end{aligned}
\label{eq:method_decomposition}
\end{equation}
Here $r_{\ell,n,k}^{(g)}$ is the weighted contribution of prototype $k$ at patch $n$ and forecast lead $\ell$ in group $g$.

\paragraph{Contribution magnitude and deletion.} We rank active prototypes, defined by $\alpha_{n,k}^{(g)}>0$, using:
\begin{equation}
    m_{n,k}^{(g)}=\frac{1}{T_{\mathrm{out}}}\sum_{\ell=1}^{T_{\mathrm{out}}}\left\|r_{\ell,n,k}^{(g)}\right\|_2.
    \label{eq:supp_contribution_magnitude}
\end{equation}
The largest-$m$ contribution is removed from $\widetilde h_{\ell,n}^{(g)}$ at all 12 forecast leads for one group and patch. Ties use the smaller prototype index. The random control samples uniformly with replacement from the other active prototypes three times, each time deleting one contribution. We retain the original matching weights without renormalization and reuse the pre-correction features and all other contributions. Post-correction fusion, the final spatial block, and every group decoder are recomputed from the features after removal. The resulting forecast changes capture the contribution's effect with routing and model parameters fixed.

\paragraph{Response and error metrics.} Let $\widehat{Y}^{-k}$ and $\widehat{Y}$ denote the predictions with and without contribution removal after inverse normalization. For output variable $v$, we average over the selected patch's $4\times4$ output cells and all 12 forecast leads, denoted by $\Omega_n$. We compute:
\begin{equation}
\begin{aligned}
    D_v^{-k}&=\operatorname{mean}_{\Omega_n}\left|\widehat{Y}_v^{-k}-\widehat{Y}_v\right|,\\
    \Delta\mathrm{MAE}_v^{-k}&=\operatorname{mean}_{\Omega_n}\Bigl(\left|\widehat{Y}_v^{-k}-Y_v\right|\\
    &\hspace{5.5em}-\left|\widehat{Y}_v-Y_v\right|\Bigr).
\end{aligned}
\label{eq:supp_intervention_metrics}
\end{equation}
$D$ measures the forecast response, whereas positive $\Delta\mathrm{MAE}$ means deletion increases error. UV10 averages elementwise absolute differences over U10 and V10. For the random control, we average the three draws within each case before averaging across cases. Both conditions use the same paired cases. Temperature, wind, pressure, and relative humidity use K, m\,s$^{-1}$, Pa, and percentage points, respectively. TCC uses a fraction in $[0,1]$.

\paragraph{Numerical consistency.} We check eight GlobalForecast cases across all four groups and four TyphoonTracking cases. Replaying the full downstream path and applying a zero deletion reproduce the complete forecast exactly. The largest discrepancy when reconstructing a residual from its prototype terms is $4.58\times10^{-5}$ in feature space. These checks verify that removal affects only the selected contribution while preserving the rest of the computation.

\subsection{Data Preparation and Evaluation}
\label{sec:supp_data_evaluation}

\paragraph{Global data.} GlobalForecast is derived from WeatherBench~\cite{rasp2020weatherbench} at $1.40625^\circ$ resolution. Its five channels are 10-m wind components (U10 and V10), 2-m temperature (T2M), total cloud cover (TCC), and relative humidity at 1000 hPa (R). We subsample the original grid to $64\times128$ and average non-overlapping $2\times2$ blocks to obtain a $32\times64$ grid with $5.625^\circ$ spacing. The 2010-2015 training split has 52,559 windows, the 2016 validation split has 8,759 windows, and the 2017-2018 test split has 17,495 windows.

\paragraph{Regional data.} TyphoonTracking follows the regional domain and variable selection of Met$^2$Net~\cite{li2025met2net}, using ERA5~\cite{hersbach2020era5} over the western North Pacific and adjacent East Asia. Its channels are mean sea level pressure (MSL), U10, V10, and T2M. Its $128\times128$ fields are downsampled offline by $4\times4$ block averaging to a $32\times32$ grid with $1^\circ$ spacing. The training, validation, and test splits cover 2017-2021, 2022, and 2023, with 43,801, 8,737, and 8,737 windows, respectively. Both benchmarks use 12 hourly observations to predict all channels at leads of 1-12 hours. Section~\ref{sec:supp_typhoon_evaluation} details track evaluation at leads of 1-12 hours for Mawar and Doksuri.

\paragraph{Normalization.} GlobalForecast channels are standardized using means and population standard deviations recomputed on the downsampled training split. TyphoonTracking uses channel-wise min-max normalization with training-set extrema. Validation and test data reuse the corresponding training statistics. We invert these transformations before computing the forecast metrics.

\paragraph{Metric aggregation.} MSE, MAE, RMSE, and $R^2$ are computed over all test windows, forecast leads, and grid elements after inverse normalization. For UV10, MSE and MAE are the arithmetic means of the corresponding U10 and V10 scores, and RMSE follows from the mean MSE:
\begin{equation*}
\begin{aligned}
\mathrm{MSE}_{\mathrm{UV10}} &= \tfrac12(\mathrm{MSE}_{U}+\mathrm{MSE}_{V}),\\
\mathrm{MAE}_{\mathrm{UV10}} &= \tfrac12(\mathrm{MAE}_{U}+\mathrm{MAE}_{V}),\\
\mathrm{RMSE}_{\mathrm{UV10}} &= \sqrt{\mathrm{MSE}_{\mathrm{UV10}}},
\end{aligned}
\end{equation*}
With $U$ and $V$ denoting U10 and V10, $R^2$ is:
\begin{equation}
    R^2_{\mathrm{UV10}} = 1 - \frac{\mathrm{SSE}_U + \mathrm{SSE}_V}{\mathrm{SST}_U + \mathrm{SST}_V},
\end{equation}
where $\mathrm{SSE}$ denotes the sum of squared prediction errors and $\mathrm{SST}$ denotes the total sum of squares, with each component centered at its own test-set mean. This is equivalent to averaging the component $R^2$ scores with weights proportional to their target variances. Wind, temperature, pressure, and relative humidity use m\,s$^{-1}$, K, Pa, and \%, respectively. TCC is evaluated as a fraction in $[0,1]$. The 1st Count column counts the best scores across the 16 variable-metric combinations in each table.

\subsection{Typhoon Track Evaluation}
\label{sec:supp_typhoon_evaluation}

The regional test split provides 33 initializations for Mawar, from 21:00 UTC on 25 May to 21:00 UTC on 2 June 2023, and 25 for Doksuri, from 09:00 UTC on 22 July to 09:00 UTC on 28 July 2023, sampled every six hours. Each initialization uses 12 hourly input frames to predict leads of 1-12 hours independently of other initializations. The maps display centers from the third output frame, while the error curves cover all 12 leads.

Figures~\ref{fig:typhoon_tracks} and~\ref{fig:supp_typhoon_mawar} use 10 and 16 consecutive initializations, respectively. The Doksuri window runs from 09:00 UTC on 22 July to 15:00 UTC on 24 July, and the Mawar window from 21:00 UTC on 28 May to 15:00 UTC on 1 June. At each lead, we average the forecast-to-reference distances over the corresponding initializations. The map uses their three-hour forecasts.

Typhoon centers are extracted from physical-scale MSL fields. ProtoCast uses its final MSL forecast after post-correction fusion and decoding. A JMA best-track position timestamped before forecast initialization guides a minimum-pressure search in the last historical field within a $5^\circ$ coordinate radius. Predicted and target centers are then located independently within $4^\circ$ of that historical center, with constrained quadratic refinement over a $3\times3$ neighborhood. All methods share this procedure at every lead. The reference centers are derived from the ERA5 target fields. Track error is the Haversine distance between forecast and reference centers, using an Earth radius of 6371.0088km.

\subsection{Model Size and Training Cost}
\label{sec:supp_training_cost}

\paragraph{Model size.} Both benchmarks use the complete ProtoCast framework. GlobalForecast uses $K=64$ prototypes per variable group and five input/output channels, while TyphoonTracking uses $K=32$ and four channels. These settings determine the sizes of the prototype-specific correction operators and variable encoders and decoders, yielding 4,545,645 and 4,146,269 trainable parameters, respectively. The shared spatiotemporal blocks and both cross-variable attention blocks retain the same configuration. Prototype keys and preprocessing statistics are stored as buffers and excluded from the parameter counts.

\begin{table}[H]
\centering
\footnotesize
\setlength{\tabcolsep}{3pt}
\renewcommand{\arraystretch}{1.08}
\caption{Model size and training time of the complete ProtoCast on both benchmarks. Training uses one NVIDIA H200 GPU, batch size 128, and BF16 precision.}
\label{tab:supp_full_training_cost}
\begin{tabular*}{\columnwidth}{@{\extracolsep{\fill}}lrrr@{}}
\toprule
Benchmark & Parameters & s/epoch & 50 epochs (h) \\
\midrule
GlobalForecast & 4,545,645 & 335.10 & 4.65 \\
TyphoonTracking & 4,146,269 & 296.48 & 4.12 \\
\bottomrule
\end{tabular*}
\end{table}

\paragraph{Training time.} Table~\ref{tab:supp_full_training_cost} summarizes the cost of 50-epoch training. The average epoch takes 335.10s on GlobalForecast and 296.48s on TyphoonTracking, totaling 4.65h and 4.12h, respectively. Each epoch includes scheduled prototype refitting, parameter optimization, validation, and checkpoint saving. Prototype refitting is performed only during training. These timings were recorded under the original run schedules.

\paragraph{Comparison with ablated models.} Table~\ref{tab:supp_ablation_training_cost} compares the trainable parameter counts and peak training memory of the complete model and the correction-free and plain-adapter variants on TyphoonTracking. The plain adapter uses unconditioned rank-256 correction branches. The complete model adds only 768 trainable parameters relative to this control, an increase of 0.019\%. Its measured peak training memory is 34.70GB.

\begin{table}[H]
\centering
\footnotesize
\setlength{\tabcolsep}{2pt}
\renewcommand{\arraystretch}{1.08}
\caption{Trainable parameters and peak training memory for the complete model and ablated variants on TyphoonTracking.}
\label{tab:supp_ablation_training_cost}
\begin{tabular*}{\columnwidth}{@{\extracolsep{\fill}}lrr@{}}
\toprule
Model & Parameters & Peak memory \\
 & & (GB) \\
\midrule
w/o correction & 3,738,461 & 33.60 \\
Plain adapter & 4,145,501 & 34.66 \\
ProtoCast & 4,146,269 & 34.70 \\
\bottomrule
\end{tabular*}
\end{table}

\subsection{Baseline Configurations and Result Sources}
\label{sec:supp_baselines}

\paragraph{Common training and evaluation protocol.} We retrained all ten baseline methods on GlobalForecast and TyphoonTracking using OpenSTL~\cite{tan2023openstl}. All methods used one GPU, BF16 mixed precision, and 50 training epochs, with batch sizes of 128 for training and 64 for evaluation. We used Adam with zero weight decay and an epoch-wise cosine learning-rate schedule, with no warm-up or gradient clipping. PredFormer~\cite{tang2026predformer} used the learning-rate settings from its official implementation, while the remaining baselines used those from the official implementation of Met$^2$Net~\cite{li2025met2net}. These settings were retained without modification. After training, we evaluated the checkpoint with the lowest validation loss on the test set. The methods used the same benchmark splits, historical inputs, forecast horizons, preprocessing, and metrics, as described in Section~\ref{sec:supp_data_evaluation}.

\paragraph{Network configurations.} Table~\ref{tab:supp_baseline_configs} summarizes the architectures used in both benchmarks. ConvNeXt, HorNet, and MogaNet are used as translator blocks within the SimVP forecasting framework, with four spatial blocks, eight translator blocks, a spatial hidden width of 32, and a translator hidden width of 256.

\begin{table}[t]
\centering
\small
\caption{Baseline architecture configurations shared across GlobalForecast and TyphoonTracking. Input and output dimensions follow the corresponding benchmark.}
\label{tab:supp_baseline_configs}
\begin{tabular}{@{}p{0.21\columnwidth}p{\dimexpr0.79\columnwidth-2\tabcolsep\relax}@{}}
\toprule
Method & Architecture configuration \\
\midrule
ConvLSTM & Four recurrent layers, 128 hidden channels per layer, kernel size 5, patch size 4. \\
PredRNN & Four recurrent layers, 128 hidden channels per layer, kernel size 5, patch size 4. \\
MAU & Four recurrent layers, 64 hidden channels per layer, temporal window 5, spatial reduction factor 4. \\
SimVP & Four spatial blocks and eight IncepU translator blocks, with hidden widths of 32 and 256. \\
ConvNeXt & SimVP framework with ConvNeXt translator blocks. \\
HorNet & SimVP framework with HorNet translator blocks. \\
MogaNet & SimVP framework with MogaNet translator blocks. \\
TAU & Four spatial blocks and eight translator blocks, hidden widths of 32 and 256, regularization coefficient 0.1. \\
Met$^2$Net & Variable-specific encoders and decoders, an eight-block cross-variable translator, and a momentum coefficient of 0.999. \\
PredFormer & Quadruplet-TSST with two quadruplet blocks, one layer per operation, embedding width 256, eight attention heads, and patch size 4. \\
\bottomrule
\end{tabular}
\end{table}

\paragraph{PredFormer.} For PredFormer~\cite{tang2026predformer}\footnote{\url{https://github.com/yyyujintang/PredFormer}}, we retained the network architecture reported in the original paper and used the Quadruplet-TSST variant with two quadruplet blocks. Each block contained one Gated Transformer layer for each temporal or spatial operation in the temporal-spatial-spatial-temporal sequence. The embedding dimension was 256, with eight attention heads, a head dimension of 32, a feed-forward expansion factor of two, and a patch size of $4\times4$. Dropout and attention dropout were both 0.1, and the configured stochastic-depth rate was 0.25. The model jointly encoded and predicted all five channels on GlobalForecast and all four channels on TyphoonTracking. The input and output dimensions matched each benchmark, while the backbone configuration was retained.

\paragraph{Met$^2$Net.} For Met$^2$Net~\cite{li2025met2net}\footnote{\url{https://github.com/ShremG/Met2Net}}, we retained the network architecture reported in the original paper, together with its two-stage objective within each training iteration. The first stage optimized the variable-specific online encoders and decoders using a forecasting MSE through the momentum translator. The second stage optimized the online translator using a latent MSE, with input and target representations supplied by the momentum encoders. The two losses were summed with unit weights, and the momentum coefficient was 0.999. Inference used the online encoders, translator, and decoders. Each encoder and decoder used four spatial blocks with a hidden width of 32 and a kernel size of three. The translator used eight cross-variable blocks with a hidden width of 256 and eight attention heads. GlobalForecast used the variable groups (U10, V10), T2M, TCC, and R, while TyphoonTracking used separate groups for MSL, U10, V10, and T2M.

\section{Additional Experiments}
\label{sec:supp_experiments}
\setcounter{topnumber}{1}
\makeatletter
\setlength{\@fptop}{0pt}
\setlength{\@dblfptop}{0pt}
\makeatother

\subsection{Core Ablation Settings and Analysis}
\label{sec:supp_ablation_protocol}

We evaluate all ablations on TyphoonTracking using the ProtoCast training configuration reported for Table~\ref{tab:era5_main}, except for the component or hyperparameter under study. The data splits, 50-epoch training schedule, validation-based checkpoint selection, and evaluation protocol remain unchanged. Each run uses batch size 128, Adam with learning rate $2\times10^{-3}$ and $\beta=(0.9,0.995)$, cosine decay, and BF16 mixed precision. Evaluation covers all 8,737 test windows after inverse normalization. We aggregate errors over windows, forecast leads, and grid elements as in Section~\ref{sec:experimental_setup}.

The reference configuration uses $K=32$ prototypes per group, correction rank $r=16$, matching temperature $\tau=0.2$, and matching dimension $d_q=64$. All variants are trained from scratch. Table~\ref{tab:ablation_core} isolates the roles of local correction, prototype weighting, and cross-variable interactions by changing the specified component while retaining the remaining configuration.

\paragraph{w/o correction.} We remove the group-specific correction branches and their prototype banks while retaining both cross-variable blocks and the forecasting modules. The uncorrected group features therefore pass through post-correction fusion before the final spatial block and decoding. This variant has higher MSEs than the full model for all four variables, showing that the local correction branches improve prediction beyond the retained forecasting and cross-variable components.

\paragraph{Plain adapter.} We replace each prototype-conditioned branch with an unconditioned residual adapter of rank 256 and retain both cross-variable blocks. The adapters are trained jointly with the forecasting modules but do not retrieve or combine prototype-specific operators. Their 407,040 branch parameters closely match the 407,808 parameters in the full correction mechanism. The full model has lower MSEs for all four variables, supporting the benefit of conditioning corrections on prototype matches at a similar parameter budget.

\paragraph{Uniform weights.} We retain the prototype-specific correction operators but assign every operator the fixed weight $1/K$ at every patch. The operators remain trainable, while query-dependent matching and prototype refitting are omitted. Both cross-variable blocks remain unchanged. MSE increases for all four variables, indicating that a fixed average of the operators is less effective than selecting their mixture according to each local state.

\paragraph{w/o pre-attention.} We replace pre-correction cross-variable attention with an identity mapping. Each group still constructs its matching descriptor and applies its correction, but neither the forecast context used for matching nor the correction input receives information from the other groups at this stage. Post-correction fusion remains active. The MSE of MSL rises by 7.44\%, and errors also increase for U10, V10, and T2M. Exchanging information before correction thus helps the model account for related variables when selecting and applying local updates.

\paragraph{w/o post-fusion.} We remove post-correction cross-variable attention and pass the corrected group features directly to the final spatial block and decoding. Pre-correction attention, prototype matching, and local correction remain unchanged. The MSE of MSL rises by 6.27\%, with higher errors for both wind components and temperature. The result shows the benefit of exchanging the corrected features across groups after their local updates.

\paragraph{Fixed gate ($\gamma=1$).} We retain trainable post-correction attention but fix its gate to one throughout training. This replaces the learnable, zero-initialized gate and applies the full fusion update from the start. Relative to the full model, the MSEs of MSL, U10, and V10 increase, while the MSE of T2M decreases from 0.5529 to 0.5491. The learned fusion strength benefits pressure and wind, while the fixed gate slightly improves T2M prediction.

\paragraph{Full model (Ours).} The full model combines entmax-based prototype matching, prototype-conditioned low-rank corrections, pre-correction cross-variable attention, and post-correction fusion with a learnable gate initialized to zero. Prototype banks are refitted with slot tracking before each epoch. It achieves the lowest MSEs for MSL, U10, and V10 and the second-lowest MSE for T2M among the seven configurations. The comparisons jointly support adapting local corrections to prototype matches and exchanging information across variables before and after those corrections. Table~\ref{tab:ablation_complete} reports all four metrics for these variants.

\subsection{Prototype Matching and Refitting}
\label{sec:supp_ablation_routing}

\begin{table}[t]
\centering
\footnotesize
\setlength{\tabcolsep}{2pt}
\renewcommand{\arraystretch}{1.08}
\caption{Matching and refitting ablations on TyphoonTracking, measured by MSE. Ours uses entmax and refits before every epoch. \textcolor{cvprblue}{\textbf{Best}} and \underline{second-best} values are marked.}
\label{tab:ablation_routing}
\begin{tabular*}{\columnwidth}{@{\extracolsep{\fill}}lrrrr@{}}
\toprule
Variant & MSL & U10 & V10 & T2M \\
\midrule
Uniform weights & 6124.51 & 1.0138 & 1.1481 & 0.5605 \\
Softmax & 5973.97 & \underline{0.9903} & \underline{1.1316} & 0.5543 \\
Initial refit only & 6145.60 & 0.9972 & 1.1405 & 0.5582 \\
Refit for 40 epochs & \underline{5954.62} & 0.9991 & 1.1537 & \textcolor{cvprblue}{\textbf{0.5528}} \\
\textbf{Ours} & \textcolor{cvprblue}{\textbf{5888.81}} & \textcolor{cvprblue}{\textbf{0.9886}} & \textcolor{cvprblue}{\textbf{1.1292}} & \underline{0.5529} \\
\bottomrule
\end{tabular*}
\end{table}

Table~\ref{tab:ablation_routing} examines the matching rule and the update schedule for the prototype references. Replacing entmax with softmax increases MSE for all four variables, while uniform weights produce larger errors. Matching-dependent mixtures improve on uniform combinations, and sparse matching further improves on dense softmax weights.

Fitting the banks only before training increases all four MSEs compared with refitting before every epoch. Stopping refitting after epoch 40 also increases MSL and wind errors, while the MSE of T2M remains close to the full model at 0.5528 versus 0.5529. Updating the references throughout training therefore benefits pressure and wind forecasts as the learned representations evolve.

\subsection{Matching Descriptor Components}
\label{sec:supp_descriptor_components}

We test whether historical change and forecast context improve prototype matching beyond the mean historical state. Removing pre-correction attention changes both the matching context and the features receiving corrections. To isolate the contribution of each descriptor term, we separately set the historical-change term $\delta u_n^{(g)}$ or the forecast-context term $\bar{h}_n^{(g)}$ in Equation~\eqref{eq:method_descriptor} to zero. Both variants retain the historical mean, the remaining descriptor term, both cross-variable attention blocks, and the features entering the correction branches.

The variants follow the TyphoonTracking protocol in Table~\ref{tab:era5_main} and Section~\ref{sec:supp_ablation_protocol}, with the same data splits, 50-epoch training schedule, validation-based checkpoint selection, and full-test evaluation. Both are trained from scratch. Each variant refits its own descriptor standardization, PCA projection, and prototype banks from training samples before every epoch. Zeroing the selected component preserves the $3D=576$ descriptor size, $d_q=64$ matching dimension, $K=32$ operators per group, and correction rank $r=16$.

\begin{table}[t]
\centering
\footnotesize
\setlength{\tabcolsep}{2pt}
\renewcommand{\arraystretch}{1.18}
\caption{Matching-descriptor ablations on TyphoonTracking. Parentheses give percentage changes from Ours. The final row reports mean normalized MSE, scaled by $10^4$. Lower is better. \textcolor{cvprblue}{\textbf{Best}} and \underline{second-best} values are marked.}
\label{tab:supp_descriptor_components}
\begin{tabular*}{\columnwidth}{@{\extracolsep{\fill}}lrrr@{}}
\toprule
Variable & Ours & \shortstack[r]{No history\\change} & \shortstack[r]{No forecast\\context} \\
\midrule
MSL (Pa$^2$) & \textcolor{cvprblue}{\textbf{5888.8075}} & \shortstack[r]{\underline{5993.1686}\\($+1.77\%$)} & \shortstack[r]{6032.4644\\($+2.44\%$)} \\
U10 (m$^2$\,s$^{-2}$) & \underline{0.9886} & \shortstack[r]{1.0067\\($+1.82\%$)} & \shortstack[r]{\textcolor{cvprblue}{\textbf{0.9842}}\\($-0.45\%$)} \\
V10 (m$^2$\,s$^{-2}$) & \textcolor{cvprblue}{\textbf{1.1292}} & \shortstack[r]{1.1745\\($+4.01\%$)} & \shortstack[r]{\underline{1.1403}\\($+0.98\%$)} \\
T2M (K$^2$) & \underline{0.5529} & \shortstack[r]{0.5634\\($+1.90\%$)} & \shortstack[r]{\textcolor{cvprblue}{\textbf{0.5483}}\\($-0.83\%$)} \\
\midrule
\shortstack[l]{Mean normalized\\MSE ($\times10^{-4}$)} & \textcolor{cvprblue}{\textbf{1.816433}} & \shortstack[r]{1.866009\\($+2.73\%$)} & \shortstack[r]{\underline{1.821510}\\($+0.28\%$)} \\
\bottomrule
\end{tabular*}
\end{table}

Table~\ref{tab:supp_descriptor_components} reports physical-scale MSE for each variable and an aggregate in normalized space. For the aggregate, we first compute each variable's MSE over all 8,737 test windows, 12 forecast leads, and grid elements using its training-set min-max normalization. We then take the arithmetic mean of the four normalized MSEs, giving equal weight to MSL, U10, V10, and T2M.

Removing historical change increases the MSE of every variable, with the largest increase of $4.01\%$ for V10. Mean normalized MSE rises from $1.816433\times10^{-4}$ to $1.866009\times10^{-4}$, an increase of $2.73\%$. Explicit temporal-change information therefore improves matching-based correction across pressure, wind, and temperature.

Removing forecast context increases the MSEs of MSL and V10 by $2.44\%$ and $0.98\%$, while reducing those of U10 and T2M by $0.45\%$ and $0.83\%$. Mean normalized MSE increases by $0.28\%$ to $1.821510\times10^{-4}$. Historical change provides the larger and more uniform benefit, while forecast context adds a smaller, variable-dependent contribution. The complete descriptor achieves the lowest mean normalized MSE.

\subsection{Learned Routing and Slot Tracking}
\label{sec:supp_router_tracking}

We examine whether explicit prototype references improve on learned mixture weights and whether tracking prototype slots benefits training. A learned router tests whether the same matching information can select correction operators effectively without a prototype bank. Refitting without slot tracking tests whether updating the references is sufficient when their correspondence with learned operators is not carried across epochs.

Both controls follow the TyphoonTracking experiment in Table~\ref{tab:era5_main} and the protocol in Section~\ref{sec:supp_ablation_protocol}. They retain the forecasting backbone, both cross-variable attention blocks, and $K=32$ rank-$16$ correction operators per group. Each is trained from scratch for 50 epochs with the same data splits, optimization settings, validation-based checkpoint selection, and full-test evaluation as Ours.

\paragraph{Learned router.} For each group, we apply LayerNorm and a trainable linear map to the same $3D=576$ descriptor in Equation~\eqref{eq:method_descriptor}, producing $K$ mixture logits. The descriptor remains detached, while the router learns from the forecasting loss. Temperature $\tau=0.2$ and entmax normalization are retained. This replaces descriptor standardization, PCA, and cosine matching to fitted prototypes with a learned routing function. No prototype refitting is performed. The four routers introduce 78,464 trainable parameters, while the correction operators remain unchanged.

\paragraph{No slot tracking.} We retain descriptor standardization, PCA, spherical $k$-means, and refitting before every epoch. Each refit uses the same fixed training samples, matching dimension $d_q=64$, and 25 clustering iterations. Instead of initializing centers from previous sample assignments as in Equation~\eqref{eq:method_refitting}, every refit uses the sample-based initialization of the first fit. The learned correction operators keep their slot indices and are neither reinitialized nor permuted. This removes membership-based continuity during training while retaining repeated prototype updates.

\begin{table}[t]
\centering
\footnotesize
\setlength{\tabcolsep}{2pt}
\renewcommand{\arraystretch}{1.18}
\caption{Routing and slot-tracking controls on TyphoonTracking, measured by MSE. Parentheses give percentage changes from Ours. The final row reports mean normalized MSE, scaled by $10^4$. Lower is better. \textcolor{cvprblue}{\textbf{Best}} and \underline{second-best} values are marked.}
\label{tab:supp_router_tracking}
\begin{tabular*}{\columnwidth}{@{\extracolsep{\fill}}lrrr@{}}
\toprule
Variable & Ours & \shortstack[r]{Learned\\router} & \shortstack[r]{No slot\\tracking} \\
\midrule
MSL (Pa$^2$) & \textcolor{cvprblue}{\textbf{5888.8075}} & \shortstack[r]{6130.3865\\($+4.10\%$)} & \shortstack[r]{\underline{6032.5248}\\($+2.44\%$)} \\
U10 (m$^2$\,s$^{-2}$) & \textcolor{cvprblue}{\textbf{0.9886}} & \shortstack[r]{1.0047\\($+1.62\%$)} & \shortstack[r]{\underline{0.9890}\\($+0.04\%$)} \\
V10 (m$^2$\,s$^{-2}$) & \textcolor{cvprblue}{\textbf{1.1292}} & \shortstack[r]{1.1480\\($+1.67\%$)} & \shortstack[r]{\underline{1.1435}\\($+1.27\%$)} \\
T2M (K$^2$) & \textcolor{cvprblue}{\textbf{0.5529}} & \shortstack[r]{\underline{0.5580}\\($+0.93\%$)} & \shortstack[r]{0.5588\\($+1.06\%$)} \\
\midrule
\shortstack[l]{Mean normalized\\MSE ($\times10^{-4}$)} & \textcolor{cvprblue}{\textbf{1.816433}} & \shortstack[r]{1.847644\\($+1.72\%$)} & \shortstack[r]{\underline{1.831505}\\($+0.83\%$)} \\
\bottomrule
\end{tabular*}
\end{table}

Table~\ref{tab:supp_router_tracking} reports the MSE of each variable after inverse normalization. The aggregate first computes each variable's MSE over all 8,737 test windows, 12 forecast leads, and grid elements under its training-set min-max normalization, then takes the arithmetic mean across the four variables, as in Section~\ref{sec:supp_descriptor_components}.

The learned router increases the MSEs of MSL, U10, V10, and T2M by $4.10\%$, $1.62\%$, $1.67\%$, and $0.93\%$, respectively. Mean normalized MSE rises from $1.816433\times10^{-4}$ to $1.847644\times10^{-4}$, an increase of $1.72\%$. Explicit prototype matching therefore improves pressure, wind, and temperature forecasts over learned linear routing of the same descriptor.

Removing slot tracking increases the MSEs of MSL, U10, V10, and T2M by $2.44\%$, $0.04\%$, $1.27\%$, and $1.06\%$. Mean normalized MSE increases by $0.83\%$ to $1.831505\times10^{-4}$. The larger changes in MSL, V10, and T2M show that preserving prototype-operator correspondence across refits benefits these forecasts, while U10 changes little.

\subsection{Prototype and Correction Capacity}
\label{sec:supp_ablation_capacity}

\begin{table}[t]
\centering
\footnotesize
\setlength{\tabcolsep}{2pt}
\renewcommand{\arraystretch}{1.08}
\caption{Prototype-bank size and correction rank on TyphoonTracking. Entries are MSEs. \textcolor{cvprblue}{\textbf{Best}} and \underline{second-best} values are marked. Rankings are computed within each panel.}
\label{tab:ablation_capacity}
\par\noindent\textbf{(a) Prototypes per group}\par\smallskip
\begin{tabular*}{\columnwidth}{@{\extracolsep{\fill}}lrrrr@{}}
\toprule
Variant & MSL & U10 & V10 & T2M \\
\midrule
$K=8$ & 6059.33 & 0.9973 & 1.1522 & 0.5545 \\
$K=16$ & 6002.10 & 0.9937 & 1.1528 & 0.5538 \\
\textbf{$K=32$ (Ours)} & \textcolor{cvprblue}{\textbf{5888.81}} & \underline{0.9886} & \textcolor{cvprblue}{\textbf{1.1292}} & \underline{0.5529} \\
$K=64$ & \underline{5986.45} & \textcolor{cvprblue}{\textbf{0.9855}} & \underline{1.1301} & \textcolor{cvprblue}{\textbf{0.5467}} \\
\bottomrule
\end{tabular*}
\vspace{3pt}
\par\noindent\textbf{(b) Correction rank}\par\smallskip
\begin{tabular*}{\columnwidth}{@{\extracolsep{\fill}}lrrrr@{}}
\toprule
Variant & MSL & U10 & V10 & T2M \\
\midrule
$r=4$ & 6175.58 & 1.0018 & \underline{1.1452} & 0.5621 \\
$r=8$ & \underline{5960.50} & \underline{0.9888} & 1.1487 & \underline{0.5614} \\
\textbf{$r=16$ (Ours)} & \textcolor{cvprblue}{\textbf{5888.81}} & \textcolor{cvprblue}{\textbf{0.9886}} & \textcolor{cvprblue}{\textbf{1.1292}} & \textcolor{cvprblue}{\textbf{0.5529}} \\
$r=32$ & 6033.49 & 0.9908 & 1.1453 & 0.5615 \\
\bottomrule
\end{tabular*}
\end{table}

Table~\ref{tab:ablation_capacity} varies the number of prototypes and the correction rank. With $K=32$, the model achieves the lowest MSEs of MSL and V10. Increasing $K$ to 64 improves U10 and T2M but does not further reduce pressure error. A larger bank therefore changes which variables benefit most, rather than improving all four variables. For the correction rank, $r=16$ gives the lowest MSE for all four variables compared with $r\in\{4,8,32\}$. Increasing the rank beyond this setting does not improve forecast accuracy.

\subsection{Matching Temperature and Dimension}
\label{sec:supp_ablation_retrieval}

\begin{table}[!b]
\centering
\footnotesize
\setlength{\tabcolsep}{2pt}
\renewcommand{\arraystretch}{1.08}
\caption{Matching temperature and dimension on TyphoonTracking. Entries are MSEs. \textcolor{cvprblue}{\textbf{Best}} and \underline{second-best} values are marked. Rankings are computed within each panel.}
\label{tab:ablation_retrieval}
\par\noindent\textbf{(a) Matching temperature}\par\smallskip
\begin{tabular*}{\columnwidth}{@{\extracolsep{\fill}}lrrrr@{}}
\toprule
Variant & MSL & U10 & V10 & T2M \\
\midrule
$\tau=0.1$ & \underline{6021.44} & 0.9896 & 1.1468 & \textcolor{cvprblue}{\textbf{0.5465}} \\
\textbf{$\tau=0.2$ (Ours)} & \textcolor{cvprblue}{\textbf{5888.81}} & \underline{0.9886} & \textcolor{cvprblue}{\textbf{1.1292}} & \underline{0.5529} \\
$\tau=0.4$ & 6029.71 & \textcolor{cvprblue}{\textbf{0.9826}} & \underline{1.1370} & 0.5572 \\
\bottomrule
\end{tabular*}
\vspace{3pt}
\par\noindent\textbf{(b) Matching dimension}\par\smallskip
\begin{tabular*}{\columnwidth}{@{\extracolsep{\fill}}lrrrr@{}}
\toprule
Variant & MSL & U10 & V10 & T2M \\
\midrule
$d_q=16$ & \underline{5976.02} & 0.9983 & 1.1448 & \textcolor{cvprblue}{\textbf{0.5492}} \\
$d_q=32$ & 5989.47 & \underline{0.9895} & 1.1490 & 0.5528 \\
\textbf{$d_q=64$ (Ours)} & \textcolor{cvprblue}{\textbf{5888.81}} & \textcolor{cvprblue}{\textbf{0.9886}} & \textcolor{cvprblue}{\textbf{1.1292}} & 0.5529 \\
$d_q=128$ & 5994.06 & 0.9965 & \underline{1.1382} & \underline{0.5525} \\
\bottomrule
\end{tabular*}
\end{table}

Table~\ref{tab:ablation_retrieval} examines the matching temperature and dimension. The reference temperature $\tau=0.2$ gives the lowest MSEs of MSL and V10, whereas $\tau=0.4$ improves U10 and $\tau=0.1$ improves T2M. Similarly, $d_q=64$ yields the lowest pressure and wind errors, while $d_q=16$ yields the lowest T2M error. These results show that matching settings affect meteorological variables differently. The reference configuration favors pressure and wind accuracy without requiring the largest prototype bank or matching dimension.

\subsection{Complete Ablation Results}
\label{sec:supp_ablation_complete}

Table~\ref{tab:ablation_complete} reports MSE, MAE, RMSE, and $R^2$ for all 21 configurations and four meteorological variables. The full model improves all 16 metrics over the correction-free, plain-adapter, uniform-weight, softmax, initial-refit-only, and attention-removal variants. The parameter sweeps reveal additional variable-specific trade-offs, including lower T2M errors with $K=64$ and $\tau=0.1$.

\begin{table*}[!t]
\centering
\caption{Complete ablation results on TyphoonTracking. Lower errors and higher $R^2$ indicate better performance. The \textcolor{cvprblue}{\textbf{best}} and \underline{second-best} values are highlighted. The 1st Count column records the number of best scores.}
\label{tab:ablation_complete}
\small
\setlength{\tabcolsep}{2pt}
\renewcommand{\arraystretch}{1.12}
\resizebox{\textwidth}{!}{%
\begin{tabular}{l*{16}{r}c}
\toprule
\multirow[c]{2}{*}{Variant} & \multicolumn{4}{c}{MSL} & \multicolumn{4}{c}{U10} & \multicolumn{4}{c}{V10} & \multicolumn{4}{c}{T2M} & \multirow[c]{2}{*}{\shortstack{1st\\Count}} \\
\cmidrule(lr){2-5}\cmidrule(lr){6-9}\cmidrule(lr){10-13}\cmidrule(lr){14-17}
 & MSE$\downarrow$ & MAE$\downarrow$ & RMSE$\downarrow$ & $R^2\!\uparrow$ & MSE$\downarrow$ & MAE$\downarrow$ & RMSE$\downarrow$ & $R^2\!\uparrow$ & MSE$\downarrow$ & MAE$\downarrow$ & RMSE$\downarrow$ & $R^2\!\uparrow$ & MSE$\downarrow$ & MAE$\downarrow$ & RMSE$\downarrow$ & $R^2\!\uparrow$ & \\
\midrule
w/o correction & 6074.21 & 53.7532 & 77.9372 & 0.9895 & 0.9993 & 0.6842 & 0.9996 & 0.9474 & 1.1361 & 0.7280 & 1.0659 & 0.9383 & 0.5560 & 0.4811 & 0.7457 & 0.9943 & 0 \\
Plain adapter & 6117.54 & 53.9834 & 78.2147 & 0.9894 & 0.9957 & 0.6813 & 0.9979 & 0.9476 & 1.1414 & 0.7289 & 1.0684 & 0.9380 & 0.5548 & 0.4814 & 0.7449 & 0.9943 & 0 \\
Uniform weights & 6124.51 & 54.1195 & 78.2592 & 0.9894 & 1.0138 & 0.6899 & 1.0069 & 0.9466 & 1.1481 & 0.7316 & 1.0715 & 0.9376 & 0.5605 & 0.4834 & 0.7487 & 0.9942 & 0 \\
w/o pre-attention & 6326.96 & 54.8428 & 79.5422 & 0.9891 & 1.0145 & 0.6900 & 1.0072 & 0.9466 & 1.1585 & 0.7362 & 1.0763 & 0.9371 & 0.5758 & 0.4897 & 0.7588 & 0.9941 & 0 \\
w/o post-fusion & 6258.18 & 54.3255 & 79.1087 & 0.9892 & 1.0101 & 0.6870 & 1.0050 & 0.9468 & 1.1713 & 0.7370 & 1.0823 & 0.9364 & 0.5626 & 0.4834 & 0.7500 & 0.9942 & 0 \\
Fixed gate ($\gamma=1$) & 5993.01 & 53.6708 & 77.4145 & 0.9897 & 1.0006 & 0.6827 & 1.0003 & 0.9473 & 1.1405 & 0.7283 & 1.0679 & 0.9381 & 0.5491 & 0.4790 & 0.7410 & 0.9944 & 0 \\
Softmax & 5973.97 & 53.4097 & 77.2915 & 0.9897 & 0.9903 & 0.6820 & 0.9952 & 0.9479 & 1.1316 & 0.7267 & 1.0638 & 0.9385 & 0.5543 & 0.4814 & 0.7445 & 0.9943 & 0 \\
Initial refit only & 6145.60 & 53.8139 & 78.3939 & 0.9894 & 0.9972 & 0.6826 & 0.9986 & 0.9475 & 1.1405 & 0.7283 & 1.0679 & 0.9381 & 0.5582 & 0.4814 & 0.7471 & 0.9943 & 0 \\
Refit for 40 epochs & \underline{5954.62} & \underline{53.2468} & \underline{77.1662} & \underline{0.9897} & 0.9991 & 0.6823 & 0.9996 & 0.9474 & 1.1537 & 0.7326 & 1.0741 & 0.9373 & 0.5528 & 0.4808 & 0.7435 & 0.9943 & 0 \\
$K=8$ & 6059.33 & 53.6445 & 77.8417 & 0.9895 & 0.9973 & 0.6828 & 0.9986 & 0.9475 & 1.1522 & 0.7331 & 1.0734 & 0.9374 & 0.5545 & 0.4812 & 0.7446 & 0.9943 & 0 \\
$K=16$ & 6002.10 & 53.5075 & 77.4732 & 0.9896 & 0.9937 & 0.6808 & 0.9968 & 0.9477 & 1.1528 & 0.7330 & 1.0737 & 0.9374 & 0.5538 & 0.4810 & 0.7442 & 0.9943 & 0 \\
$K=64$ & 5986.45 & 53.3235 & 77.3721 & 0.9897 & \underline{0.9855} & \textcolor{cvprblue}{\textbf{0.6778}} & \underline{0.9927} & \underline{0.9481} & \underline{1.1301} & \textcolor{cvprblue}{\textbf{0.7242}} & \underline{1.0630} & \underline{0.9386} & \underline{0.5467} & \textcolor{cvprblue}{\textbf{0.4779}} & \underline{0.7394} & \underline{0.9944} & 3 \\
$r=4$ & 6175.58 & 54.2768 & 78.5849 & 0.9893 & 1.0018 & 0.6853 & 1.0009 & 0.9473 & 1.1452 & 0.7309 & 1.0701 & 0.9378 & 0.5621 & 0.4843 & 0.7497 & 0.9942 & 0 \\
$r=8$ & 5960.50 & 53.4622 & 77.2043 & 0.9897 & 0.9888 & 0.6800 & 0.9944 & 0.9479 & 1.1487 & 0.7321 & 1.0718 & 0.9376 & 0.5614 & 0.4834 & 0.7492 & 0.9942 & 0 \\
$r=32$ & 6033.49 & 53.6051 & 77.6756 & 0.9896 & 0.9908 & 0.6799 & 0.9954 & 0.9478 & 1.1453 & 0.7301 & 1.0702 & 0.9378 & 0.5615 & 0.4838 & 0.7493 & 0.9942 & 0 \\
$\tau=0.1$ & 6021.44 & 53.6372 & 77.5979 & 0.9896 & 0.9896 & 0.6806 & 0.9948 & 0.9479 & 1.1468 & 0.7297 & 1.0709 & 0.9377 & \textcolor{cvprblue}{\textbf{0.5465}} & \underline{0.4783} & \textcolor{cvprblue}{\textbf{0.7393}} & \textcolor{cvprblue}{\textbf{0.9944}} & 3 \\
$\tau=0.4$ & 6029.71 & 53.6418 & 77.6512 & 0.9896 & \textcolor{cvprblue}{\textbf{0.9826}} & \underline{0.6784} & \textcolor{cvprblue}{\textbf{0.9913}} & \textcolor{cvprblue}{\textbf{0.9483}} & 1.1370 & 0.7279 & 1.0663 & 0.9382 & 0.5572 & 0.4821 & 0.7464 & 0.9943 & 3 \\
$d_q=16$ & 5976.02 & 53.3279 & 77.3047 & 0.9897 & 0.9983 & 0.6838 & 0.9992 & 0.9474 & 1.1448 & 0.7301 & 1.0699 & 0.9378 & 0.5492 & 0.4789 & 0.7411 & 0.9944 & 0 \\
$d_q=32$ & 5989.47 & 53.4819 & 77.3917 & 0.9897 & 0.9895 & 0.6798 & 0.9947 & 0.9479 & 1.1490 & 0.7315 & 1.0719 & 0.9376 & 0.5528 & 0.4804 & 0.7435 & 0.9943 & 0 \\
$d_q=128$ & 5994.06 & 53.4231 & 77.4213 & 0.9897 & 0.9965 & 0.6816 & 0.9982 & 0.9475 & 1.1382 & 0.7271 & 1.0669 & 0.9382 & 0.5525 & 0.4804 & 0.7433 & 0.9943 & 0 \\
\midrule
\textbf{Ours} & \textcolor{cvprblue}{\textbf{5888.81}} & \textcolor{cvprblue}{\textbf{53.1169}} & \textcolor{cvprblue}{\textbf{76.7386}} & \textcolor{cvprblue}{\textbf{0.9898}} & 0.9886 & 0.6793 & 0.9943 & 0.9479 & \textcolor{cvprblue}{\textbf{1.1292}} & \underline{0.7248} & \textcolor{cvprblue}{\textbf{1.0626}} & \textcolor{cvprblue}{\textbf{0.9387}} & 0.5529 & 0.4804 & 0.7436 & 0.9943 & \textcolor{cvprblue}{\textbf{7}} \\
\bottomrule
\end{tabular}%
}
\end{table*}

\subsection{Extended Prototype Contribution Results}
\label{sec:supp_prototype_results}

\begin{figure*}[!tp]
    \centering
    \includegraphics[width=\textwidth]{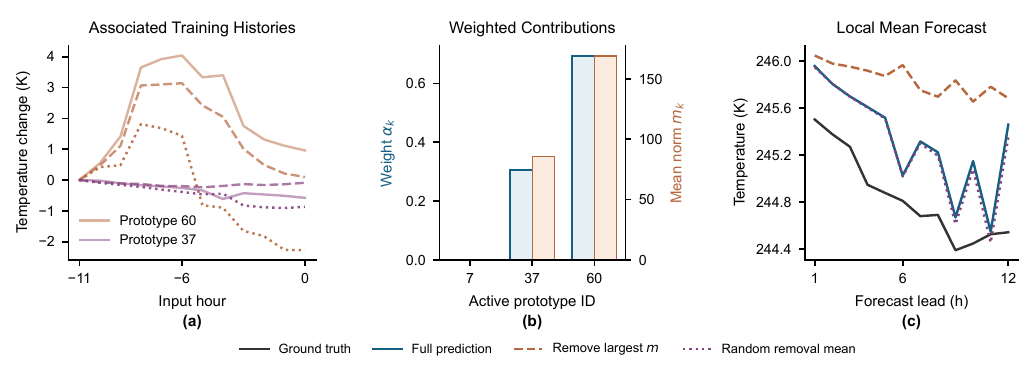}
    \caption{Interpreting a local temperature correction on GlobalForecast. (a) Three training histories associated with each of two active prototypes, shown relative to the first observation. (b) Matching weights $\alpha$ and mean weighted-residual norms $m$. (c) Patch-mean temperature forecasts with the full model, the largest-$m$ contribution removed, or a random other active contribution removed, alongside ground truth. Random-deletion forecasts average three draws. Matching weights stay fixed in both removal tests.}
    \label{fig:prototype_contributions}
\end{figure*}

\paragraph{GlobalForecast temperature.} Figure~\ref{fig:prototype_contributions} connects prototype-associated training histories to local corrections and forecast responses on GlobalForecast. The matching weights identify the active references, and their weighted residuals quantify how each reference contributes to the feature update. The contribution magnitude $m$ is defined in Equation~\eqref{eq:supp_contribution_magnitude}. Removing the largest-$m$ term changes the temperature forecast more than removing another active term in the illustrated case, making the link between a retrieved pattern and its prediction effect explicit. Section~\ref{sec:supp_prototype_cases} presents additional examples linking prototype matches and training histories to forecast accuracy across all four variable groups.

Across 96 test windows, deleting the largest-$m$ temperature contribution increases local MAE by 0.1361K, compared with 0.0072K for a randomly selected other active contribution. Error increases in 83 of the 96 cases. The mean absolute prediction response is 0.4480K versus 0.0674K, a 6.65-fold difference. Deleting the largest-$m$ term therefore produces a larger mean forecast response and error increase than random deletion. The decomposition thus identifies useful local corrections and makes their role in forecasting more interpretable. Matching weights and other contributions remain fixed during deletion.

\begin{table}[t]
\centering
\footnotesize
\setlength{\tabcolsep}{2pt}
\caption{Local error changes after prototype deletion. Largest-$m$ and random report mean $\Delta\mathrm{MAE}$ in each variable's physical units. The last column counts cases with increased error after largest-$m$ deletion. R uses percentage points and TCC a fraction.}
\label{tab:supp_prototype_summary}
\begin{tabular*}{\columnwidth}{@{\extracolsep{\fill}}lrrrr@{}}
\toprule
Group & $N$ & Largest-$m$ & Random & Increased \\
\midrule
Global T2M & 96 & +0.13608 & +0.00723 & 83/96 \\
Global R & 96 & +0.44656 & +0.00950 & 80/96 \\
Global UV10 & 96 & +0.10028 & +0.00591 & 87/96 \\
Global TCC & 96 & +0.00137 & -0.00003 & 61/96 \\
Regional T2M & 45 & +0.03150 & -0.00193 & 31/45 \\
\bottomrule
\end{tabular*}
\end{table}

\begin{table*}[!tp]
\centering
\footnotesize
\setlength{\tabcolsep}{4pt}
\caption{Cross-variable responses to local prototype deletion on GlobalForecast, averaged over the same 96 cases for each condition. $D$ is the mean absolute forecast response and $\Delta$MAE is the error after deletion minus the full-model error. T2M, R, UV10, and TCC use K, percentage points, m\,s$^{-1}$, and a fraction, respectively.}
\label{tab:supp_cross_global}
\begin{tabular*}{\textwidth}{@{\extracolsep{\fill}}llrrrrrrrr@{}}
\toprule
Modified group & Removed term & \multicolumn{2}{c}{T2M} & \multicolumn{2}{c}{R} & \multicolumn{2}{c}{UV10} & \multicolumn{2}{c}{TCC} \\
\cmidrule(lr){3-4}\cmidrule(lr){5-6}\cmidrule(lr){7-8}\cmidrule(lr){9-10}
 & & $D$ & $\Delta$MAE & $D$ & $\Delta$MAE & $D$ & $\Delta$MAE & $D$ & $\Delta$MAE \\
\midrule
T2M & Largest-$m$ & 0.448038 & +0.136081 & 0.229989 & +0.018922 & 0.074353 & +0.006112 & 0.007780 & +0.000114 \\
T2M & Random & 0.067408 & +0.007234 & 0.039725 & -0.001520 & 0.011510 & +0.000700 & 0.001279 & -0.000026 \\
\addlinespace
R & Largest-$m$ & 0.073885 & +0.008201 & 1.296052 & +0.446564 & 0.037506 & +0.002874 & 0.005317 & +0.000123 \\
R & Random & 0.008442 & +0.000006 & 0.140833 & +0.009496 & 0.005244 & +0.000098 & 0.000716 & +0.000023 \\
\addlinespace
UV10 & Largest-$m$ & 0.047069 & +0.002947 & 0.223304 & +0.015973 & 0.328560 & +0.100281 & 0.007632 & +0.000694 \\
UV10 & Random & 0.007951 & -0.000709 & 0.042334 & +0.000328 & 0.054729 & +0.005906 & 0.001229 & -0.000083 \\
\addlinespace
TCC & Largest-$m$ & 0.015303 & -0.000575 & 0.047329 & -0.003739 & 0.005889 & -0.000205 & 0.017231 & +0.001368 \\
TCC & Random & 0.002290 & -0.000158 & 0.007454 & -0.000224 & 0.000981 & +0.000038 & 0.002751 & -0.000033 \\
\bottomrule
\end{tabular*}
\end{table*}

\begin{table*}[!tp]
\centering
\footnotesize
\setlength{\tabcolsep}{4pt}
\caption{Cross-variable responses to temperature-prototype deletion on TyphoonTracking over 45 paired cases. Definitions follow Table~\ref{tab:supp_cross_global}. MSL uses Pa, U10 and V10 use m\,s$^{-1}$, and T2M uses K.}
\label{tab:supp_cross_regional}
\begin{tabular*}{\textwidth}{@{\extracolsep{\fill}}llrrrrrrrr@{}}
\toprule
Modified group & Removed term & \multicolumn{2}{c}{MSL} & \multicolumn{2}{c}{U10} & \multicolumn{2}{c}{V10} & \multicolumn{2}{c}{T2M} \\
\cmidrule(lr){3-4}\cmidrule(lr){5-6}\cmidrule(lr){7-8}\cmidrule(lr){9-10}
 & & $D$ & $\Delta$MAE & $D$ & $\Delta$MAE & $D$ & $\Delta$MAE & $D$ & $\Delta$MAE \\
\midrule
T2M & Largest-$m$ & 2.000565 & +0.149463 & 0.019452 & -0.001806 & 0.027219 & +0.002074 & 0.148729 & +0.031502 \\
T2M & Random & 0.417193 & -0.019567 & 0.003102 & -0.000138 & 0.003632 & +0.000052 & 0.027290 & -0.001931 \\
\bottomrule
\end{tabular*}
\end{table*}

\paragraph{Effects within variable groups.} Table~\ref{tab:supp_prototype_summary} extends the temperature analysis in Section~\ref{sec:prototype_analysis}. On GlobalForecast, removing the largest-$m$ contribution increases mean MAE for all four groups. Wind and humidity errors rise in 87 and 80 of the 96 cases, respectively, compared with 83 for temperature. Cloud-cover error rises in 61 cases, showing a less uniform benefit across examples. The corresponding forecast response is 6.00 times the random-control response for wind, 9.20 times for humidity, and 6.26 times for cloud cover. Thus, the contribution magnitude identifies terms that strongly affect predictions across different variable groups, while the error changes reveal how often those terms improve the forecast.

\paragraph{Propagation across variables.} Table~\ref{tab:supp_cross_global} reports all output variables after removing one contribution from a selected group. Removing the largest temperature contribution changes humidity forecasts by 0.2300 percentage points on average and wind forecasts by 0.0744m\,s$^{-1}$. These responses show that a local group correction propagates to other variables through subsequent processing. The direction of the error change depends on the output variable. For example, cloud-cover deletion slightly reduces wind, temperature, and humidity errors while increasing cloud-cover error. This analysis traces each contribution's effects across variables.

\begin{figure*}[!tp]
\centering
\includegraphics[width=0.82\textwidth]{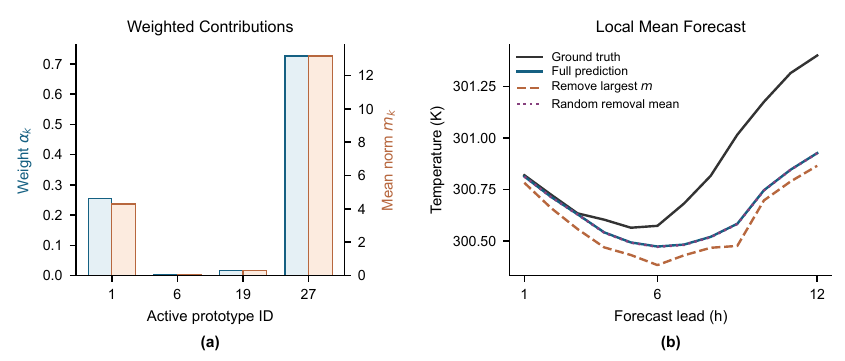}
\caption{Local temperature contributions and forecast responses on TyphoonTracking. (a) Matching weights $\alpha$ and mean weighted-residual norms $m$ for the active prototypes. (b) Patch-mean forecasts from the full model and the two deletion conditions, alongside ground truth. The random-deletion curve averages three draws from the other active prototypes.}
\label{fig:supp_prototype_regional}
\end{figure*}

\paragraph{Regional forecasting.} On TyphoonTracking, the largest-$m$ temperature deletion raises mean local MAE by 0.0315K and increases error in 31 of 45 paired cases. Random deletion changes MAE by $-0.00193$K. The largest-$m$ term produces a 0.1487K forecast response, compared with 0.0273K for the random control. Figure~\ref{fig:supp_prototype_regional} shows the representative case, and Table~\ref{tab:supp_cross_regional} reports the effects on pressure, wind, and temperature. For the three single-active cases, removing the sole temperature contribution increases mean MAE by 0.0853K, with error increasing in two cases. These cases have no random-other-active control and remain separate from the paired comparison.

\begin{figure*}[!tp]
\centering
\includegraphics[width=\textwidth]{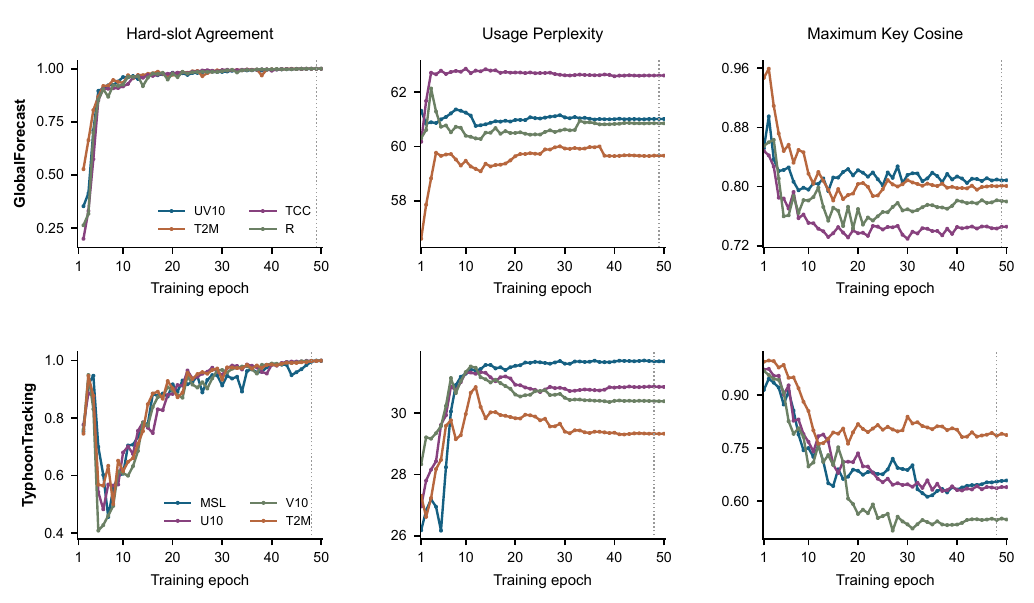}
\caption{Prototype-bank statistics over 50 training epochs on GlobalForecast (top) and TyphoonTracking (bottom). Columns show agreement of tracked slot assignments, usage perplexity, and the largest pairwise key cosine similarity. Statistics use queries from fixed training reference samples. No empty slots occur in either benchmark. Vertical dotted lines mark the validation-selected checkpoints.}
\label{fig:supp_bank_usage}
\end{figure*}

\paragraph{Prototype-bank usage during training.} Figure~\ref{fig:supp_bank_usage} tracks all 50 training epochs. Let $p_k$ be the fraction of queries from fixed training reference samples assigned to slot $k$ after refitting. Usage perplexity is $\exp(-\sum_k p_k\log p_k)$, and slot agreement is the fraction retaining the same tracked slot between consecutive refits. No slot is empty for either benchmark. At the selected checkpoints, usage perplexity ranges from 59.66 to 62.62 for the 64-slot GlobalForecast banks and from 29.33 to 31.68 for the 32-slot TyphoonTracking banks. Slot agreement ranges from 99.976\% to 100\% and from 99.573\% to 99.878\%, respectively. These statistics show broad use of the available slots and continuity of the training-reference assignments as refitting converges.

\subsection{Prototype History Consistency}
\label{sec:supp_history_consistency}

We examine whether histories associated with the same prototype exhibit consistent physical evolution on GlobalForecast. This analysis complements the contribution-removal tests in Section~\ref{sec:supp_prototype_results} by examining the historical patterns organized by prototype matching. We use all 17,495 test windows from 2017-2018 at a one-hour stride, with 128 spatial patches per window, giving 2,239,360 patch histories per variable group. The validation-selected model, prototype banks, and preprocessing statistics remain fixed. All 64 prototypes in each group are evaluated using their actual soft matching weights, without truncating the active set.

\paragraph{Physical histories.} For each $4\times4$ patch, we spatially average the 12 hourly observations in physical units and subtract the final observation from the entire history. This removes the local offset and exposes temporal evolution. U10 and V10 remain separate components. Let $d_i\in\mathbb{R}^{M}$ concatenate the resulting values at $-11,\ldots,-1$ hours, with $M=11C_g$ for a group containing $C_g$ variables. The zero-valued anchor at hour 0 is excluded from the dispersion calculation. Each window-patch history enters the analysis once per group, since its matching weights are shared across forecast leads.

\paragraph{Matched reference.} We compare each prototype-associated distribution with a reference matched by calendar month, patch latitude row, and local time of the final observation. We use all eight latitude rows and four six-hour local-solar-time bins. Calendar month follows UTC, and local solar time is obtained from UTC and patch longitude. Within each group defined by these three conditions, the reference uses the complete empirical distribution of physical histories. These distributions are mixed using the prototype's own matching-weight mass across these groups. This preserves its seasonal, latitudinal, and diurnal composition while removing the association between its weights and individual histories within each group. Reference moments are computed directly from the full distributions.

\paragraph{Dispersion measures.} Let $\alpha_{ik}$ be the weight of prototype $k$ for history $i$, $W_k=\sum_i\alpha_{ik}$, and $q_k=W_k/\sum_jW_j$. The weighted prototype mean is $\mu_k=\sum_i\alpha_{ik}d_i/W_k$, and the overall mean is $\mu=\sum_kq_k\mu_k$. We measure within-prototype and between-prototype dispersion as:
\begin{equation}
\begin{aligned}
D_{\mathrm{within}}^2 &= \sum_k q_k\frac{\sum_i\alpha_{ik}\lVert d_i-\mu_k\rVert_2^2}{M W_k},\\
D_{\mathrm{between}}^2 &= \sum_k q_k\frac{\lVert\mu_k-\mu\rVert_2^2}{M}.
\end{aligned}
\label{eq:supp_history_dispersion}
\end{equation}
The same definitions apply to the matched reference. Both measures are reported as root-mean-square (RMS) values in physical units. The relative reduction in within-prototype RMS is $1-D_{\mathrm{within}}/D_{\mathrm{within,ref}}$. UV10 averages squared deviations equally over its two components and the 11 history times before taking the square root.

\begin{table}[H]
\centering
\footnotesize
\setlength{\tabcolsep}{2pt}
\caption{Physical-history dispersion on GlobalForecast. $D$ and $D_{\mathrm{ref}}$ denote prototype and matched-reference RMS, respectively. Reduction refers to within-prototype RMS. T2M, R, UV10, and TCC use K, percentage points, m\,s$^{-1}$, and a fraction.}
\label{tab:supp_history_consistency}
\begin{tabular*}{\columnwidth}{@{\extracolsep{\fill}}lrrrrr@{}}
\toprule
& \multicolumn{2}{c}{Within RMS} & \multicolumn{2}{c}{Between RMS} & \\
\cmidrule(lr){2-3}\cmidrule(lr){4-5}
Group & $D$ & $D_{\mathrm{ref}}$ & $D$ & $D_{\mathrm{ref}}$ & Reduction\\
\midrule
T2M & 1.2915 & 1.7536 & 1.3362 & 0.6151 & 26.4\%\\
R & 3.6392 & 5.1764 & 4.2837 & 2.1907 & 29.7\%\\
UV10 & 0.7228 & 0.7924 & 0.3247 & 0.0021 & 8.8\%\\
TCC & 0.05786 & 0.05902 & 0.01170 & 0.00076 & 2.0\%\\
\bottomrule
\end{tabular*}
\end{table}

\paragraph{Results.} Table~\ref{tab:supp_history_consistency} shows lower within-prototype dispersion than the matched reference for all four groups. The reductions are 26.4\% for T2M, 29.7\% for R, 8.8\% for UV10, and 2.0\% for TCC. Between-prototype dispersion is also higher than its reference counterpart in each group. The prototype weights therefore organize physical variation into more consistent histories within prototypes and more distinct mean histories between prototypes, with the strongest reductions for temperature and humidity. Removing local time from the reference grouping retains the same direction of change, with within-reference ratios of 0.695, 0.647, 0.912, and 0.980, respectively.

\begin{figure*}[!tp]
\centering
\includegraphics[width=\textwidth,trim=0bp 32bp 0bp 0bp,clip]{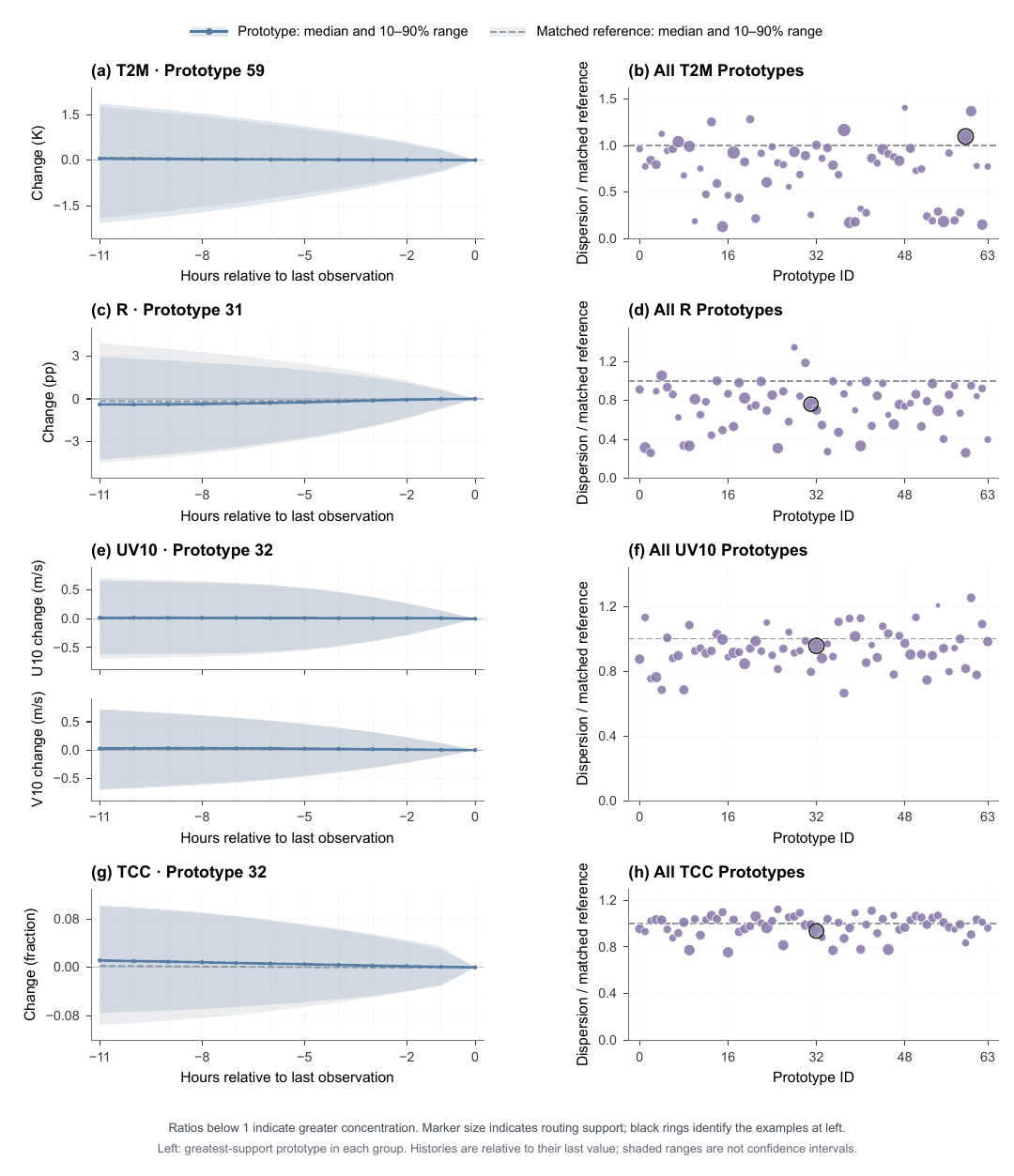}
\caption{Prototype-associated physical histories on GlobalForecast. Rows show T2M, R, UV10, and TCC. Left panels show the prototype with the greatest total matching-weight mass in each group. Curves are weighted medians, and shaded regions span the pointwise 10th-90th percentiles. Each history is relative to its final observation. UV10 retains U10 and V10 as separate components. Right panels show the within-prototype RMS divided by its matched-reference RMS for all 64 prototypes. Values below one indicate more concentrated histories. Marker area is proportional to matching-weight mass, and black rings identify the examples at left.}
\label{fig:supp_history_consistency}
\end{figure*}

Figure~\ref{fig:supp_history_consistency} displays the physical-history distributions and the dispersion ratios of all 64 prototypes per group. The history examples use the prototype with the greatest total matching-weight mass in each group. Weighted medians and pointwise 10th-90th percentile ranges summarize the associated histories and their matched references. Percentiles use 2,048 equal-width bins spanning the complete observed range for each component and history time. These full-test statistics link prototype matching to physical temporal patterns. Together with the contribution-removal results in Section~\ref{sec:supp_prototype_results}, they associate historical patterns with useful local forecast corrections.

\subsection{Prototype-Operator Pairing}
\label{sec:supp_pairing_intervention}

Section~\ref{sec:supp_history_consistency} relates prototype matching to physical historical patterns. In this section, we test whether forecast accuracy depends on the pairing between matched prototypes and their learned correction operators.

\paragraph{Pairing test.} We use the validation-selected models from Tables~\ref{tab:weatherbench_main} and~\ref{tab:era5_main}, with 64 and 32 prototypes per group, respectively. For each repetition, we independently sample a permutation $\pi_g$ within each group such that $\pi_g(k)\ne k$ for every prototype. All four groups are permuted simultaneously. Equation~\eqref{eq:method_operators} becomes:
\begin{equation}
\Delta h_{\ell,n}^{(g)}=\left(\sum_{k=1}^{K}\alpha_{n,k}^{(g)}B_{g,\pi_g(k)}\right)v_{\ell,n}^{(g)}.
\label{eq:supp_pairing_intervention}
\end{equation}
This retains every learned operator and the original matching weights, but assigns each prototype a different operator. The permutation remains fixed across all test windows, patches, and forecast leads within a repetition. Prototype keys and model parameters remain frozen, with no training or refitting. Post-correction fusion and decoding process the resulting features.

We reevaluate the original pairing and ten independently permuted pairings using BF16 inference and batch size 64. Each evaluation covers all 17,495 GlobalForecast test windows or all 8,737 TyphoonTracking test windows and all 12 hourly forecast leads. Metrics follow the physical-scale aggregation in Section~\ref{sec:supp_data_evaluation}. We report the mean and sample standard deviation across the ten permutations.

\begin{table*}[!tp]
\centering
\small
\setlength{\tabcolsep}{3pt}
\renewcommand{\arraystretch}{1.06}
\caption{Prototype-operator pairing test. Original denotes the unchanged pairing, and Permuted reports the mean $\pm$ sample standard deviation over ten permutations. Each metric is first computed over the complete test set for one permutation. Lower MSE, MAE, and RMSE and higher $R^2$ indicate better forecasts. The better value within each pair is blue and bold, and the other mean is underlined.}
\label{tab:supp_pairing_intervention}
\begin{tabular*}{\textwidth}{@{\extracolsep{\fill}}llrrrr@{}}
\toprule
Variable & Pairing & MSE$\downarrow$ & MAE$\downarrow$ & RMSE$\downarrow$ & $R^2\uparrow$ \\
\midrule
\multicolumn{6}{l}{\textit{GlobalForecast}} \\
T2M & Original & \textcolor{cvprblue}{\textbf{0.7850}} & \textcolor{cvprblue}{\textbf{0.5611}} & \textcolor{cvprblue}{\textbf{0.8860}} & \textcolor{cvprblue}{\textbf{0.9982}} \\
 & Permuted & $\underline{3.6588}\pm 0.2986$ & $\underline{1.1443}\pm 0.0357$ & $\underline{1.9113}\pm 0.0792$ & $\underline{0.9917}\pm 0.0007$ \\
R & Original & \textcolor{cvprblue}{\textbf{10.5224}} & \textcolor{cvprblue}{\textbf{2.2646}} & \textcolor{cvprblue}{\textbf{3.2438}} & \textcolor{cvprblue}{\textbf{0.9619}} \\
 & Permuted & $\underline{24.9588}\pm 1.1772$ & $\underline{3.3368}\pm 0.0541$ & $\underline{4.9946}\pm 0.1172$ & $\underline{0.9096}\pm 0.0043$ \\
UV10 & Original & \textcolor{cvprblue}{\textbf{0.7626}} & \textcolor{cvprblue}{\textbf{0.6013}} & \textcolor{cvprblue}{\textbf{0.8733}} & \textcolor{cvprblue}{\textbf{0.9671}} \\
 & Permuted & $\underline{1.4381}\pm 0.0529$ & $\underline{0.8104}\pm 0.0119$ & $\underline{1.1990}\pm 0.0220$ & $\underline{0.9380}\pm 0.0023$ \\
TCC & Original & \textcolor{cvprblue}{\textbf{0.0165}} & \textcolor{cvprblue}{\textbf{0.0892}} & \textcolor{cvprblue}{\textbf{0.1285}} & \textcolor{cvprblue}{\textbf{0.8199}} \\
 & Permuted & $\underline{0.0190}\pm 0.0001$ & $\underline{0.0970}\pm 0.0003$ & $\underline{0.1379}\pm 0.0003$ & $\underline{0.7928}\pm 0.0009$ \\
\midrule
\multicolumn{6}{l}{\textit{TyphoonTracking}} \\
MSL & Original & \textcolor{cvprblue}{\textbf{5888.75}} & \textcolor{cvprblue}{\textbf{53.1166}} & \textcolor{cvprblue}{\textbf{76.7382}} & \textcolor{cvprblue}{\textbf{0.9898}} \\
 & Permuted & $\underline{7043.71}\pm 146.46$ & $\underline{59.2875}\pm 0.9050$ & $\underline{83.9227}\pm 0.8736$ & $\underline{0.9878}\pm 0.0003$ \\
U10 & Original & \textcolor{cvprblue}{\textbf{0.9886}} & \textcolor{cvprblue}{\textbf{0.6793}} & \textcolor{cvprblue}{\textbf{0.9943}} & \textcolor{cvprblue}{\textbf{0.9479}} \\
 & Permuted & $\underline{1.1405}\pm 0.0093$ & $\underline{0.7418}\pm 0.0038$ & $\underline{1.0679}\pm 0.0044$ & $\underline{0.9400}\pm 0.0005$ \\
V10 & Original & \textcolor{cvprblue}{\textbf{1.1292}} & \textcolor{cvprblue}{\textbf{0.7248}} & \textcolor{cvprblue}{\textbf{1.0626}} & \textcolor{cvprblue}{\textbf{0.9387}} \\
 & Permuted & $\underline{1.3082}\pm 0.0057$ & $\underline{0.7934}\pm 0.0019$ & $\underline{1.1438}\pm 0.0025$ & $\underline{0.9290}\pm 0.0003$ \\
T2M & Original & \textcolor{cvprblue}{\textbf{0.5529}} & \textcolor{cvprblue}{\textbf{0.4804}} & \textcolor{cvprblue}{\textbf{0.7436}} & \textcolor{cvprblue}{\textbf{0.9943}} \\
 & Permuted & $\underline{1.2386}\pm 0.1276$ & $\underline{0.6972}\pm 0.0364$ & $\underline{1.1116}\pm 0.0576$ & $\underline{0.9873}\pm 0.0013$ \\
\bottomrule
\end{tabular*}
\end{table*}

\paragraph{Results.} Table~\ref{tab:supp_pairing_intervention} shows that permuting the pairing worsens all four metrics for every reported variable in every repetition. On GlobalForecast, mean MSE increases by 366.10\% for T2M, 137.20\% for R, 88.59\% for UV10, and 15.06\% for TCC. On TyphoonTracking, the increases are 19.61\% for MSL, 15.36\% for U10, 15.86\% for V10, and 124.02\% for T2M. Temperature shows the largest relative increase on both benchmarks. Lead-wise evaluation also gives higher MSE for every variable at each of the 12 forecast leads in all ten repetitions. The learned operators therefore provide more accurate corrections when paired with their original prototype references. Together with the history analysis, this result links the patterns selected by prototype matching to the operators that use those patterns to correct forecasts.

\subsection{Comparison with Persistence}
\label{sec:supp_persistence}

Persistence tests how much forecasting improves over leaving the observed weather unchanged. For each input window, it copies the final observed field to every forecast lead, $\widehat{Y}_{\ell}=X_{T_{\mathrm{in}}}$ for $\ell=1,\ldots,12$, without training. We evaluate all 17,495 GlobalForecast test windows from 2017-2018 and all 8,737 TyphoonTracking test windows from 2023 at a one-hour stride. ProtoCast uses the same validation-selected models and test results as Tables~\ref{tab:weatherbench_main} and~\ref{tab:era5_main}.

The comparison follows the preprocessing and metric definitions in Section~\ref{sec:supp_data_evaluation}. Both predictions and targets undergo the same inverse normalization. Errors are aggregated over all test windows, 12 forecast leads, and grid cells with equal element weights, and RMSE is the square root of the aggregated MSE. For UV10, MSE and MAE average the corresponding U10 and V10 scores, while $R^2$ weights the component scores by their target variances. Temperature, wind, pressure, and relative humidity use K, m\,s$^{-1}$, Pa, and percentage points, respectively, while TCC uses a fraction. Relative MSE reduction is $100(1-\mathrm{MSE}_{\mathrm{ours}}/\mathrm{MSE}_{\mathrm{persistence}})\%$.

\begin{table*}[!tp]
\centering
\small
\setlength{\tabcolsep}{3pt}
\renewcommand{\arraystretch}{1.14}
\caption{Comparison with persistence on both benchmarks. Metrics aggregate the complete test sets and all forecast leads. Lower MSE, MAE, and RMSE and higher $R^2$ indicate better forecasts. The final column reports relative MSE reduction.}
\label{tab:supp_persistence}
\begin{tabular*}{\textwidth}{@{\extracolsep{\fill}}l*{9}{r}@{}}
\toprule
& \multicolumn{4}{c}{Persistence} & \multicolumn{4}{c}{ProtoCast} & MSE \\
\cmidrule(lr){2-5}\cmidrule(lr){6-9}
Variable & MSE$\downarrow$ & MAE$\downarrow$ & RMSE$\downarrow$ & $R^2\uparrow$ & MSE$\downarrow$ & MAE$\downarrow$ & RMSE$\downarrow$ & $R^2\uparrow$ & reduction \\
\midrule
\multicolumn{10}{l}{\textit{GlobalForecast}} \\
T2M & 5.9785 & 1.2859 & 2.4451 & 0.9865 & 0.7850 & 0.5611 & 0.8860 & 0.9982 & 86.87\% \\
R & 65.2149 & 4.9748 & 8.0756 & 0.7637 & 10.5224 & 2.2646 & 3.2438 & 0.9619 & 83.87\% \\
UV10 & 3.9284 & 1.1971 & 1.9820 & 0.8307 & 0.7626 & 0.6013 & 0.8733 & 0.9671 & 80.59\% \\
TCC & 0.0302 & 0.1137 & 0.1739 & 0.6704 & 0.0165 & 0.0892 & 0.1285 & 0.8199 & 45.36\% \\
\midrule
\multicolumn{10}{l}{\textit{TyphoonTracking}} \\
MSL & 37712.56 & 136.6503 & 194.1972 & 0.9349 & 5888.81 & 53.1169 & 76.7386 & 0.9898 & 84.39\% \\
U10 & 3.3109 & 1.1938 & 1.8196 & 0.8257 & 0.9886 & 0.6793 & 0.9943 & 0.9479 & 70.14\% \\
V10 & 4.4007 & 1.3376 & 2.0978 & 0.7610 & 1.1292 & 0.7248 & 1.0626 & 0.9387 & 74.34\% \\
T2M & 6.7803 & 1.3978 & 2.6039 & 0.9303 & 0.5529 & 0.4804 & 0.7436 & 0.9943 & 91.85\% \\
\bottomrule
\end{tabular*}
\end{table*}

Table~\ref{tab:supp_persistence} shows that ProtoCast improves all four metrics for every variable on both benchmarks. On GlobalForecast, MSE falls by 86.87\% for T2M, 83.87\% for R, 80.59\% for UV10, and 45.36\% for TCC. The largest relative reduction is for temperature, while cloud cover shows a smaller but positive improvement. On TyphoonTracking, MSE decreases by 84.39\% for MSL, 70.14\% for U10, 74.34\% for V10, and 91.85\% for T2M. These reductions show that the model predicts changes over the 12-hour horizon more accurately than retaining the final observation.

\subsection{Area-Weighted Global Forecast Evaluation}
\label{sec:supp_area_weighted}

GlobalForecast uses a latitude-longitude grid with equally spaced coordinates, but its cells cover less surface area towards the poles. Equal-grid averaging therefore gives high-latitude regions more influence per unit area and can favor models whose improvements are concentrated there. Cosine-latitude weighting makes each cell contribute approximately in proportion to its surface area. This experiment tests whether ProtoCast's improvements and the method ranking persist when errors represent the Earth's surface more evenly. We evaluate every test window and all 12 hourly leads using the stored grid latitudes. The four learned models retain the validation-selected checkpoints in Table~\ref{tab:weatherbench_main}. Both weighting schemes share each model's predictions from a common BF16 environment, without retraining. Persistence follows Section~\ref{sec:supp_persistence}.

For latitude $\varphi_i$ in radians, we define the normalized row weight and weighted average of a scalar field $z$ as:
\begin{equation}
\begin{aligned}
w_i&=\frac{\cos\varphi_i}{H^{-1}\sum_{h=1}^{H}\cos\varphi_h},\\
\langle z\rangle_w&=\frac{1}{NLHW}\sum_{n=1}^{N}\sum_{\ell=1}^{L}\sum_{i=1}^{H}\sum_{j=1}^{W}w_i z_{n\ell ij},
\end{aligned}
\label{eq:supp_area_weights}
\end{equation}
where $N=17{,}495$, $L=12$, $H=32$, and $W=64$. For prediction error $e=\widehat{Y}-Y$, area-weighted MSE and MAE are $\langle e^2\rangle_w$ and $\langle|e|\rangle_w$, respectively. RMSE is the square root of the aggregated MSE, and $R^2=1-\langle e^2\rangle_w/\langle(Y-\langle Y\rangle_w)^2\rangle_w$. For UV10, MSE and MAE average the corresponding area-weighted U10 and V10 scores. Its $R^2$ weights the component scores by their area-weighted target variances, each measured around its own weighted target mean. All metrics use the physical scales in Section~\ref{sec:supp_data_evaluation}.

\begin{table*}[!tp]
\centering
\small
\setlength{\tabcolsep}{4pt}
\renewcommand{\arraystretch}{1.06}
\caption{Area-weighted evaluation on GlobalForecast. Equal-grid MSE uses the same predictions for comparison. Area-weighted metrics aggregate the complete test set and all forecast leads. The best area-weighted value for each variable and metric is blue and bold, and the second best is underlined.}
\label{tab:supp_area_weighted}
\begin{tabular*}{\textwidth}{@{\extracolsep{\fill}}llrrrrr@{}}
\toprule
 & & Equal-grid & \multicolumn{4}{c}{Area-weighted} \\
\cmidrule(lr){4-7}
Variable & Method & MSE$\downarrow$ & MSE$\downarrow$ & MAE$\downarrow$ & RMSE$\downarrow$ & $R^2\uparrow$ \\
\midrule
T2M & Persistence & 5.9785 & 6.8438 & 1.3327 & 2.6161 & 0.9700 \\
 & TAU & 2.3887 & 1.9577 & 0.9995 & 1.3992 & 0.9914 \\
 & Met2Net & 0.9608 & \underline{0.8038} & \underline{0.6102} & \underline{0.8965} & \underline{0.9965} \\
 & PredFormer & 1.1814 & 0.9798 & 0.6813 & 0.9899 & 0.9957 \\
 & ProtoCast & 0.7850 & \textcolor{cvprblue}{\textbf{0.6560}} & \textcolor{cvprblue}{\textbf{0.5046}} & \textcolor{cvprblue}{\textbf{0.8100}} & \textcolor{cvprblue}{\textbf{0.9971}} \\
\midrule
R & Persistence & 65.2149 & 73.0354 & 5.2861 & 8.5461 & 0.7341 \\
 & TAU & 19.6103 & 18.5490 & 3.2047 & 4.3068 & 0.9325 \\
 & Met2Net & 12.3275 & 11.6101 & 2.4235 & 3.4074 & 0.9577 \\
 & PredFormer & 11.0174 & \underline{10.2466} & \underline{2.2813} & \underline{3.2010} & \underline{0.9627} \\
 & ProtoCast & 10.5224 & \textcolor{cvprblue}{\textbf{9.7492}} & \textcolor{cvprblue}{\textbf{2.1949}} & \textcolor{cvprblue}{\textbf{3.1224}} & \textcolor{cvprblue}{\textbf{0.9645}} \\
\midrule
UV10 & Persistence & 3.9284 & 3.6726 & 1.1532 & 1.9164 & 0.8345 \\
 & TAU & 1.3588 & 1.2805 & 0.8336 & 1.1316 & 0.9423 \\
 & Met2Net & 1.0297 & 0.9842 & 0.7197 & 0.9921 & 0.9556 \\
 & PredFormer & 0.7817 & \underline{0.7356} & \underline{0.6020} & \underline{0.8576} & \underline{0.9669} \\
 & ProtoCast & 0.7626 & \textcolor{cvprblue}{\textbf{0.7106}} & \textcolor{cvprblue}{\textbf{0.5849}} & \textcolor{cvprblue}{\textbf{0.8430}} & \textcolor{cvprblue}{\textbf{0.9680}} \\
\midrule
TCC & Persistence & 0.0302 & 0.0299 & 0.1180 & 0.1730 & 0.6667 \\
 & TAU & 0.0218 & 0.0218 & 0.1107 & 0.1476 & 0.7572 \\
 & Met2Net & 0.0175 & 0.0177 & 0.0964 & 0.1332 & 0.8025 \\
 & PredFormer & 0.0162 & \textcolor{cvprblue}{\textbf{0.0163}} & \underline{0.0929} & \textcolor{cvprblue}{\textbf{0.1276}} & \textcolor{cvprblue}{\textbf{0.8187}} \\
 & ProtoCast & 0.0165 & \underline{0.0164} & \textcolor{cvprblue}{\textbf{0.0920}} & \underline{0.1279} & \underline{0.8178} \\
\bottomrule
\end{tabular*}
\end{table*}

As shown in Table~\ref{tab:supp_area_weighted}, ProtoCast achieves the best area-weighted result in 13 of the 16 variable-metric combinations. Its MSE is 18.38\% lower than Met2Net for T2M and 4.85\% and 3.39\% lower than PredFormer for R and UV10, respectively. PredFormer retains the lowest MSE and RMSE of TCC and the highest $R^2$. ProtoCast has 0.50\% higher MSE of TCC but the lowest MAE of TCC. The temperature, humidity, and wind improvements therefore persist when high-latitude cells contribute in proportion to their approximate surface area.

\subsection{Predictive Role of Prototype Corrections}
\label{sec:supp_correction_effects}

ProtoCast links matched historical patterns to explicit feature corrections. To understand the predictive role of these corrections, we ask whether keeping an identified prototype contribution improves the forecast and at which forecast leads its effect is strongest. Removing one contribution provides a direct comparison with the unchanged model. A random perturbation of the same magnitude provides a separate comparison for the direction of the feature change. This analysis connects the contribution decomposition in Equation~\eqref{eq:method_decomposition} to observable changes in forecast accuracy.

\begin{figure*}[!tp]
\centering
\includegraphics[width=\textwidth]{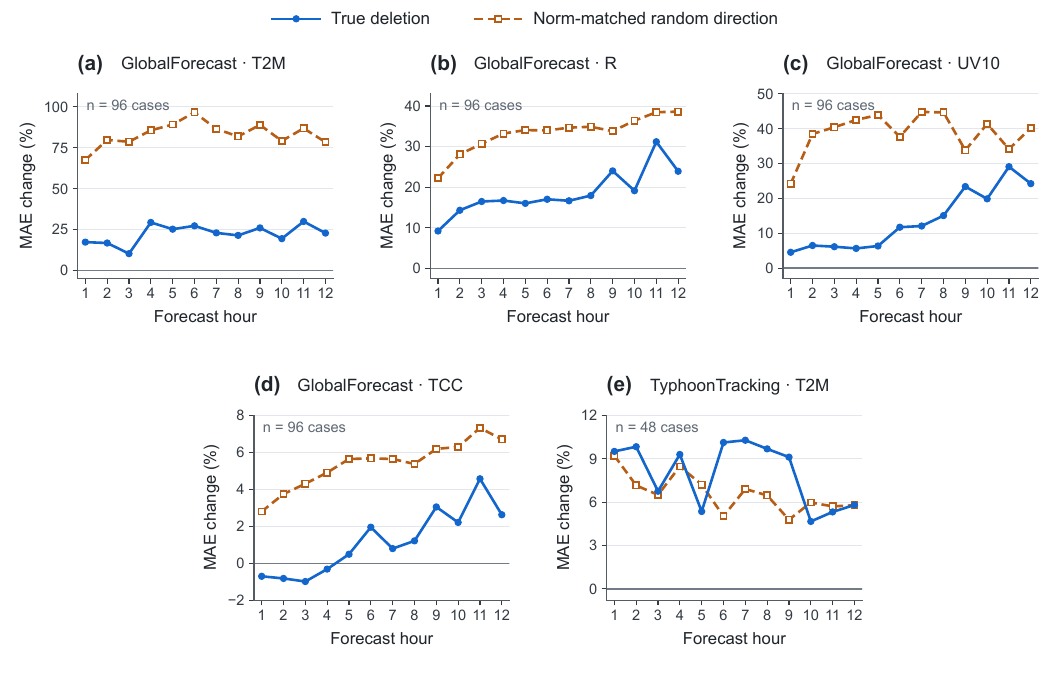}
\caption{Forecast error changes after removing one prototype contribution or applying a norm-matched random perturbation. Panels (a)-(d) show GlobalForecast T2M, R, UV10, and TCC, respectively. Panel (e) shows TyphoonTracking T2M. Each curve reports the percentage change in mean patch MAE relative to the full prediction at each lead, using Equation~\eqref{eq:supp_correction_percent}. Positive values indicate worse predictions. Blue denotes contribution deletion, and orange denotes the random control.}
\label{fig:supp_correction_horizon}
\end{figure*}

\paragraph{Fixed-model tests.} We reuse the validation-selected models and sampled windows and patches in Section~\ref{sec:supp_prototype_analysis}. We analyze 96 GlobalForecast cases, each evaluated separately for T2M, R, UV10, and TCC. For TyphoonTracking, we test the T2M group in 48 cases. All 48 enter this analysis because a random orthogonal transformation does not require a second active prototype. For each case and group, we select the single weighted contribution with the largest average norm across the 12 forecast leads, using Equation~\eqref{eq:supp_contribution_magnitude}. We keep the routing weights, other contributions, model parameters, and normalization fixed. At the selected group and patch, we compare the full corrected feature with contribution removal and a perturbation in a random direction:
\begin{equation}
\begin{aligned}
\widetilde h_{\ell}^{\mathrm{del}}&=\widetilde h_{\ell}-r_{\ell},\\
\widetilde h_{\ell}^{\mathrm{rnd},j}&=\widetilde h_{\ell}-Q_jr_{\ell},
\end{aligned}
\label{eq:supp_norm_matched_intervention}
\end{equation}
where $r_{\ell}$ is the selected prototype contribution and $Q_j^\top Q_j=I$. We draw ten random orthogonal transformations per case and selected group. Each transformation is shared across all 12 leads, preserving the perturbation norm at each lead and the inner products between leads. We recompute post-correction variable fusion, the final spatial block, and decoding after each feature modification. The analysis contains 432 case-group comparisons and 4,320 random directions, with frozen FP32 inference and no retraining or prototype refitting.

\paragraph{Error calculation.} For case $i$ and lead $\ell$, $E_{i\ell}^{c}$ is the physical-unit MAE of condition $c$ over the selected patch's $4\times4$ output cells. For UV10, we average the U10 and V10 absolute errors. We first average the ten random directions within each case and lead to obtain $E_{i\ell}^{\mathrm{rnd}}$. We then average the selected leads within each case and the cases within each benchmark-variable comparison. The percentage change for a set of leads $S$ is:
\begin{equation}
P_S^{c}=100\left(\frac{\sum_i\sum_{\ell\in S}E_{i\ell}^{c}}{\sum_i\sum_{\ell\in S}E_{i\ell}^{\mathrm{full}}}-1\right).
\label{eq:supp_correction_percent}
\end{equation}
Positive values indicate higher error than the full prediction. The case percentage in Table~\ref{tab:supp_correction_effects} counts cases whose 12-lead MAE increases after deletion.

\begin{table*}[!tp]
\centering
\small
\setlength{\tabcolsep}{3.5pt}
\renewcommand{\arraystretch}{1.12}
\caption{Effect of removing one prototype contribution. MAE uses the selected patch and all 12 leads. Random denotes ten norm-matched directions averaged within each case. The last two columns compare deletion with the full prediction. R uses percentage points, and TCC uses a fraction.}
\label{tab:supp_correction_effects}
\begin{tabular*}{\textwidth}{@{\extracolsep{\fill}}llrrrrrr@{}}
\toprule
Benchmark & Variable & Cases & Full MAE & Deletion MAE & Random MAE & MAE increase & Cases worse \\
\midrule
GlobalForecast & T2M (K) & 96 & 0.5831 & 0.7192 & 1.0743 & 23.3\% & 86.5\% \\
GlobalForecast & R (pp) & 96 & 2.2370 & 2.6835 & 3.0101 & 20.0\% & 83.3\% \\
GlobalForecast & UV10 (m\,s$^{-1}$) & 96 & 0.6147 & 0.7150 & 0.8579 & 16.3\% & 90.6\% \\
GlobalForecast & TCC & 96 & 0.0894 & 0.0908 & 0.0945 & 1.5\% & 63.5\% \\
TyphoonTracking & T2M (K) & 48 & 0.4475 & 0.4823 & 0.4758 & 7.8\% & 68.8\% \\
\bottomrule
\end{tabular*}
\end{table*}

\paragraph{Predictive benefit of the selected contributions.} Removing the selected contribution increases average MAE in all five comparisons. On GlobalForecast, the MAEs of T2M, R, and UV10 increase by 23.3\%, 20.0\%, and 16.3\%, respectively. Their case-level errors increase in 86.5\%, 83.3\%, and 90.6\% of cases. The effect is smaller for TCC, with a 1.5\% increase in average MAE and higher error in 63.5\% of cases. On TyphoonTracking, the MAE of T2M increases by 7.8\%, with 68.8\% of cases becoming less accurate. These results show that the selected prototype contributions reduce local forecast error on average.

\paragraph{Dependence on the forecast lead.} Figure~\ref{fig:supp_correction_horizon} shows all 12 leads for every benchmark-variable comparison. On GlobalForecast, the relative effect of removing a wind contribution increases from 5.9\% in hours 1-4 to 24.2\% in hours 9-12. The corresponding effect for humidity grows from 15.1\% to 24.6\%. Temperature corrections also have a larger average effect in the final four hours than in the first four hours. TCC has a different pattern, with deletion reducing MAE by 0.7\% in hours 1-4 but increasing it by 3.1\% in hours 9-12. TyphoonTracking temperature changes by 8.7\%, 9.0\%, and 6.2\% over the three consecutive four-hour periods. The role of a prototype contribution therefore varies with the predicted variable and forecast lead. The decomposition makes these differences measurable by linking each feature correction to its forecast effect.

\begin{table*}[!tp]
\centering
\small
\setlength{\tabcolsep}{4pt}
\renewcommand{\arraystretch}{1.62}
\caption{Complete 12-lead paired random-control comparison. Differences use each output variable's physical units. Patch and full-field columns use the same cases and feature modifications. The case percentages compare deletion with the case's mean random control.}
\label{tab:supp_correction_random_all}
\begin{tabular*}{\textwidth}{@{\extracolsep{\fill}}llrrrrrr@{}}
\toprule
 & & \multicolumn{3}{c}{Selected patch} & \multicolumn{3}{c}{Full output field} \\
\cmidrule(lr){3-5}\cmidrule(lr){6-8}
Modified group & Output & Mean $d_i$ & Median $d_i$ & Cases $d_i>0$ & Mean $d_i$ & Median $d_i$ & Cases $d_i>0$ \\
\midrule
\multicolumn{8}{l}{\textit{GlobalForecast, 96 cases per modified group}} \\
T2M & T2M (K) & -0.3551 & -0.3577 & 4.2\% & -0.0029 & -0.0029 & 5.2\% \\
T2M & R (pp) & -0.0080 & -0.0093 & 42.7\% & $-7.719\!\times\!10^{-5}$ & $-8.447\!\times\!10^{-5}$ & 40.6\% \\
T2M & UV10 (m\,s$^{-1}$) & -0.0020 & -0.0012 & 42.7\% & $-2.174\!\times\!10^{-5}$ & $-2.381\!\times\!10^{-5}$ & 40.6\% \\
T2M & TCC & -0.0013 & -0.0009 & 35.4\% & $-9.735\!\times\!10^{-6}$ & $-5.810\!\times\!10^{-6}$ & 36.5\% \\
T2M & U10 (m\,s$^{-1}$) & -0.0004 & -0.0028 & 41.7\% & $-1.019\!\times\!10^{-5}$ & $-2.354\!\times\!10^{-5}$ & 41.7\% \\
T2M & V10 (m\,s$^{-1}$) & -0.0036 & -0.0023 & 41.7\% & $-3.329\!\times\!10^{-5}$ & $-3.065\!\times\!10^{-5}$ & 41.7\% \\
\addlinespace[3pt]
R & T2M (K) & 0.0007 & -0.0010 & 45.8\% & $-2.136\!\times\!10^{-7}$ & $-6.180\!\times\!10^{-6}$ & 45.8\% \\
R & R (pp) & -0.3266 & -0.2498 & 15.6\% & -0.0028 & -0.0023 & 14.6\% \\
R & UV10 (m\,s$^{-1}$) & $6.175\!\times\!10^{-5}$ & -0.0004 & 46.9\% & $2.792\!\times\!10^{-6}$ & $-1.850\!\times\!10^{-6}$ & 47.9\% \\
R & TCC & -0.0004 & -0.0003 & 39.6\% & $-3.328\!\times\!10^{-6}$ & $-2.004\!\times\!10^{-6}$ & 43.8\% \\
R & U10 (m\,s$^{-1}$) & 0.0006 & 0.0007 & 53.1\% & $8.447\!\times\!10^{-6}$ & $4.719\!\times\!10^{-6}$ & 53.1\% \\
R & V10 (m\,s$^{-1}$) & -0.0005 & -0.0004 & 45.8\% & $-2.863\!\times\!10^{-6}$ & $-6.190\!\times\!10^{-6}$ & 45.8\% \\
\addlinespace[3pt]
UV10 & T2M (K) & -0.0026 & -0.0021 & 38.5\% & $-1.983\!\times\!10^{-5}$ & $-1.475\!\times\!10^{-5}$ & 38.5\% \\
UV10 & R (pp) & -0.0267 & -0.0288 & 29.2\% & -0.0002 & -0.0003 & 36.5\% \\
UV10 & UV10 (m\,s$^{-1}$) & -0.1429 & -0.1505 & 6.3\% & -0.0012 & -0.0011 & 7.3\% \\
UV10 & TCC & -0.0004 & -0.0002 & 44.8\% & $-2.868\!\times\!10^{-6}$ & $-4.031\!\times\!10^{-7}$ & 44.8\% \\
UV10 & U10 (m\,s$^{-1}$) & -0.1619 & -0.1614 & 7.3\% & -0.0013 & -0.0012 & 9.4\% \\
UV10 & V10 (m\,s$^{-1}$) & -0.1240 & -0.1159 & 15.6\% & -0.0010 & -0.0009 & 10.4\% \\
\addlinespace[3pt]
TCC & T2M (K) & -0.0012 & -0.0001 & 44.8\% & $-1.357\!\times\!10^{-5}$ & $-2.004\!\times\!10^{-6}$ & 42.7\% \\
TCC & R (pp) & -0.0050 & -0.0020 & 42.7\% & $-4.328\!\times\!10^{-5}$ & $-6.194\!\times\!10^{-6}$ & 46.9\% \\
TCC & UV10 (m\,s$^{-1}$) & -0.0003 & -0.0001 & 42.7\% & $-2.716\!\times\!10^{-6}$ & $-2.039\!\times\!10^{-7}$ & 46.9\% \\
TCC & TCC & -0.0037 & -0.0028 & 20.8\% & $-3.532\!\times\!10^{-5}$ & $-2.581\!\times\!10^{-5}$ & 21.9\% \\
TCC & U10 (m\,s$^{-1}$) & -0.0005 & -0.0002 & 38.5\% & $-3.585\!\times\!10^{-6}$ & $-9.333\!\times\!10^{-7}$ & 42.7\% \\
TCC & V10 (m\,s$^{-1}$) & -0.0001 & $-5.078\!\times\!10^{-5}$ & 44.8\% & $-1.846\!\times\!10^{-6}$ & $-7.118\!\times\!10^{-7}$ & 43.8\% \\
\midrule
\multicolumn{8}{l}{\textit{TyphoonTracking, 48 cases per modified group}} \\
T2M & MSL (Pa) & 0.0790 & -0.0583 & 43.8\% & 0.0102 & 0.0020 & 52.1\% \\
T2M & U10 (m\,s$^{-1}$) & -0.0011 & -0.0011 & 41.7\% & $-2.750\!\times\!10^{-5}$ & $-2.815\!\times\!10^{-5}$ & 37.5\% \\
T2M & V10 (m\,s$^{-1}$) & 0.0016 & 0.0034 & 56.3\% & $-5.864\!\times\!10^{-6}$ & $-1.056\!\times\!10^{-5}$ & 50.0\% \\
T2M & T2M (K) & 0.0065 & -0.0022 & 45.8\% & $-1.740\!\times\!10^{-5}$ & $2.156\!\times\!10^{-5}$ & 50.0\% \\
\bottomrule
\end{tabular*}
\end{table*}

\paragraph{Norm-matched random comparison.} On GlobalForecast, random perturbations cause larger mean MAE increases than contribution deletion in each modified group's output. All four groups have negative mean and median paired differences. On TyphoonTracking, deleting the temperature contribution is more harmful by 0.0065K on average, but the median difference is $-0.0022$K and deletion is more harmful in 45.8\% of cases. Thus, random directions cause greater disruption on GlobalForecast, while TyphoonTracking gives mixed results. Table~\ref{tab:supp_correction_effects} compares the full forecast with contribution deletion to measure predictive benefit. Together with the leadwise analysis, it shows whether and when individual contributions help forecasts, making their predictive roles more interpretable.

Table~\ref{tab:supp_correction_random_all} reports comparisons for every recorded output variable within the selected patch and across the full field. For each case, $d_i=\overline E_i^{\mathrm{del}}-\overline E_i^{\mathrm{rnd}}$, with bars averaging all 12 leads after averaging random directions. We report mean and median $d_i$ and the percentage with $d_i>0$. Positive values mean deletion is more harmful than the random control.

\clearpage
\subsection{Comparison Across Training Runs}
\label{sec:supp_training_runs}

Random initialization and training order can affect forecasting accuracy even when hyperparameters remain fixed. To examine these variations, we compare ProtoCast with w/o correction and Plain adapter on TyphoonTracking using three different random seeds. Each seed is shared by all three methods. For each method, we retain the data splits, model configuration, optimizer settings, and 50-epoch schedule in Section~\ref{sec:supp_ablation_protocol}. Weights and optimizer states are initialized from scratch. We evaluate each validation-best checkpoint on all 8,737 test windows.

Table~\ref{tab:supp_training_runs} reports each run and the arithmetic mean across the three runs. ProtoCast achieves the lowest MAE for MSL, U10, V10, and T2M in every run. Its mean MAEs are also lower than those of both controls for all four variables, reproducing the benefit of prototype-conditioned corrections over removing corrections or using an unconditioned adapter in these runs.

We measure the variation of ProtoCast for variable $v$ as $\rho_v=100(\max_s e_{s,v}-\min_s e_{s,v})/\bar e_v$, where $e_{s,v}$ is the MAE in run $s$ and $\bar e_v$ is its arithmetic mean across the three runs. The relative ranges are 1.2456\%, 1.2640\%, 1.3023\%, and 1.4940\% for MSL, U10, V10, and T2M, respectively. The relative range remains below 1.5\% for all four variables in these runs.

\begin{table}[H]
\centering
\caption{MAE for three matched TyphoonTracking training runs. Lower is better. The final block reports arithmetic means.}
\label{tab:supp_training_runs}
\small
\setlength{\tabcolsep}{2pt}
\renewcommand{\arraystretch}{1.12}
\begin{tabular*}{\columnwidth}{@{\extracolsep{\fill}}llrrrr@{}}
\toprule
Run & Method & \shortstack{MSL\\(Pa)} & \shortstack{U10\\(m/s)} & \shortstack{V10\\(m/s)} & \shortstack{T2M\\(K)} \\
\midrule
1 & Ours & \textcolor{cvprblue}{\textbf{53.1169}} & \textcolor{cvprblue}{\textbf{0.6793}} & \textcolor{cvprblue}{\textbf{0.7248}} & \textcolor{cvprblue}{\textbf{0.4804}} \\
 & w/o correction & \underline{53.7532} & 0.6842 & \underline{0.7280} & \underline{0.4811} \\
 & Plain adapter & 53.9834 & \underline{0.6813} & 0.7289 & 0.4814 \\
\midrule
2 & Ours & \textcolor{cvprblue}{\textbf{53.7829}} & \textcolor{cvprblue}{\textbf{0.6879}} & \textcolor{cvprblue}{\textbf{0.7343}} & \textcolor{cvprblue}{\textbf{0.4876}} \\
 & w/o correction & 54.7514 & \underline{0.6973} & \underline{0.7443} & \underline{0.5032} \\
 & Plain adapter & \underline{54.4896} & 0.7104 & 0.7544 & 0.5063 \\
\midrule
3 & Ours & \textcolor{cvprblue}{\textbf{53.4914}} & \textcolor{cvprblue}{\textbf{0.6828}} & \textcolor{cvprblue}{\textbf{0.7314}} & \textcolor{cvprblue}{\textbf{0.4869}} \\
 & w/o correction & 53.8690 & 0.6925 & 0.7374 & 0.4896 \\
 & Plain adapter & \underline{53.7899} & \underline{0.6894} & \underline{0.7341} & \underline{0.4892} \\
\midrule
Mean & Ours & \textcolor{cvprblue}{\textbf{53.4637}} & \textcolor{cvprblue}{\textbf{0.6833}} & \textcolor{cvprblue}{\textbf{0.7302}} & \textcolor{cvprblue}{\textbf{0.4850}} \\
 & w/o correction & 54.1245 & \underline{0.6913} & \underline{0.7366} & \underline{0.4913} \\
 & Plain adapter & \underline{54.0876} & 0.6937 & 0.7392 & 0.4923 \\
\bottomrule
\end{tabular*}
\end{table}

\section{Additional Visualizations}
\label{sec:supp_visualizations}

\subsection{Forecast Fields and Spatial Errors}
\label{sec:supp_forecast_fields}

Figures~\ref{fig:supp_forecast_00480}, \ref{fig:supp_forecast_14712}, \ref{fig:supp_forecast_04188}, \ref{fig:supp_forecast_13440}, \ref{fig:supp_forecast_08496}, and~\ref{fig:supp_forecast_16296} extend Figure~\ref{fig:qualitative_forecast} with six additional 12-hour forecasts from the 2017-2018 GlobalForecast test split. Each comparison shows the predicted weather fields and spatial errors for TAU, Met$^2$Net, PredFormer, and ProtoCast, using the same color scales across methods and cases.

ProtoCast achieves the lowest full-field MAE for UV10, T2M, TCC, and R in all six examples.

\onecolumn
\makeatletter
\global\@colht\textheight
\global\@colroom\textheight
\global\vsize\textheight
\makeatother
\noindent\begin{minipage}{\textwidth}
    \captionsetup{type=figure,skip=4pt}
    \centering
    \includegraphics[width=\textwidth]{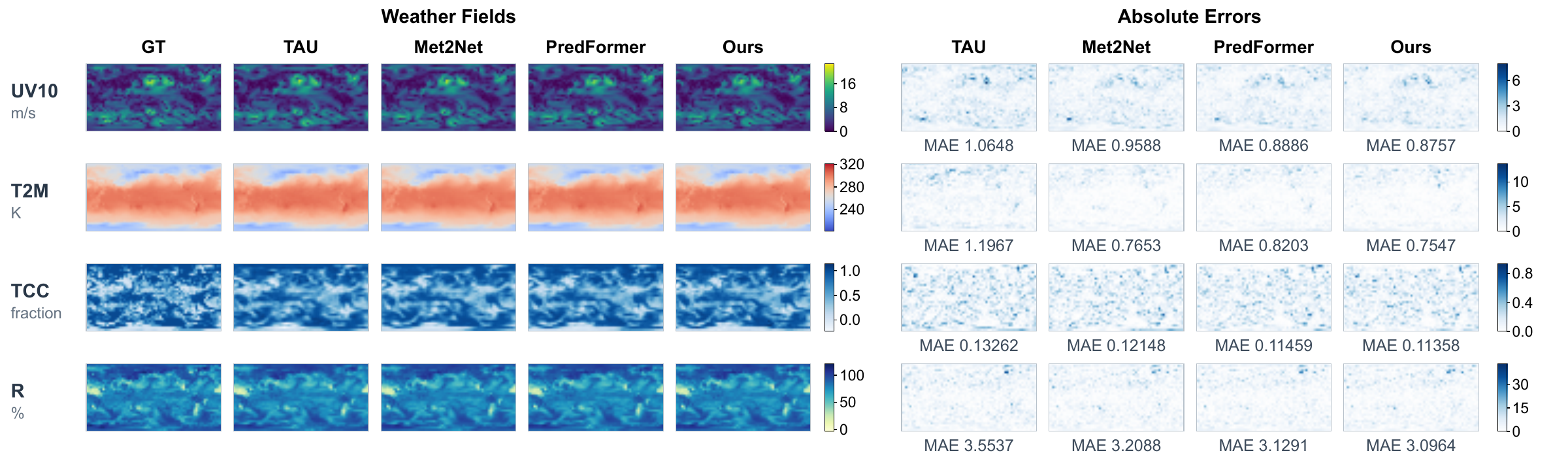}
    \caption{Qualitative comparison of 12-hour forecasts on a GlobalForecast test example. The left panels show ground truth (GT) and predictions from TAU, Met$^2$Net, PredFormer, and Ours. The right panels show absolute errors. Rows show UV10, T2M, TCC, and R. UV10 displays wind speed $\sqrt{u^2+v^2}$ with component error $(|\hat{u}-u|+|\hat{v}-v|)/2$, where hats denote predictions. Scalar errors are absolute differences from GT in physical units. Each MAE averages the full $32\times64$ field.}
    \label{fig:supp_forecast_00480}
\end{minipage}\par\addvspace{6pt}

\noindent\begin{minipage}{\textwidth}
    \captionsetup{type=figure,skip=4pt}
    \centering
    \includegraphics[width=\textwidth]{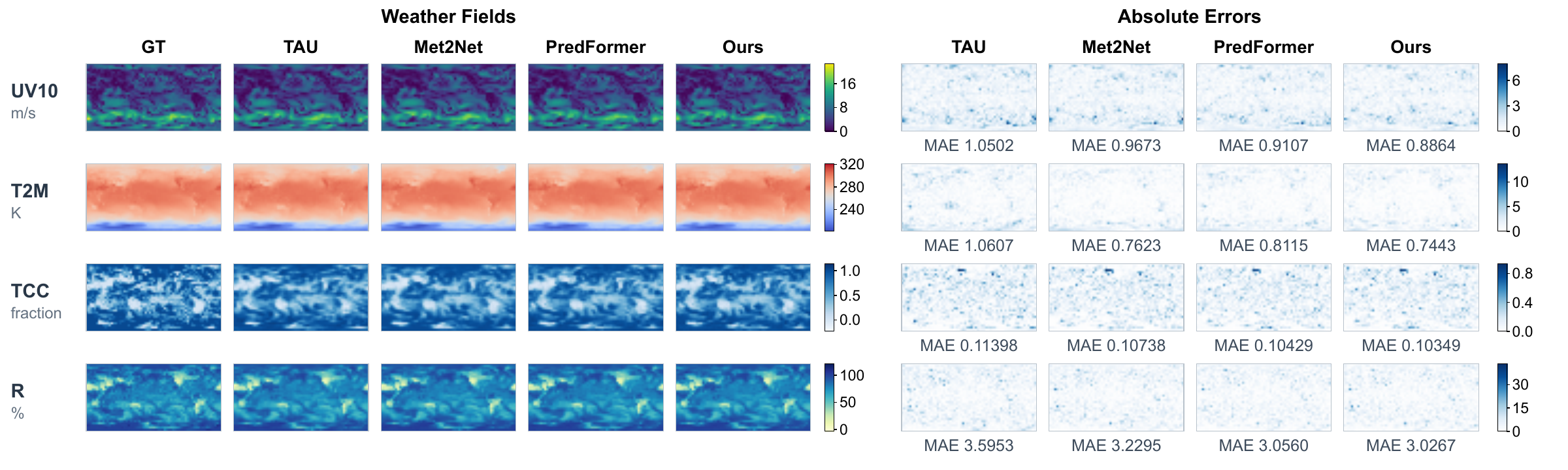}
    \caption{Qualitative comparison of 12-hour forecasts on a GlobalForecast test example. Weather fields, absolute errors, color scales, and full-field MAE follow the conventions in Figure~\ref{fig:supp_forecast_00480}.}
    \label{fig:supp_forecast_14712}
\end{minipage}\par\addvspace{6pt}

\noindent\begin{minipage}{\textwidth}
    \captionsetup{type=figure,skip=4pt}
    \centering
    \includegraphics[width=\textwidth]{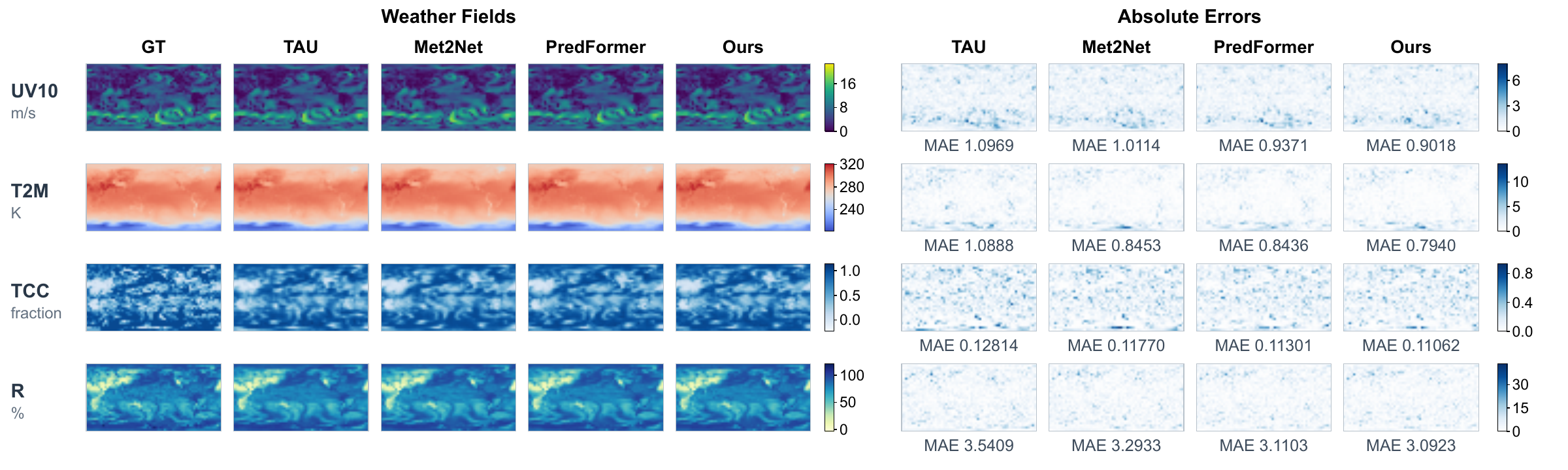}
    \caption{Qualitative comparison of 12-hour forecasts on a GlobalForecast test example. Weather fields, absolute errors, color scales, and full-field MAE follow the conventions in Figure~\ref{fig:supp_forecast_00480}.}
    \label{fig:supp_forecast_04188}
\end{minipage}\par\addvspace{6pt}

\noindent\begin{minipage}{\textwidth}
    \captionsetup{type=figure,skip=4pt}
    \centering
    \includegraphics[width=\textwidth]{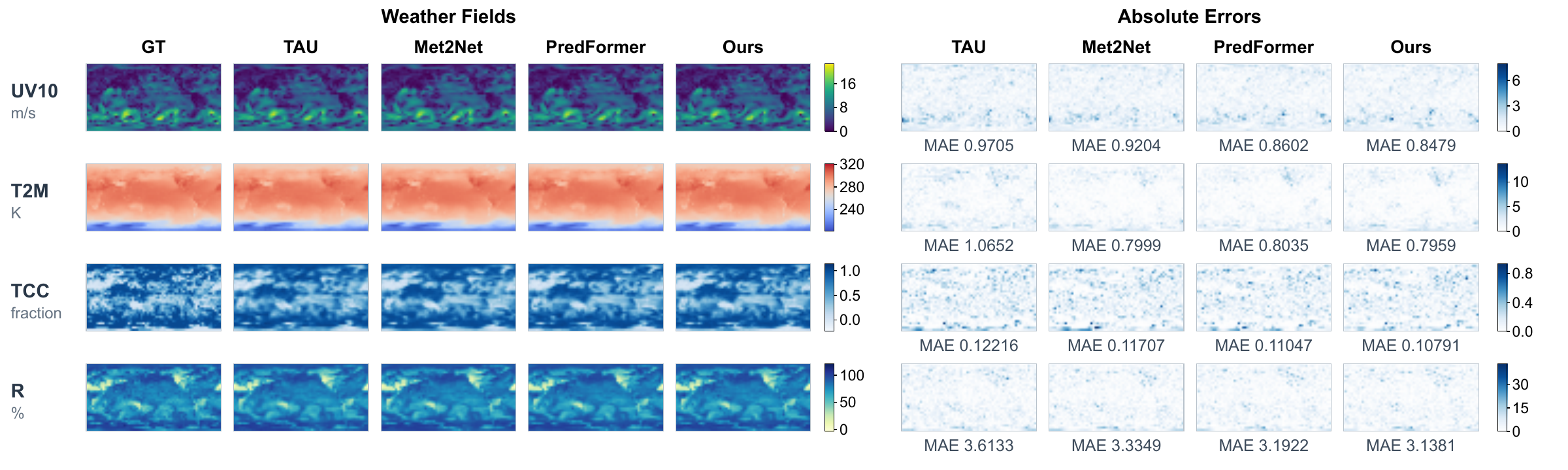}
    \caption{Qualitative comparison of 12-hour forecasts on a GlobalForecast test example. Weather fields, absolute errors, color scales, and full-field MAE follow the conventions in Figure~\ref{fig:supp_forecast_00480}.}
    \label{fig:supp_forecast_13440}
\end{minipage}\par\addvspace{6pt}

\noindent\begin{minipage}{\textwidth}
    \captionsetup{type=figure,skip=4pt}
    \centering
    \includegraphics[width=\textwidth]{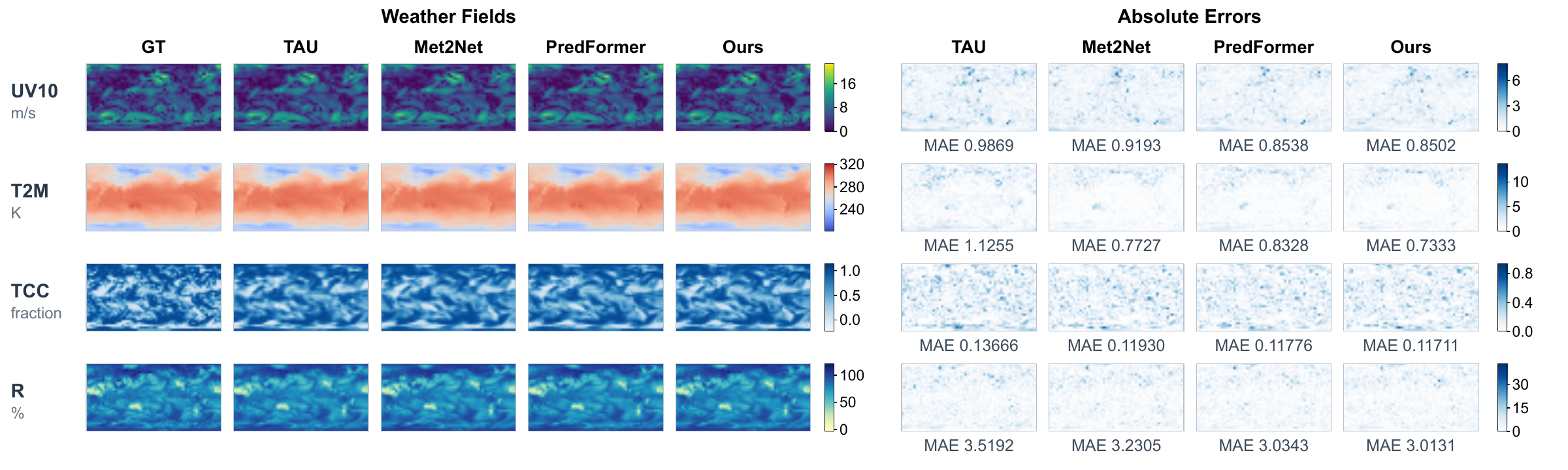}
    \caption{Qualitative comparison of 12-hour forecasts on a GlobalForecast test example. Weather fields, absolute errors, color scales, and full-field MAE follow the conventions in Figure~\ref{fig:supp_forecast_00480}.}
    \label{fig:supp_forecast_08496}
\end{minipage}\par\addvspace{6pt}

\noindent\begin{minipage}{\textwidth}
    \captionsetup{type=figure,skip=4pt}
    \centering
    \includegraphics[width=\textwidth]{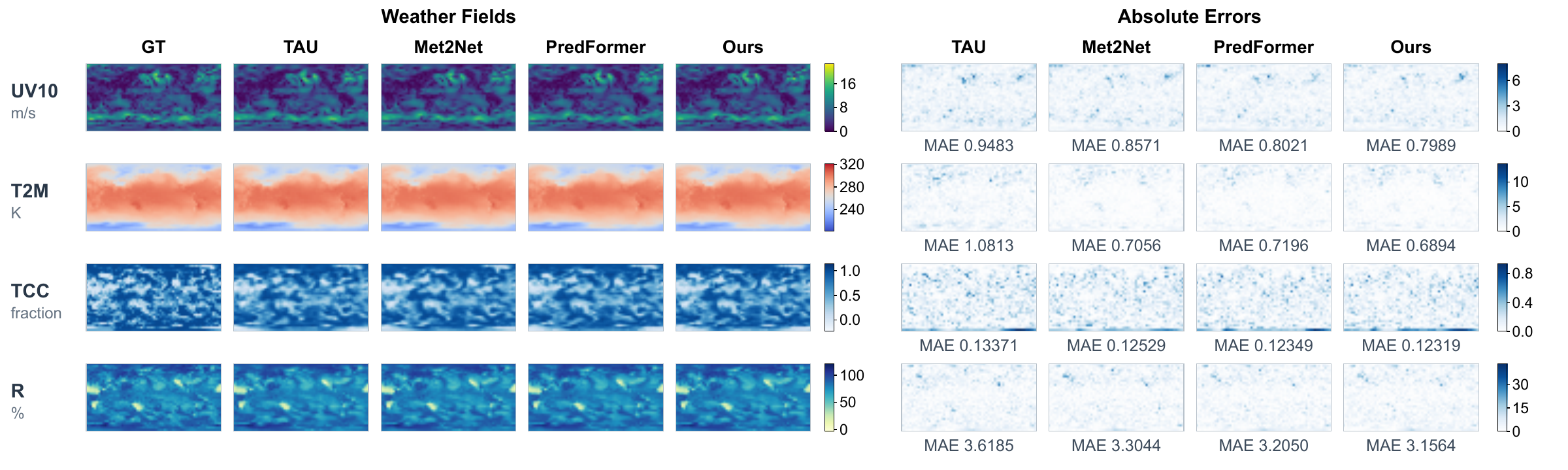}
    \caption{Qualitative comparison of 12-hour forecasts on a GlobalForecast test example. Weather fields, absolute errors, color scales, and full-field MAE follow the conventions in Figure~\ref{fig:supp_forecast_00480}.}
    \label{fig:supp_forecast_16296}
\end{minipage}\par\addvspace{6pt}

\clearpage
\noindent\begin{minipage}{\textwidth}
\subsection{Errors Across Forecast Leads}
\label{sec:supp_leadwise_mse}

Figures~\ref{fig:supp_leadwise_group_two}, \ref{fig:supp_leadwise_group_three}, and~\ref{fig:supp_leadwise_group_four} show three additional groups of GlobalForecast test cases. Each group contains one case per variable and follows the evaluation conventions in Figure~\ref{fig:leadwise_mse}. In these displayed cases, ProtoCast achieves the lowest MSE at every lead from 1 to 12 hours.

\end{minipage}\par\addvspace{6pt}

\noindent\begin{minipage}{\textwidth}
    \captionsetup{type=figure}
    \centering
    \includegraphics[width=\textwidth]{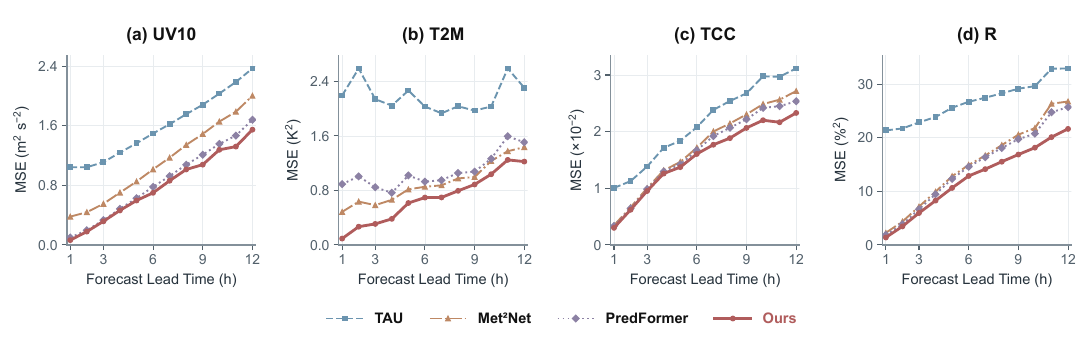}
    \caption{MSE across 1-12-hour forecast leads for an additional group of GlobalForecast test cases. Panels (a)-(d) show UV10, T2M, TCC, and R. Each panel uses one test case and follows the evaluation conventions in Figure~\ref{fig:leadwise_mse}.}
    \label{fig:supp_leadwise_group_two}
\end{minipage}\par\vspace{6pt}

\noindent\begin{minipage}{\textwidth}
    \captionsetup{type=figure}
    \centering
    \includegraphics[width=\textwidth]{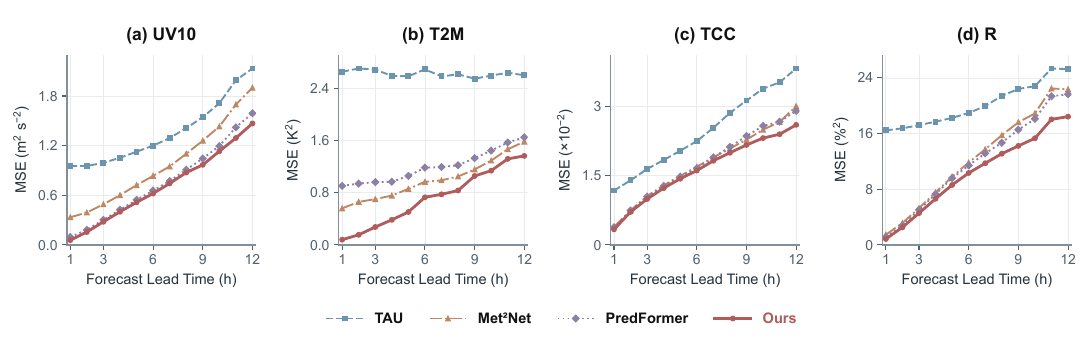}
    \caption{MSE across 1-12-hour forecast leads for an additional group of GlobalForecast test cases. Panels (a)-(d) show UV10, T2M, TCC, and R. Each panel uses one test case and follows the evaluation conventions in Figure~\ref{fig:leadwise_mse}.}
    \label{fig:supp_leadwise_group_three}
\end{minipage}\par\vspace{6pt}

\noindent\begin{minipage}{\textwidth}
    \captionsetup{type=figure}
    \centering
    \includegraphics[width=\textwidth]{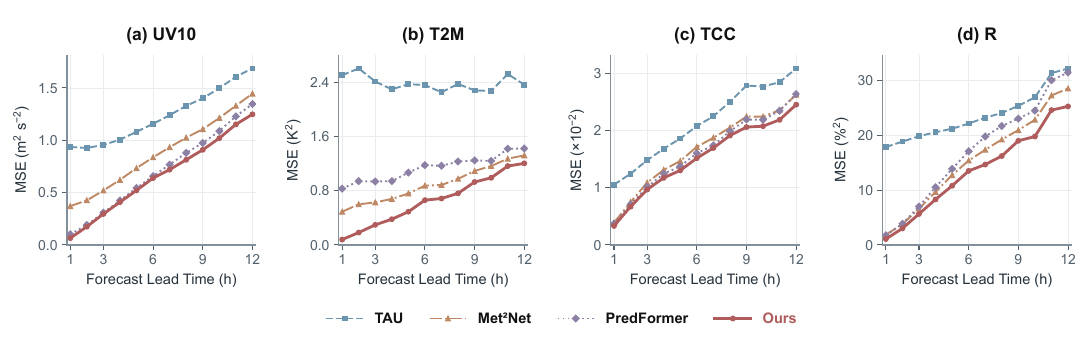}
    \caption{MSE across 1-12-hour forecast leads for an additional group of GlobalForecast test cases. Panels (a)-(d) show UV10, T2M, TCC, and R. Each panel uses one test case and follows the evaluation conventions in Figure~\ref{fig:leadwise_mse}.}
    \label{fig:supp_leadwise_group_four}
\end{minipage}\par\vspace{6pt}

\clearpage
\subsection{Additional Typhoon Track Prediction}
\label{sec:supp_typhoon_visualization}

\begin{figure}[H]
    \centering
    \begin{minipage}{0.78\textwidth}
        \centering
        \includegraphics[width=\linewidth]{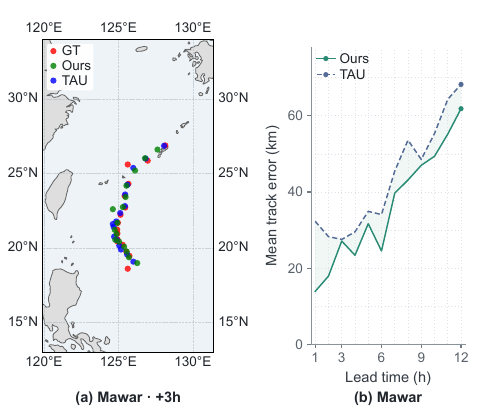}
        \caption{Typhoon track prediction for Mawar (2023), comparing TAU and Ours. (a) Three-hour predicted centers and ground-truth (GT) centers extracted from ERA5 target MSL fields. (b) Mean Haversine track error at leads of 1-12 hours. Both panels use 16 forecasts initialized at six-hour intervals.}
        \label{fig:supp_typhoon_mawar}
    \end{minipage}
\end{figure}

Figure~\ref{fig:supp_typhoon_mawar} complements the Doksuri comparison in Figure~\ref{fig:typhoon_tracks} with a Mawar comparison against TAU. The two panels follow the same center-extraction and averaging procedure described in Section~\ref{sec:supp_typhoon_evaluation}. Across the 16 evaluated Mawar start times, ProtoCast achieves lower mean track error than TAU at each forecast lead from 1 to 12 hours.

We measure track error as the great-circle distance between predicted and ERA5 reference centers. Let $(\hat{\phi}_{i,h},\hat{\lambda}_{i,h})$ and $(\phi_{i,h},\lambda_{i,h})$ denote their latitudes and longitudes in radians for initialization $i$ and forecast lead $h$. With $\Delta\phi_{i,h}=\hat{\phi}_{i,h}-\phi_{i,h}$ and $\Delta\lambda_{i,h}=\hat{\lambda}_{i,h}-\lambda_{i,h}$, the mean Haversine error is:
\begin{equation}
\begin{aligned}
 a_{i,h}={}&\sin^2\!\left(\frac{\Delta\phi_{i,h}}{2}\right)\\
 &+\cos\hat{\phi}_{i,h}\cos\phi_{i,h}\sin^2\!\left(\frac{\Delta\lambda_{i,h}}{2}\right),\\
 E_h={}&\frac{2R_{\oplus}}{N}\sum_{i=1}^{N}\arcsin\sqrt{a_{i,h}}.
\end{aligned}
\label{eq:supp_track_error}
\end{equation}
Here, $R_{\oplus}=6371.0088$km and $N$ is the number of forecast initializations, with $N=16$ for Mawar and $N=10$ for Doksuri. Each curve reports $E_h$ separately at leads of 1-12 hours. Lower values indicate more accurate predicted centers.

\clearpage
\subsection{Prototype Matching and Forecast Contributions}
\label{sec:supp_prototype_cases}

We present the main-text example in Figure~\ref{fig:prototype_t2m_case} and nine additional GlobalForecast examples. Panel (a) projects training queries, prototype keys, and the current query into two dimensions using PCA. The displayed weights come from cosine-based matching in the $d_q$-dimensional space defined in Section~\ref{sec:method_matching}. Panel (b) shows the current history and three retrieved training histories per displayed prototype. Each curve is obtained by subtracting its own final observation from every value in the history, placing its final point at zero. Prototype IDs are group-specific.

Following Section~\ref{sec:supp_prototype_analysis}, we subtract the largest-magnitude weighted contribution from one group and patch at all 12 leads. Matching weights and other contributions remain fixed, and downstream processing is recomputed without retraining. Panel (c) reports per-lead MAE by taking absolute errors at each of the 16 cells in the $4\times4$ patch before averaging. U10 and V10 use separate axes in m\,s$^{-1}$, T2M uses K, and TCC and R use percentage points (pp). Across all ten examples, the mean removal-induced MAE increase at 7-12 hours exceeds that at 1-6 hours.

The UV10 examples in Figure~\ref{fig:supp_prototype_uv10_cases} assign weights of 0.87 to P14 and 0.60 to P40, respectively. Both wind components show larger average removal effects at 7-12 hours. Figure~\ref{fig:supp_prototype_tcc_cases} shows a sustained contribution from P15 throughout the forecast and a more pronounced later effect from P52. The humidity examples in Figure~\ref{fig:supp_prototype_r_cases} also benefit from their local terms. For P44, the mean MAE increase is 0.1468 pp at 1-6 hours and 2.0723 pp at 7-12 hours.

Figure~\ref{fig:supp_prototype_t2m_cases} complements the main-text P27 example with three temperature cases. The P20 example shows warming histories, the P26 example assigns weights of 0.45 and 0.36 to P26 and P51, and the P23 example associates late historical cooling with its retrieved references. Retaining the largest contribution lowers MAE at every lead in all four temperature examples. At each patch, the same matching weights act on lead-conditioned features in Equation~\eqref{eq:method_decomposition}, producing different corrections across leads. Connecting actual weights and training references to these forecast changes makes local prototype contributions more interpretable.

\noindent\begin{minipage}{\textwidth}
    \captionsetup{type=figure,skip=4pt}
    \centering
    \includegraphics[width=\textwidth,trim=0bp 6bp 0bp 12bp,clip]{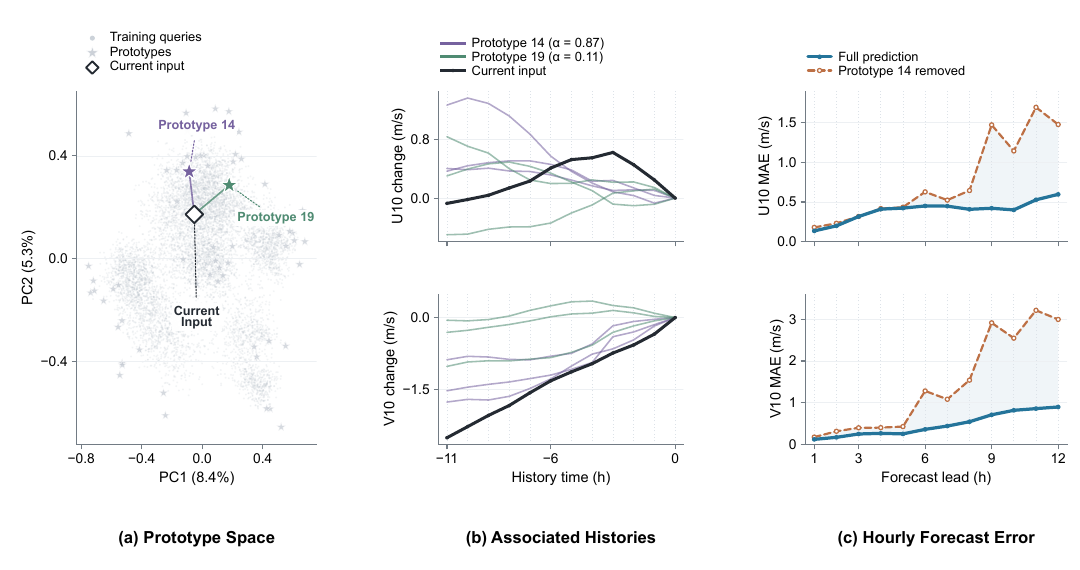}
    \caption{UV10 prototype contribution: removing P14. Panels show prototype space, associated histories, and hourly patch MAE. Blue and orange denote the full model and contribution removal.}
    \label{fig:supp_prototype_uv10_cases}
\end{minipage}\par\addvspace{2pt}

\noindent\begin{minipage}{\textwidth}
    \captionsetup{type=figure,skip=4pt}
    \ContinuedFloat
    \centering
    \includegraphics[width=\textwidth,trim=0bp 6bp 0bp 12bp,clip]{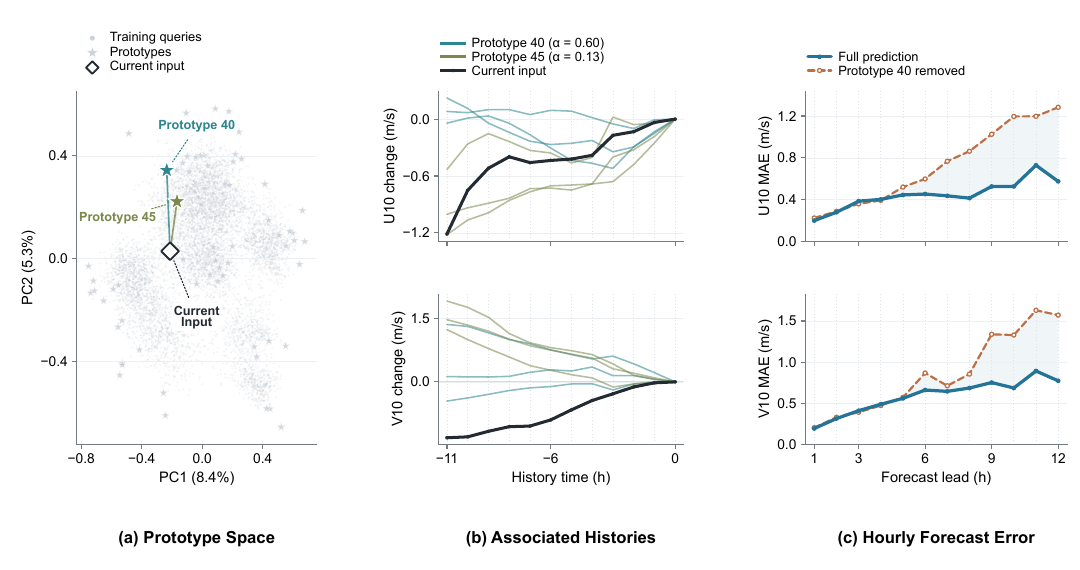}
    \caption{\textit{Continued.} Prototype contributions to UV10 forecasting, removing P40. Panel conventions follow the preceding example.}
\end{minipage}\par\addvspace{2pt}

\noindent\begin{minipage}{\textwidth}
    \captionsetup{type=figure,skip=4pt}
    \centering
    \includegraphics[width=\textwidth,trim=0bp 6bp 0bp 12bp,clip]{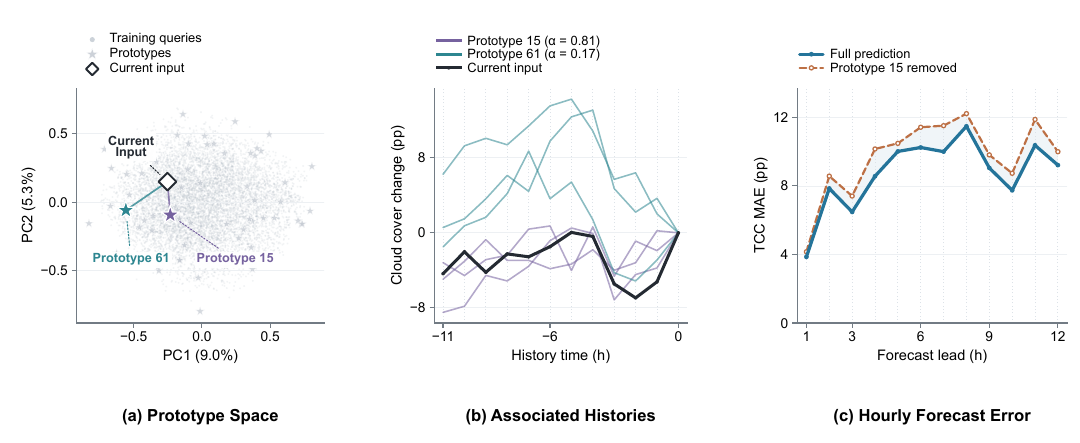}
    \caption{Prototype contributions to TCC forecasting, removing P15. Panel conventions follow Figure~\ref{fig:supp_prototype_uv10_cases}. MAE is measured in percentage points.}
    \label{fig:supp_prototype_tcc_cases}
\end{minipage}\par\addvspace{2pt}

\noindent\begin{minipage}{\textwidth}
    \captionsetup{type=figure,skip=4pt}
    \ContinuedFloat
    \centering
    \includegraphics[width=\textwidth,trim=0bp 6bp 0bp 12bp,clip]{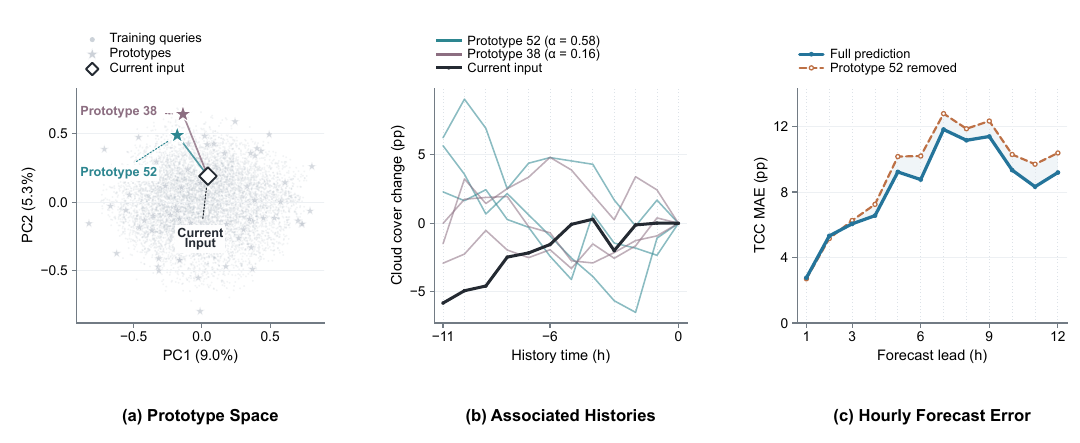}
    \caption{\textit{Continued.} Prototype contributions to TCC, removing P52. Conventions and units follow the preceding example.}
\end{minipage}\par\addvspace{2pt}

\noindent\begin{minipage}{\textwidth}
    \captionsetup{type=figure,skip=4pt}
    \centering
    \includegraphics[width=\textwidth,trim=0bp 6bp 0bp 12bp,clip]{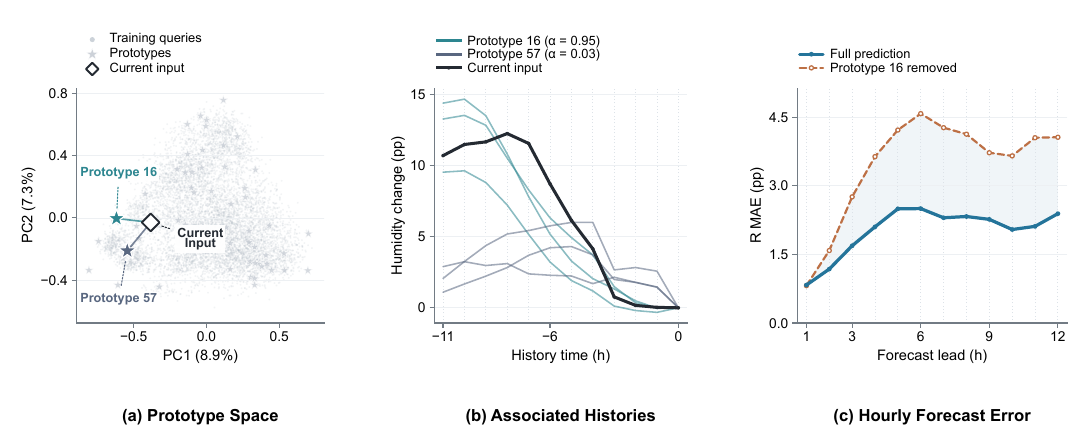}
    \caption{Prototype contributions to relative-humidity forecasting, removing P16. Panel conventions follow Figure~\ref{fig:supp_prototype_uv10_cases}. MAE is measured in percentage points.}
    \label{fig:supp_prototype_r_cases}
\end{minipage}\par\addvspace{2pt}

\noindent\begin{minipage}{\textwidth}
    \captionsetup{type=figure,skip=4pt}
    \ContinuedFloat
    \centering
    \includegraphics[width=\textwidth,trim=0bp 6bp 0bp 12bp,clip]{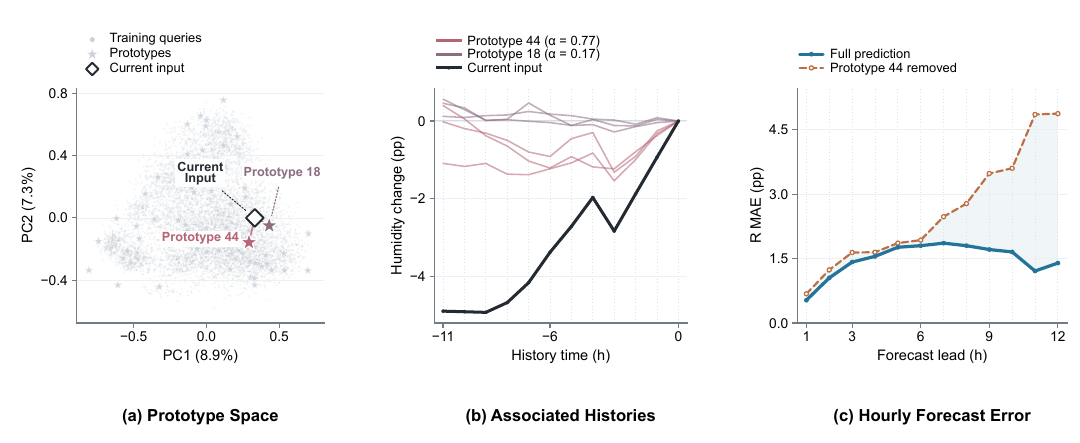}
    \caption{\textit{Continued.} Prototype contributions to relative-humidity forecasting, removing P44. Panel conventions and units follow the preceding example.}
\end{minipage}\par\addvspace{2pt}

\noindent\begin{minipage}{\textwidth}
    \captionsetup{type=figure,skip=4pt}
    \centering
    \includegraphics[width=\textwidth,trim=0bp 6bp 0bp 12bp,clip]{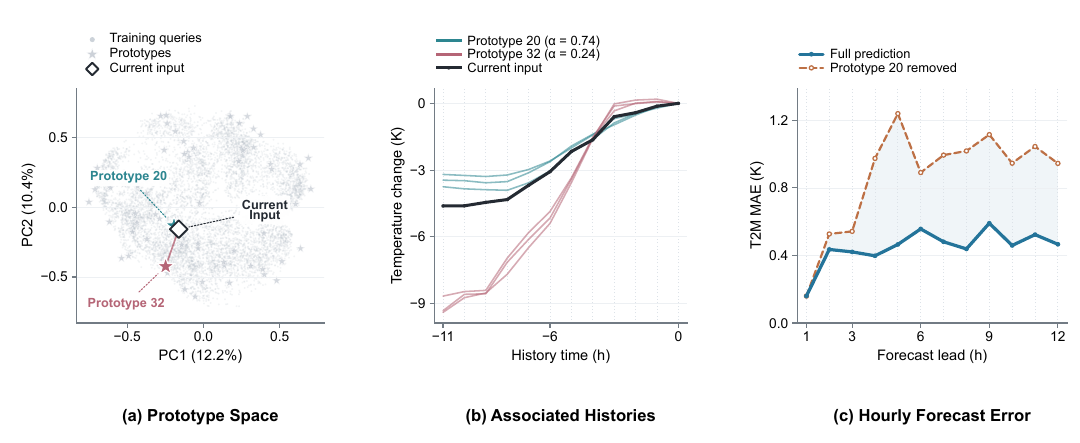}
    \caption{Additional T2M example of prototype matching and forecast contributions, removing P20.}
    \label{fig:supp_prototype_t2m_cases}
\end{minipage}\par\addvspace{2pt}

\noindent\begin{minipage}{\textwidth}
    \captionsetup{type=figure,skip=4pt}
    \ContinuedFloat
    \centering
    \includegraphics[width=\textwidth,trim=0bp 6bp 0bp 12bp,clip]{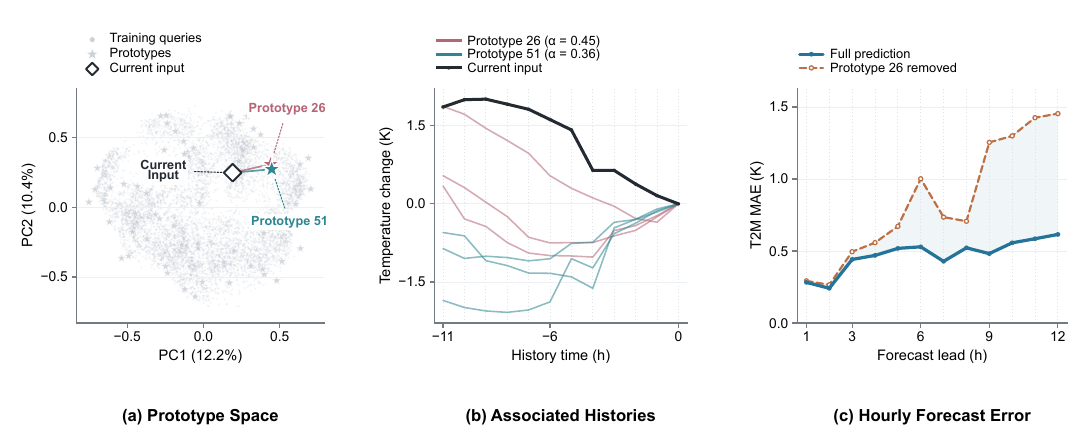}
    \caption{\textit{Continued.} Additional T2M example of prototype matching and forecast contributions, removing P26.}
\end{minipage}\par\addvspace{2pt}

\noindent\begin{minipage}{\textwidth}
    \captionsetup{type=figure,skip=4pt}
    \ContinuedFloat
    \centering
    \includegraphics[width=\textwidth,trim=0bp 6bp 0bp 12bp,clip]{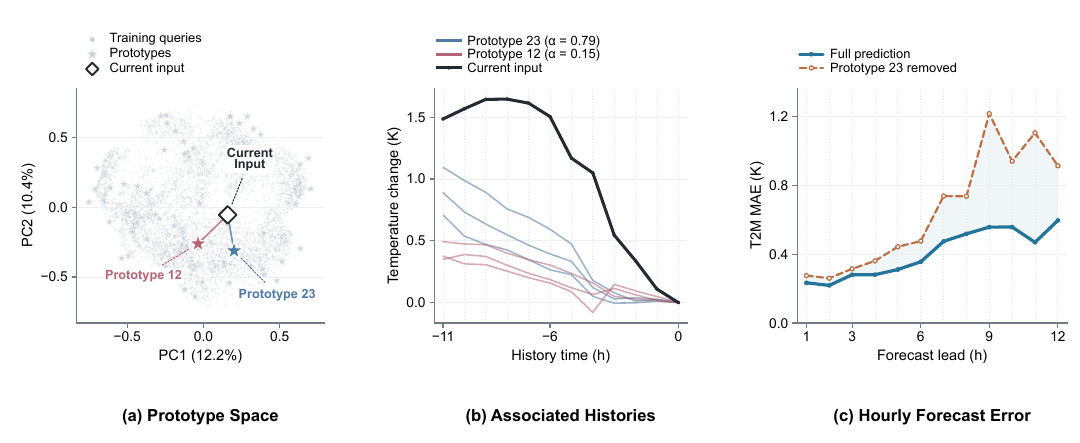}
    \caption{\textit{Continued.} Additional T2M example of prototype matching and forecast contributions, removing P23.}
\end{minipage}\par\addvspace{2pt}

\end{document}